\documentclass{article}

\usepackage{microtype}
\usepackage{graphicx}
\usepackage{subcaption}
\usepackage{booktabs} 
\usepackage[hyperindex,breaklinks]{hyperref}
\hypersetup{hidelinks}

\usepackage{algorithm}
\usepackage{algorithmic}
\usepackage{amsmath}
\usepackage{amssymb}
\usepackage{mathtools}
\usepackage{amsthm}
\usepackage[capitalize,noabbrev]{cleveref}
\usepackage{enumitem}
\usepackage{mathtools}
\usepackage{comment}
\usepackage{xcolor}         
\usepackage{todonotes}
\usepackage{longtable}

\usepackage[main, final]{neurips_2026}

\newcommand{\Fcal}{\mathcal{F}}

\newcommand{\Ncal}{\mathcal{N}}
\newcommand{\Ocal}{\mathcal{O}}

\newcommand{\Rcal}{\mathcal{R}}

\newcommand{\EE}{\mathbb{E}}

\newcommand{\OO}{\mathbb{O}}
\newcommand{\PP}{\mathbb{P}}

\newcommand{\RR}{\mathbb{R}}

\newcommand{\redtext}[1]{\textcolor{red}{#1}}

\theoremstyle{plain}
\newtheorem{theorem}{Theorem}[section]
\newtheorem{proposition}[theorem]{Proposition}
\newtheorem{lemma}[theorem]{Lemma}
\newtheorem{corollary}[theorem]{Corollary}
\theoremstyle{definition}

\newtheorem{assumption}[theorem]{Assumption}
\theoremstyle{remark}
\newtheorem{remark}[theorem]{Remark}

\newcommand{\E}{\mathbb{E}}
\newcommand{\R}{\mathbb{R}}
\DeclareMathOperator{\dist}{dist}

\DeclareMathOperator{\op}{op}

\title{Factor Augmented High-Dimensional SGD}

\author{%
  Shubo Li\\
  Department of Statistics\\
  The Pennsylvania State University\\
  University Park, PA 16802, USA\\
  \texttt{skl6034@psu.edu}
  \And
  Yuefeng Han
  \quad Xiufan Yu\\ 
  Department of Applied and Computational\\
  Mathematics and Statistics\\
  University of Notre Dame\\
  Notre Dame, IN 46556, USA\\
  \texttt{\{yuefeng.han,xiufan.yu\}@nd.edu}
}

\begin{document}

\maketitle

\begin{abstract}
Stochastic gradient descent (SGD) is a fundamental optimization algorithm widely used in modern machine learning.
In this paper, we propose \emph{Factor-Augmented SGD} (FSGD), a new optimization method that leverages latent factor representations in high-dimensional learning tasks.
Unlike standard two-stage dimension reduction approaches that rely on offline representation learning and full data storage, a key novelty of FSGD is that it operates in a single pass over streaming data, with an online Oja warm-up before downstream SGD begins, making it scalable to large-scale and high-dimensional problems.
Furthermore, we establish a new theoretical framework that explicitly incorporates latent factor estimation error into the analysis of SGD, and provide moment convergence in $\ell^s$ norm under decaying step sizes and mini-batch updates.
Our results provide a new foundation for employing SGD reliably and scalably in high-dimensional machine learning systems.
\end{abstract}

\section{Introduction}
\label{sec: Introduction}
Stochastic gradient descent (SGD) has been a cornerstone of machine learning since the pioneering work of \citet{robbins1951stochastic}. 
Beyond its algorithmic simplicity and scalability, SGD has also become a central object of theoretical study, with refined analyses linking its dynamics to implicit regularization, generalization performance, and algorithmic stability. 
For decades, theoretical analyses of SGD have largely resided within the realm of classical stochastic approximation \citep{polyak1992acceleration,lai2003stochastic,bottou2018optimization}, where the data dimension is considered fixed while the sample size tends to infinity. 
While this regime has yielded foundational insights, it no longer fully reflects the characteristics of modern learning systems.
Contemporary applications often operate in regimes where data dimension, sample size, and model complexity grow together, calling for new theoretical tools and perspectives that go beyond traditional asymptotic analyses.

In this study, we focus on the learning tasks involving high-dimensional predictors. When SGD is applied directly to such data, the dimensionality of the feature space propagates into the optimization process, resulting in a high-dimensional (HD) parameter space. 
Algorithmically, one trending strategy is to approximate the gradient updates using a low-rank representation to reduce memory costs and accelerate computation \citep{wang2018atomo, vogels2019powersgd,kozak2019stochastic,kasiviswanathan2021sgd,zhao2024galore}. 
Theoretically, despite the vast literature on SGD, convergence guarantees of HD-SGD remain limited \citep{garrigos2023handbook,li2025statistical}. 

Despite recent progress, the behavior of SGD in HD regimes remains only partially understood, and its direct application to high-dimensional datasets can be empirically fragile.
Constructing models directly in high-dimensional feature spaces therefore leads to excessive parameter complexity and exacerbates the curse of dimensionality \citep{arous2021online, schmidt2020nonparametric}. Moreover, models defined on high-dimensional feature spaces often involve a large number of parameters. Applying SGD in such high-dimensional parameter regimes is computationally costly and is known to exhibit unstable or oscillatory behavior around flat minima \citep{hoffer2017train, wu2018sgd, cohen2021gradient}.

To address the challenges induced by HD predictors, instead of directly optimizing HD-SGD in the high-dimensional feature space, we propose a new strategy, named \textbf{Factor-Augmented Stochastic Gradient Descent (FSGD)}, that augments SGD with a factor model. 
The key insight is that the effective information content of HD predictors often lies in a much lower-dimensional latent space, which can be captured by the ``factors''. Ignoring this structure leads to inefficient updates that waste computation on noisy directions, slow convergence along dominant subspaces, and obscure the true geometry of the optimization landscape. By modeling predictors dynamically through a factor decomposition, we can extract useful representations from predictors and design SGD updates that adapt to the intrinsic dimensionality of the problem.

Integrating factor models into SGD is far from a straightforward adaptation, but involves several non-trivial challenges to be carefully addressed. 

First, existing factor-augmented learning approaches  \citep{fan2024factor,zhou2025covariate,zhou2025factor,luo2025factor,luo2026supervised} in other applications typically adopt a two-stage paradigm, where a low-dimensional representation is first learned using an offline procedure, and then treated as fixed during downstream learning tasks. 
Such two-stage approaches rely on full access to the dataset and require storing or repeatedly scanning all samples, which is inefficient for iterative SGD updates and may even be infeasible in large-scale learning settings where data arrive sequentially and memory resources are limited.
To tackle this challenge, we propose an online factor-learning procedure designed for a one-pass streaming setting and seamlessly integrate it with SGD updates, enabling memory-efficient and scalable learning.

Second, a key advantage of factor models is that they explicitly allow for the presence of idiosyncratic errors, capturing variation in predictors that is not explained by a small number of common factors. 
Most existing works in representation learning literature focus exclusively on learning low-dimensional embeddings and implicitly absorb idiosyncratic components into the representation itself, ignoring both the error structure and the associated estimation uncertainty. 
There is limited understanding of how idiosyncratic noise and representation estimation error interact with stochastic optimization procedures, particularly when SGD is employed. Our work bridges this gap by rigorously quantifying how representation estimation error and idiosyncratic noise propagate into the convergence rate.

These limitations raise a fundamental question: 

\vspace{-0.5ex}
    {\it Can the latent factor structure of high-dimensional predictors be explicitly utilized to improve the efficiency and theoretical understanding of SGD in high dimensions?}

In this work, we provide a positive answer to the above question. 
We summarize our main contributions as follows.

\begin{enumerate}[leftmargin = *]
    \item \textbf{Factor-Augmented SGD (FSGD).} We propose a new optimization method that utilizes the latent structure of high-dimensional predictors. Specifically, our method learns a low-dimensional latent representation via online PCA and simultaneously updates the model parameters using SGD on the induced factor space. 
    While our theory is developed for strongly convex population objectives, the neural network experiments illustrate the empirical robustness of the FSGD principle beyond the convex setting. 


     \item \textbf{Streaming unsupervised factor learning coupled with SGD.} 
    We develop a streaming learning framework in which both unsupervised latent factors and model parameters are updated using SGD-type procedures after an online Oja warm-up. Unlike classical two-stage representation learning approaches that require an offline PCA step with full data access, FSGD operates on streaming data and does not store the entire dataset, crucial for practical large-scale learning scenarios. As a result, factor estimates can be continuously refined during model training rather than being fixed a priori, causing the population minimizer to shift over time. In Proposition \ref{prop: Stability of theta* in Oja's Update}, we establish a novel bound for minimizer shift to address the nontrivial statistical dependence between factor estimation and gradient updates.


    \item \textbf{New theoretical guarantee for SGD with latent factor estimation.}
    Theorem \ref{thm:Moment_Convergence_of_SGD} provides a new $q$-th moment convergence analysis of SGD that accounts for both latent factor estimation error and idiosyncratic noise in high-dimensional settings. Unlike existing approaches that treat the learned representation as fixed or known and ignore the idiosyncratic component in the factor structure, we rigorously quantify how the representation estimation error and idiosyncratic noise propagate into the convergence rate of SGD.
\end{enumerate}

\subsection{Related Works}

\paragraph{HD-SGD.} 
While the theory of SGD has been extensively studied in traditional low-dimensional settings \citep{bottou2018optimization,garrigos2023handbook}, theoretical understanding of HD-SGD remains underdeveloped, with only a handful of recent works. \citet{paquette2021sgd,paquette2022implicit,paquette2025homogenization} study HD-SGD under quadratic loss functions. 
\citet{li2025statistical} studies HD-SGD and its Ruppert-Polyak averaged variant, and provides guarantees on moment convergence and high-probability concentration bounds in HD regimes with constant learning rates. Our work provides a rigorous characterization of convergence by explicitly accounting for data dimensionality via factor models, a completely different analytical route from existing studies, offering a fresh theoretical perspective on SGD with HD predictors. 
Methodologically, we also emphasize that the proposed FSGD fundamentally differs from the SGD with low-rank gradient approximation \citep{vogels2019powersgd,kozak2019stochastic,kasiviswanathan2021sgd}: FSGD is a data-side approach that utilizes the latent structure of high-dimensional predictors, whereas the latter is an optimization-side method that enforces low-rank structure on gradient updates.





\vspace{-1ex}
\paragraph{Factor Models.} Factor models \citep{lawley1962factor,bai2003,bai2008large,han2024tensor,han2024simultaneous,han2024cp,han2023tensor,han2022rank,yu2024dynamic,yu2024power,bolivar2025threshold,chen2025modewise,chen2026estimation} have long served as a foundational tool to capture the underlying structure of high-dimensional data through a smaller set of latent variables (a.k.a., factors). In time series forecasting literature, they are often employed as a first step to extract low-dimensional representations from multivariate inputs, which are then fed into subsequent learning algorithms for prediction \citep{stock2002forecasting,stock2002macroeconomic,bair2006prediction,fan2017sufficient,luo2022inverse,yu2022nonparametric,chen2026diffusion}. 
We emphasize that FSGD fundamentally differs from these approaches: existing factor-augmented learning approaches adopt a straightforward two-stage paradigm, while FSGD involves an online factor-learning step that better aligns with the data-adaptive and stochastic nature of SGD updates.

\vspace{-2ex}
\paragraph{Notation}
For $v \in \R^d$, we denote the $l^s$-norm of $v$ by $|v|_s:= (\sum_{i=1}^d |v_i|^s )^{1/s}$. Write $v^{\odot s}=(v_1^s, v_2^s, \cdots,v_d^s)^\top$. For a matrix $A \in \R^{d_1 \times d_2}$, let $\|A\|_F$ denote its Frobenius norm, and $\|A\|_{\op}$ denote its operator norm. We write $\mathbb{O}_{d,k}$ to be all $d \times k$ orthonormal matrix and denote $\mathbb{O}_{k}=\mathbb{O}_{k,k}$. For any matrices with orthonormal columns  $V,W\in\R^{d\times k}$, we define the subspace distance
\begin{equation}
    \label{eq: subspace distance}
    \dist(W,V)
    :=
    \|VV^\top - WW^\top\|_{\op} .
\end{equation}
The expectation and covariance of random vectors are respectively denoted by $\E[\cdot]$ and $\operatorname{Cov}(\cdot)$. For $q > 0$ and a random variable $X$, we write $L^q$-norm of $X$ by $\|X\|_q:=(\E[|X|^q])^{1/q}$, and $X \in L^q$ iff $\|X\|_q < \infty$. For random variable $X\in \R$, we define sub-Gaussian norm as $\|X\|_{\psi_2}:=\inf\{t>0:\E[\exp(X^2/t^2)]\leq2 \}$. For random vector $Z \in \R^d$, we define $\|Z\|_{\psi_2}:=\sup_{|u|_2=1}\| \langle u,Z \rangle\|_{\psi_2}$. 

For two positive number sequences $(a_n)$ and $(b_n)$, we say $a_n = \Ocal(b_n)$ (resp. $a_n = \Theta( b_n)$) if there exists $c>0$, such that $a_n / b_n \leq c$ (resp. $1/c \leq a_n/b_n \leq c$) for all large $n$. We call $c$ an absolute constant if it is independent of all problem parameters, including $n$, $d$, $t$, and other model-specific quantities.
Throughout the paper, $C$ denotes an absolute constant, unless stated otherwise. A summary of the main notation is provided in Appendix~\ref{app:notation}.


\vspace{-1ex}
\section{Factor Augmented SGD}
\label{sec: Factor Augmented SGD}
\subsection{Factor Augmented Regression Model}
\label{subsec: Latent Factor Model}
High-dimensional learning poses both statistical and computational challenges. Statistically, when learning a $(\beta,C)$-smooth function, deep neural networks achieve the optimal nonparametric rate of $n^{-\frac{2\beta}{2\beta+d}}$ with sample size $n$, which deteriorates rapidly as the ambient dimension
$d$ grows. To address this challenge, we adopt a factor model for the covariate process that exploits the latent low-dimensional structure underlying the observed data $\{x_i,y_i \}$. Specifically, we assume
\begin{equation}
\label{eq:factor model}
y_i = \mathcal M(f_i) + \varepsilon_i, \text{ and } x_i = Bf_i + u_i,
\end{equation}
where $f_i, u_i,\varepsilon_i$ are mutually independent random vectors, $B \in \RR^{d \times k}$ is an unknown factor loading matrix, $f_i \in \RR^k$ is a vector of latent factors, $u_i \in \RR^d$ is the idiosyncratic component, and $\mathcal M$ is the regression function. 
Under this model, the regression function $\mathcal M$ is defined on the $k$-dimensional latent factor space, with a corresponding oracle nonparametric rate of $n^{-\frac{2\beta}{2\beta+k}}$ when the factors are observed. Algorithmically, building the downstream model on this low-dimensional space allows SGD to optimize a model with fewer parameters, while learning and applying the projection requires a cost linear in $d$ for fixed $m$ and $k$.
We denote $\Sigma:=\operatorname{Cov}(X), \; \Sigma_f := \operatorname{Cov}(f)$, and let $V \in \RR^{d \times k}$ be the orthogonal matrix whose columns are the leading $k$ eigenvectors of $\Sigma$. The following regularity conditions on the factor model in \eqref{eq:factor model} are standard in the literature \citep{bai2003,bai2012statistical}.

\begin{assumption}
\label{asp: factor model regularity}
    The idiosyncratic component $u_i \in \R^d$ is a zero-mean sub-Gaussian random vector with $\psi_2$-Orlicz norm $\|u_i\|_{\psi_2} \leq K_u$. The loading matrix satisfies $d^{-1} B^\top B = I_k $, and $\EE[f_i] = 0.$ Moreover, the minimum eigenvalue of $\Sigma_f$ is strictly positive, i.e $\lambda_k(\Sigma_f)>0$.
\end{assumption}

The following proposition establishes the convergence rate of the estimated factors.

\begin{proposition}[Factor Estimation]
    \label{prop: Factor Estimation}
    Suppose the factor model satisfies \eqref{eq:factor model} and Assumption \ref{asp: factor model regularity}. Given $Q \in \RR^{d\times k}$ with orthonormal columns, define the factor estimator for a new observation $x$ as
\(
\hat f := d^{-1/2} Q^\top x.
\)
Then there exists a rotation matrix $R\in\R^{k\times k}$, for $q\geq2$, such that
{
\small
\begin{equation}
\label{eq: Factor Estimation in Prop}
    \begin{aligned}
            &\big\|\,|\hat f-Rf|_s\,\big\|_{q}
\le 
     \sqrt{2} k^{\frac{1}{2}} \| \operatorname{dist}(Q,V) \|_q\big\|\,|f|_2\,\big\|_{q}+
     C \Big(
\frac{K_u^2}{ \lambda_k(\Sigma_f)} \big\||f|_2\,\big\|_{q}
+
K_u   ( k^{\frac{1}{2}}+q^{\frac{1}{2}}) \Big) d^{-\frac{1}{2}}.
    \end{aligned}
\end{equation}
}
\end{proposition}

The first term in \eqref{eq: Factor Estimation in Prop} captures the effect of the subspace distance between $Q$ and $V$, while the second term accounts for both the subspace distance between $V$ and $d^{-1/2}B$, and the contribution of the idiosyncratic components. 
The sub-Gaussian condition on $u_i$ bounds the noise covariance and the projected idiosyncratic noise, thereby controlling the second term. A smaller $\lambda_k(\Sigma_f)$ enlarges the bound on the subspace distance between $V$ and $d^{-1/2}B$, while heavier-tailed $u_i$ may worsen both bounds in the second term.
When $Q$ is obtained via PCA and ${\rm Cov}(f)$ has distinct eigenvalues, the factors $f$ are identifiable up to sign flips; equivalently, the indeterminacy reduces to $R\in{\rm diag}(\{1,-1\}^{k})$.


\vspace{-1ex}
\subsection{Online PCA}
\vspace{-1ex}

Principal component analysis (PCA) is widely used to estimate latent factors in high-dimensional models. However, classical PCA becomes computationally infeasible when both the sample size $n$ and the ambient dimension $d$ are large, or when data arrive in a streaming fashion. To address these challenges, we adopt a streaming PCA approach for online factor estimation.

Let $\{A_t\}_{t\ge1}\subset\mathbb{R}^{d\times d}$ be a sequence of i.i.d. symmetric random matrices with $\mathbb{E}[A_t]=\Sigma$, and let $\{\eta_t^{\mathrm{oja}}\}_{t\ge1}$ denote the step sizes. Initialize with any $Z_0\in\R^{d\times k}$ of full column rank
and set $Q_0=\mathrm{QR}[Z_0]$, where $\mathrm{QR}[\cdot]$ denotes the $Q$-factor of the thin QR decomposition. For $t\ge1$, Oja's algorithm \citep{oja1982simplified, oja1985stochastic} updates 
\begin{equation}
\label{eq:oja_update}
Q_t := \mathrm{QR}[(I+\eta_t^{\rm oja} A_t)\,Q_{t-1}].
\end{equation}

Let $V\in\R^{d\times k}$ denote the matrix whose columns are the $k$ leading eigenvectors of $\Sigma$.  We impose the following conditions on Oja's updates throughout this paper.

\begin{assumption}
\label{asp: ojs_assump}
For Oja's update, we assume 
\begin{enumerate}[leftmargin=*, itemsep=0pt, topsep=0pt]
    \item There exists a constant $M>0$ such that, 
    \vspace{-1ex}
    \[
        \sup_{P\in\mathbb{O}_{d,k}}
        \| P^{\top}(A_t - \Sigma) \|_{F}
        =
        \Big( \sum_{i=1}^k \sigma_i^2(A_t-\Sigma) \Big)^{\frac{1}{2}}
        \le M
    \]
    holds almost surely, where $\sigma_1(X)\ge\cdots\ge\sigma_d(X)$ denote the singular values of $X$. 

    \item (Warm start) The initialization $Q_0$ in stage-II satisfies
        $\dist(Q_0,V) \le \frac{1}{2}.$
\end{enumerate}
\end{assumption}
\vspace{-1ex}
The condition
\(
\sup_{ P \in\mathbb{O}_{d,k}}
\| P^{\top}(A_t - \Sigma) \|_{F} \leq M \text{ a.s.}
\)
 is standard in the analysis of online PCA; see, e.g.,
\citet{jain2016streaming, allen2017first, huang2021streaming, liu2022dp}.
Under the factor model in \eqref{eq:factor model} and Assumption \ref{asp: factor model regularity}, with $A_t=m^{-1}\sum_{i=1}^m x_i x_i^\top$ and sub-Gaussian $(f_i,u_i)$, one may take \(M = \mathcal{O}(m^{-1/2} d \log d)\) a.s.; see Remark~3.4 of \citet{liu2022dp} for further discussion. The warm-start condition in Assumption~\ref{asp: ojs_assump}(ii) is also standard and requires the initial estimate to be sufficiently close to the population principal subspace. In particular, it is guaranteed by Theorem 3 of \cite{huang2021streaming} after a burn-in period. This condition is used later in Proposition~\ref{prop: Stability of theta* in Oja's Update} to control how subspace updates affect the population minimizer.

Adapted from \citet{huang2021streaming}, the following lemma provides an estimation error bound for Oja's algorithm in terms of the subspace distance between the estimated and true principal subspaces.

\begin{lemma}[$L^q$-Moment Bound for Oja Update]
\label{lemma: Lq moment bound for oja update - main}
    Suppose Assumption \ref{asp: ojs_assump} holds. Let $\lambda_1 \geq \cdots \geq \lambda_d$ be the eigenvalues of $\Sigma$ and define the eigengap $\rho_k = \lambda_k - \lambda_{k+1}$. Set $\eta_t^{\rm oja}
=
\frac{\alpha}{(\beta+t)\rho_k}$ and define the normalized gap
    \(
\bar\rho_k
:=
\min\left\{
\frac{\rho_k}{M},
\frac{\rho_k}{\|\Sigma\|_{\op}},
1
\right\}.
\)
If
\[
\alpha\ge 8,
\qquad
\beta\ge
\max \big\{ \frac{4(1+\sqrt{2e})\alpha}{\bar\rho_k}, \,
C\cdot \frac{\alpha^2}{\bar\rho_k^2} \big\},
\]
where $C$ is an absolute constant. Then for every $q \geq 2$ and $t \geq 1$, it holds that
\[
\left\|\operatorname{dist}(Q_t,V)\right\|_q
\le
k^{1/q}
\left[
\left(\frac{\beta+1}{\beta+t}\right)^{\alpha/2}
+
C\cdot\frac{\alpha q^{1/2}}{\bar\rho_k}\cdot
\frac{t^{1/2}}{\beta+t}
+
C^{2/q} \cdot \alpha^{2/q}\bar\rho_k^{-2/q}
\exp\left\{
-\frac{\beta\bar\rho_k^2}{\alpha^2 C^2q}
\right\}
\right].
\]
\end{lemma}

\subsection{Factor Augmented SGD}
\label{subsec: Factor Augmented SGD}

When the latent factor structure in \eqref{eq:factor model} is present, it is natural to consider a factor-augmented stochastic gradient descent (FSGD) framework in which the loss function is defined on the latent factors rather than the high-dimensional covariates. Specifically, given factor representations $f_{t,i}$ and responses $y_{t,i}$, we consider a loss function of the form $l(\theta; f_{t,i}, y_{t,i})$. The optimization problem of interest is
\begin{align}\label{eq:opt_problem}
\theta^*= \arg\min_{\theta}L(\theta;I),\ \text{where}\  L(\theta;R)=\EE l(\theta; R f_{t,i},y_{t,i}),    
\end{align}
$\theta\in\RR^{p}$, $p\ll d$, and $R\in\OO_{k}$ denoting a rotation matrix. The rotation matrix $R$ accounts for the rotational ambiguity inherent in factor models; setting $R=I$ corresponds to the ideal case where the factors are perfectly identified.
At iteration $t$, given a mini-batch $B_t = \{(x_{t,i}, y_{t,i})\}_{i=1}^m$ of size $m$, the FSGD update takes the form
\begin{align}
\theta_{t+1} =\theta_t - \eta_t \cdot \frac{1}{m} \sum_{i=1}^m \nabla_\theta l(\theta_t; \hat f_{t,i}, y_{t,i}),    
\end{align}
where $\hat f_{t,i}$ denotes an estimated latent factor obtained from the observation $x_{t,i}$. Since the true factors $f_{t,i}$ are not directly observed, they must be estimated online from the streaming data via a dimension reduction procedure.

\begin{algorithm*}[ht]
\caption{Factor Augmented SGD (FSGD)}
\label{alg:FASGD}
\begin{algorithmic}[1]
\REQUIRE Step sizes $\{\eta_t,\eta^{{\rm oja},(0)},\eta_t^{\rm oja}\}$, factor dimension $k$, mini-batch size $m$
\item[] \textbf{Stage I: Online PCA Warm-up}
\STATE Initialize $Z_{0}\in\R^{d\times k}$ of full column rank and set $Q_{0}^{(0)}=QR[Z_0]$.
\FOR{$s=1,2,\dots,T_0$}
    \STATE Sample an independent mini-batch $B_{s}^{(0)}=\big\{x_{s,i}^{(0)}\big\}_{i=1}^m$ from model \eqref{eq:factor model}. 
    
    \STATE Form the mini-batch matrix $X_s^{(0)}=[x_{s,1}^{(0)},\ldots,x_{s,m}^{(0)}]\in\mathbb{R}^{d\times m}$.
    \STATE Update subspace $Q_s^{(0)} \gets QR\!\left[Q_{s-1}^{(0)}+\frac{\eta^{\rm oja,(0)}}{m}X_s^{(0)}\left((X_s^{(0)})^\top Q_{s-1}^{(0)}\right)\right]$.
    
\ENDFOR
\STATE Set $Q_0 \gets Q_{T_0}^{(0)}$. 

\vspace{0.3em}
\item[] \textbf{Stage II: Joint Online PCA and Mini-batch SGD}
\STATE Initialize $\theta_0$
\FOR{$t=1,2,\dots,T$}
    \item[] \textbf{Mini-batch SGD:}
    \STATE Sample an independent mini-batch $B_t=\{(x_{t,i},y_{t,i})\}_{i=1}^m$ from model \eqref{eq:factor model}. 
    \STATE Estimate factors: $\hat f_{t,i} \gets d^{-1/2} Q_{t-1}^\top x_{t,i}$ for all $(x_{t,i},y_{t,i}) \in B_t$.
    \STATE Compute stochastic gradient:
    \(
    g_t \gets \frac{1}{m}\sum_{i=1}^m \nabla_\theta l \big(\theta_t;\hat f_{t,i},y_{t,i}\big).
    \)
    \STATE Update parameters:
    \(
    \theta_{t+1} \gets \theta_t - \eta_t\, g_t.
    \)
    \item[] \textbf{Online PCA:}
    
    \STATE Form the mini-batch matrix $X_t=[x_{t,1},\ldots,x_{t,m}]\in\mathbb{R}^{d\times m}$.
    \STATE Update subspace $Q_t \gets QR\!\left[Q_{t-1}+\frac{\eta^{\rm oja}_{t-1}}{m}X_t\left(X_t^\top Q_{t-1}\right)\right]$.
    
    \STATE (Optional) Procrustes alignment $S_t=Q_{t}^\top Q_{t-1}$, compute SVD $S_t=U_{S_t}\Lambda_{S_t} V_{S_t}^\top$, and then set $Q_t \gets Q_t U_{S_t} V_{S_t}^\top$.
\ENDFOR
\end{algorithmic}
\end{algorithm*}


The FSGD proceeds in two stages, detailed in Algorithm~\ref{alg:FASGD}.

\vspace{-1.5ex}
\paragraph{Stage I: Initialization.}
The goal of the first stage is to obtain a good initialization satisfying the warm-start condition in Assumption~\ref{asp: ojs_assump}. One approach is to initialize $Z_0$ with i.i.d. Gaussian entries and run Oja's algorithm for a sufficient burn-in period. During this burn-in, mini-batches are processed sequentially without storing or replaying past data.
When offline initialization is permitted, one may perform offline PCA on a preliminary dataset to estimate the loading matrix, as studied in \citet{yang2025theoretical}, and use the resulting subspace estimator to initialize $Q_0$ in the second stage.

\vspace{-1.5ex}

\paragraph{Stage II: Joint Optimization.}
In this stage, at each iteration $t$, given the subspace estimate $Q_{t-1}\in\mathbb R^{d\times k}$ obtained from the previous $t-1$ mini-batches, we perform linear dimension reduction to extract $k \ll d$ latent factors by projecting each observation onto the estimated factor space:
\[
\hat f_{t,i} = d^{-\frac{1}{2}} Q_{t-1}^\top x_{t,i} \in \mathbb{R}^k,
\]
where $x_{t,i}$ denotes the $i$-th  sample in the mini-batch at iteration $t$. The stochastic gradient is then computed as $\nabla_\theta l(\theta; \hat f_{t,i}, y_{t,i})$. To fully utilize the data, the same mini-batch $B_t$ is reused to perform the Oja update in Step~15 of Algorithm~\ref{alg:FASGD}, yielding an updated subspace estimate $Q_t$. We note that each Oja update costs $O(mdk+dk^2)$ time. Specifically, computing $X_t^\top Q_{t-1}$ and then $X_t(X_t^\top Q_{t-1})$ costs $O(mdk)$ in total, while the thin QR decomposition costs $O(dk^2)$. Storing the subspace estimate $Q_t$ requires $O(dk)$ memory.

It is important to note that recovery of the factor loading matrix is subject to the intrinsic rotational ambiguity of factor models. Specifically, given the orthogonal projection matrix $Q_{t-1}$ at iteration $t$, there exists an orthogonal rotation matrix $R_{t-1} \in \OO^k$ (depending on $Q_{t-1}$) such that the estimated factor $\hat f_{t,i}=d^{-\frac{1}{2}} Q_{t-1}^\top x_{t,i}$ is close to the rotated true factor $R_{t-1} f_{t,i}$; see Proposition~\ref{prop: Factor Estimation} for a formal statement. In offline PCA, if the leading $k$ eigenvalues of the covariance matrix are distinct, then the factor loading matrix is uniquely determined up to sign changes. In our online PCA setting, we could perform a Procrustes alignment of $Q_t$ to $Q_{t-1}$ at each iteration. When the leading $k$ eigenvalues of $\Sigma$ are distinct and the subspace error of $Q_{t-1}$ is sufficiently small, the associated Procrustes rotation $R_t$ varies smoothly across iterations, i.e., remaining close to $R_{t-1}$. In Appendix~\ref{subsec: Proofs for Factor Augmented SGD}, we show that this additional alignment step does not affect the theoretical bounds. 

\section{Convergence of FSGD}
\label{sec: FSGD}
In this section, we establish the moment convergence property of FSGD. 
Recall the optimization problem defined in \eqref{eq:opt_problem} and the data $(x,f,y)$ from model \eqref{eq:factor model}. We begin by imposing the following assumptions on the objective function and the stochastic gradients.

\begin{assumption}[$\ell^s$ norm - Strong Convexity]
\label{asp:Strongly Convex}
    Let $s\geq2$ be an even integer. For any $R \in \mathbb{O}_k$, assume there exists a constant $\mu>0$ such that
    \begin{align*}
        &\left\langle(\theta_1 - \theta_2)^{\odot(s-1)},
       \nabla L(\theta_1;R) - \nabla L(\theta_2;R)
    \right\rangle  \geq\mu|\theta_1 - \theta_2|_s^s,
    \end{align*}
    holds for all $\theta_1,\theta_2 \in \R^p$.
\end{assumption}
A similar condition is adopted in \citet{li2025statistical}. When $s=2$, this reduces to standard strong convexity. For example, consider the least squares loss under the linear model \(y_i = f_i^\top \theta + \varepsilon_i.\)
The Hessian of the population loss is given by
\[
\nabla_\theta^2 L(\theta;R) = R \Sigma_f R^\top,
\quad \text{where } \Sigma_f := \mathbb{E}[f_i f_i^\top].
\]
Thus, a sufficient condition for strong convexity is $\lambda_{\min}(\Sigma_f) > 0$.

\begin{assumption}[$\ell^s$ norm - Lipschitz Continuity for $\theta$]
\label{asp:Lipschitz Continuity for theta}
Let $s\geq2$ be an even integer and $q \geq 2$.
For $(x,f,y)$ from model \eqref{eq:factor model}, assume there exists a constant $L_{s,\theta}>0$ such that
{\small
\begin{align*}
    &\Big\|| \nabla_\theta \,l(\theta_1;Rf,y) - \nabla_\theta \,l(\theta_2;Rf,y) |_s\Big\|_q \leq L_{s,\theta} \cdot \Big\| |\theta_1 - \theta_2|_s \Big\|_q, \\
    &\Big\|| \nabla_\theta \,l(\theta_1;d^{-1/2}Q^\top x,y) - \nabla_\theta \,l(\theta_2;d^{-1/2}Q^\top x,y) |_s\Big\|_q \leq L_{s,\theta} \cdot \Big\| |\theta_1 - \theta_2|_s \Big\|_q, 
\end{align*}
}
hold for all $R \in \OO_k, Q\in \OO_{d,k}$.
\end{assumption}

\begin{assumption}[Lipschitz Continuity for $f$]
\label{asp:Lipschitz Continuity for f}
Let $s\geq2$ be an even integer and $q \geq 2$.
For random vectors $y, f_1,f_2$ from model \eqref{eq:factor model}, there exists a constant $L_{s,f}>0$ such that
    \[
    \Big\| | \nabla_\theta \,l(\theta; f_1,y) - \nabla_\theta \,l(\theta;f_2,y) |_s \Big\|_q \leq L_{s,f} \cdot \Big\| |f_1 - f_2|_s \Big\|_q,
    \]
holds uniformly for all $\theta$.
\end{assumption}

Assumptions~\ref{asp:Strongly Convex} and \ref{asp:Lipschitz Continuity for theta} are standard in the analysis of SGD and have appeared extensively in the literature; see, e.g., \citet{ruppert1988efficient} and \citet{polyak1992acceleration}.

As discussed earlier, factor loading estimation is subject to rotational ambiguity.
Accordingly, given $Q_{t-1}$ obtained from the first $t-1$ Oja's updates, we define $\theta^*_{t-1}$ as the minimizer of the rotated population risk
\begin{align}
\theta^*_{t-1}&=\arg\min_{\theta} L(\theta;R_{t-1}) , \label{op:rotated}\\
L(\theta;R_{t-1})&=\mathbb E_z  \left[ l(\theta; R_{t-1} f_{t,i}, y_{t,i}) \,\middle|\, R_{t-1} \right],     
\end{align}
where $R_{t-1} \in \OO_k$ is the Procrustes rotation aligning $Q_{t-1}$ with the oracle loading matrix $V$, and $\E_z[\cdot]$ denotes the expectation taken with respect to $(f_{t,i},y_{t,i})$ only. Our goal is to characterize how close the FSGD iterate $\theta_t$ is to $\theta^*_{t-1}$, which represents a rotated version of $\theta^*$ corresponding to the rotation $R_{t-1}$. 

A key challenge arising from Oja's updates is that the rotation matrix $R_t$ varies across iterations, causing the population minimizer $\theta_t^*$ to shift over time. 
To quantify this effect, the warm-start condition and Assumption~\ref{asp:ojas_step_size} allow us to bound the change in $R_t$ induced by each update of $Q_t$. Proposition~\ref{prop: Stability of theta* in Oja's Update} then uses this bound to control the \emph{minimizer shift}, defined as the difference between successive minimizers $\theta_t^*$ and $\theta_{t-1}^*$. 
Specifically, the following result shows that $\| |\theta_{t}^*-\theta_{t-1}^*|_s \|_q$ is of the same order as the Oja's step size $\eta_t^{\mathrm{oja}}$ multiplied by the perturbation bound $M_A$. 
Consequently, the drift of the target minimizer remains sufficiently controlled, allowing the algorithm to retain its convergence guarantees.

\begin{assumption}
\label{asp:ojas_step_size}
Let $M_A<\infty$ be such that $\|A_t\|_{\rm op}\le M_A$ a.s. for all $t$. Assume the Oja's step size in Stage II is non-increasing and satisfies
\[
\eta_0^{\rm oja} \le \frac{1}{4(\sqrt{2}+1)M_A} k^{-2}.
\] 
\end{assumption}

The upper bound on $\eta_0^{\mathrm{oja}} \le \frac{1}{4(\sqrt{2}+1)M_A} k^{-2}$ depends on the uniform bound for \(A_t\), rather than for \(A_t-\Sigma\). This condition is required to control the perturbation error introduced by the QR normalization step.



\begin{proposition}[Stability of \(\theta^*\) under Oja's update]
\label{prop: Stability of theta* in Oja's Update}
Suppose Assumptions \ref{asp: ojs_assump},
\ref{asp:Strongly Convex}, \ref{asp:Lipschitz Continuity for f} and  \ref{asp:ojas_step_size} hold.
Set $\eta_t^{\rm oja} = \frac{\alpha}{(\beta+t)\rho_k}$ 
with
\[
\beta
\ge
C\bar\rho_k^{-2}
\left[
q\log d+\log\left(\frac{k}{\bar\rho_k^2}\right)
\right], \qquad \alpha \geq 8 .
\]
Then, for every \(1\le t\le T\) with \(\log T=\Ocal(\log d)\),
\[
\Big\|\,\big|\theta_t^*-\theta_{t-1}^*\big|_s\,\Big\|_q
\le
C\frac{L_{s,f}}{\mu}M_f
\left(
M_Ak^{1/2}\eta_t^{\rm oja}
+
d^{-1/2}T^{-1}
\right),
\]
where \(M_f:=\||f|_2\|_q\).
In particular, if \(\eta_t^{\rm oja}\le c_k/t\) for some $c_k$ independent of $t$, then
\[
\Big\|\,\big|\theta_t^*-\theta_{t-1}^*\big|_s\,\Big\|_q
\le
C\frac{L_{s,f}}{\mu}M_f(M_Ak^{1/2}c_k+d^{-1/2}) t^{-1}.
\]
\end{proposition}

We now state the first main result of this paper. Theorem~\ref{thm:Moment_Convergence_of_SGD} establishes the moment convergence of FSGD iterates $\theta_t$ to the rotated oracle $\theta_{t-1}^*$.

\begin{theorem}[Moment Convergence of FSGD]
\label{thm:Moment_Convergence_of_SGD}
    Suppose Assumptions \ref{asp: factor model regularity}, \ref{asp: ojs_assump}, \ref{asp:Strongly Convex}, \ref{asp:Lipschitz Continuity for theta}, \ref{asp:Lipschitz Continuity for f} and \ref{asp:ojas_step_size} hold. Let the step sizes $\{ \eta_t \}_{t=1}^\infty$ be non-negative and monotonically non-increasing, satisfying $\sum_{i=0}^\infty \eta_i = \infty, t \eta_{t-1} \to \infty$, $\eta_0 \leq \frac{\mu}{9( {\max \{q,s \} -1})L^2_{s,\theta}}$. Assume $\EE |\nabla_\theta l(\theta; f_i,y_i)|_q^q < \infty$ for any $\theta$. Set $\eta_t^{\rm oja} = \frac{\alpha}{(\beta+t)\rho_k}$ with 
    \[
    \beta \ge
 \max \Big\{  C\bar\rho_k^{-2} \big[ q\log d+\log(\frac{k}{\bar\rho_k^2}) \big], \frac{4(1+\sqrt{2e})\alpha}{\bar\rho_k}, \,
C \frac{\alpha^2}{\bar\rho_k^2} \Big\},
    \]
    where $C,\alpha$ are absolute constant with $\alpha \geq 8$.
    Under polynomial step-size $\eta_t = \tilde{\Ocal}((t+1)^{-\gamma})$ for some $\gamma \in (0,1)$ and \(\log T=\Ocal(\log d)\), we have the convergence rate for any $1\leq t \leq T$,
    \begin{equation}
        \begin{aligned}
\label{eq: Theorem 3.4 rate}
\Big\||\theta_t-\theta_{t-1}^*|_s\Big\|_q&\leq A_1d^{-1/2}+A_2(m^{-1/2}+d^{-1/2})t^{-\gamma/2}\\&+A_3 (C_{\alpha,\beta}+\frac{q^{1/2}}{\bar \rho_k}) \, t^{-1/2}+A_4 M_A \rho_k^{-1} t^{-(1-\gamma)}. 
\end{aligned}
    \end{equation}
where the constants \(A_i\) collect the remaining quantities that are independent of \(d\), \(m\), and \(t\) when \(k,q,s\) and the model constants are fixed. Their explicit definitions are given in \eqref{eq: Ai definitions} in Appendix~\ref{subsec: App Proofs for Section Sec FSGD}. $C_{\alpha,\beta} = \Ocal(\beta^{1/2})$ defined in \eqref{eq: def of C alpha beta}. 


\end{theorem}

\begin{remark}\label{rmk:sgdrate}
Suppose the conditions of Theorem~\ref{thm:Moment_Convergence_of_SGD} hold. Consider sub-Gaussian $(f_i,u_i)$ in \eqref{eq:factor model}, and assume $k=\Theta(1)$ and $m\ll d$. Then \(M = \mathcal{O}(m^{-1/2} d \log d) \text{ a.s.}, M_A = \Ocal(d \log d)\), $\rho_k = \Theta(d), \bar \rho_k^{-1} = \Theta(\max\{ m^{-1/2} \log d, 1\})$. Thus, $q \log^3 (d)$ is a sufficient upper order of $\beta$.
Therefore, the rate in \eqref{eq: Theorem 3.4 rate} becomes
\begin{align}\label{eq:fsgd}
\Big \| | \theta_t - \theta^*_{t-1} |_s \Big \|_q \lesssim&  d^{-\frac{1}{2}} + m^{-\frac12}t^{-\frac{\gamma}{2}}   + \log^{3/2}(d) \, t^{-\frac{1}{2}} +  \log( d) \, t^{-(1-\gamma)}.
\end{align}
The above bound groups the errors according to their final rates rather than by mutually exclusive sources of error. The first term contains the idiosyncratic component in factor estimation; see Proposition~\ref{prop: Factor Estimation}, and also absorbs the bad-event contribution from the Oja's random updates by choosing $\beta$ sufficiently large. 
The second term represents the standard convergence rate of mini-batch SGD \citep{tang2023acceleration}, which scales with the batch size and learning rate.
The third term captures the error propagation from Oja's online PCA into the FSGD updates. By Lemma~\ref{lemma: Lq moment bound for oja update - main}, this term consists of the warm start transient term, summarized by \(C_{\alpha,\beta}t^{-1/2}\), and the stochastic Oja term \(\bar\rho_k^{-1}t^{-1/2}\).
The fourth term accounts for the drift in the time-varying optimal rotation matrix $R_t$ relative to the population minimizer $\theta^*$.
By balancing second and fourth error terms, the optimal rate is $d^{-\frac{1}{2}}+t^{-\frac{1}{3}}$ achieved at $\gamma = \frac{2}{3}$.
\end{remark}




\begin{remark}
The $\ell^s$-norm provides a smooth surrogate for the $\ell^\infty$-norm, which is widely used in high-dimensional statistics; see, e.g., \citet{fan2001variable, wainwright2019high, chen2025concentration, li2025statistical}. Specifically, let $s = s_p$, where
\(
s_p := 2 \min\{ \ell \in \mathbb{N} : 2\ell > \log p \}.
\)
Then the $\ell^{s_p}$-norm and the $\ell^\infty$-norm are equivalent in the sense that
$\|x\|_\infty\le \|x\|_{s_p} \le e\,\|x\|_\infty .$
Consequently, by setting $s = s_p$, the moment bounds established in Theorem~\ref{thm:Moment_Convergence_of_SGD} immediately yield $\ell^\infty$-norm guarantees.    
\end{remark}




The fourth term in the convergence bounds \eqref{eq: Theorem 3.4 rate} and \eqref{eq:fsgd} arises from the time-varying target induced by online factor loading updates. This term is undesirable as it does not benefit from larger learning rates. A practical remedy is to allocate the first half of the iterations to latent subspace learning and freeze the projection matrix after iteration $t_1=\lfloor t/2 \rfloor$. As demonstrated in Corollary~\ref{cor:Moment_Convergence_of_SGD-freeze_Oja} below, this strategy ensures exponential decay of the instability induced by the time-varying minimizer $\theta_t^*$.


\begin{corollary}[Moment Convergence of FSGD (Frozen)]
\label{cor:Moment_Convergence_of_SGD-freeze_Oja}
    Suppose the assumptions of Theorem~\ref{thm:Moment_Convergence_of_SGD} hold. Consider polynomial step size $\eta_t = \tilde{\Ocal}((t+1)^{-\gamma})$ for some $\gamma \in (0,1)$, and suppose Oja's updates are frozen after iteration $t_1$. Define
    \vspace{-1ex}
\begin{align*}
F_{t_1}:=&C_2M_f t_1^{-\frac{1}{2}}+C_3d^{-1/2}, \qquad
\Xi_{t_1}^2:=C_{s,q}^2 m^{-1}+ L_{s,f}^2 F_{t_1}^2.    
\end{align*}
Then for the $\lfloor ta \rfloor$-th iterate with $t_1/t < a\leq 1$, we have the following convergence rate
    \vspace{-1ex}
\begin{equation}
\label{eq: cor--free Oja's rate}
    \begin{aligned}
            \Big \| | \theta_{\lfloor ta \rfloor} - \theta^*_{ t_1 -1} |_s \Big \|_q &\leq  \sqrt{\frac{4}{\mu}(\frac{8}{\mu}L_{s,f}^2+1)}F_{t_1}  + 2C_1 \Xi_{t_1} \cdot t^{-\frac{\gamma}{2}} , 
                \vspace{-1ex}
    \end{aligned}
\end{equation}
where constants $C_1,C_2,C_3,C_{s,q}$ do not depend on $t$ and are collected in Remark \ref{remark: proof constants}. 
In particular, if $t_1=\lfloor t/c_0 \rfloor$ for $c_0>1$, $m\ll d$ and $m\lesssim t^{1-\gamma}$, then \eqref{eq: cor--free Oja's rate} simplifies to
\vspace{-1ex}
\begin{align*}
 \Big \| | \theta_{\lfloor ta \rfloor} - \theta^*_{ t_1 -1} |_s \Big \|_q   \lesssim&  d^{-\frac{1}{2}} + m^{-\frac12}t^{-\frac{\gamma}{2}}.
\end{align*}
\end{corollary}
\begin{remark}
The term $F_{t_1}$ arises from factor estimation error, which decreases as the ambient dimension $d$ grows and the subspace estimation improves, see the bound on $ \| |\hat f - R f|_s \|_{q}$ in Proposition~\ref{prop: Factor Estimation} for details. 
Our theoretical results also apply to the case of offline PCA, where $F_{t_1}$ can be replaced by the corresponding factor estimation bound and depends on the sample size of the offline PCA rather than $t_1$. As noted in Remark~\ref{rmk:sgdrate}, the error propagation from Oja’s online PCA decays faster than the standard SGD rate. Consequently, one may choose $t_1$ of smaller order than $t$ while still matching or improving upon the mini-batch SGD rate $m^{-1/2} t^{-\gamma/2}$.
\end{remark}

Compared with the rate in Theorem~\ref{thm:Moment_Convergence_of_SGD}, the projection-frozen regime mitigates the effect of minimizer shift. As a result, it may exhibit improved convergence behavior when $\gamma$ is large and the dimension $d$ is high, at the cost of a larger constant in the factor estimation error term $F_{t_1}$.

\section{Experiments}
\label{sec: Experiments}
\vspace{-0.5ex}
We evaluate FSGD on both synthetic and real data. Complete experimental details, baseline configurations, sensitivity analyses, and additional neural network experiments are deferred to Appendix~\ref{app: simulation} and Appendix~\ref{app:real_data_details}.

\subsection{Synthetic Experiments}
\label{subsec:simulation}
We first test FSGD under a linear factor model that matches our theoretical assumptions. The goal is to verify the scaling behavior stated by Theorem~\ref{thm:Moment_Convergence_of_SGD}. We generate covariates from the factor model~\eqref{eq:factor model} and generate responses from a linear model. Unless otherwise specified, we use $k=3$, mini-batch size $m=5$, and average results over 100 repetitions. We also evaluate predictive performance and computational efficiency in a nonlinear neural network framework; details are in the Appendix~\ref{app:nn_simulation}.

Figure~\ref{fig: Linear SGD combined} summarizes two key phenomena. First, when fixing $\gamma=0.6$ and increasing $d$, the rotated parameter error decreases with the ambient dimension, consistent with the $d^{-1/2}$ term in Theorem~\ref{thm:Moment_Convergence_of_SGD}. Second, when fixing $d=100$ and varying the SGD decay exponent $\gamma$, the best performance is attained near $\gamma\approx 2/3$, matching the trade-off described in Remark~\ref{rmk:sgdrate}: small $\gamma$ leads to slow SGD convergence, while large $\gamma$ amplifies minimizer drift induced by online factor learning.

\begin{figure}[t]
    \centering
    \begin{subfigure}[t]{0.45\linewidth}
        \centering
        \includegraphics[width=\linewidth]{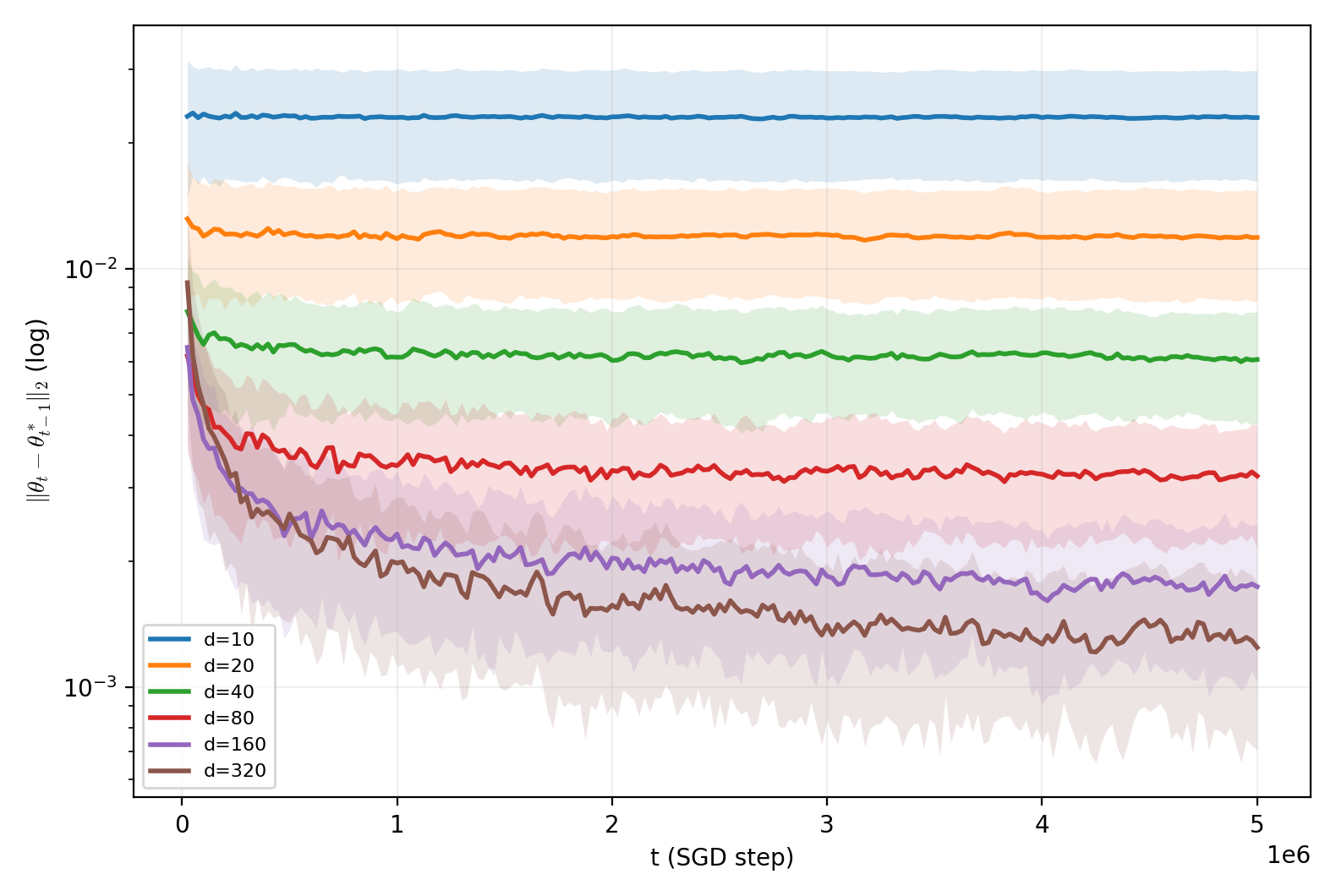}
        \caption{Varying $d$}
        \label{fig: Linear SGD d}
    \end{subfigure}
    \hfill
    \begin{subfigure}[t]{0.45\linewidth}
        \centering
        \includegraphics[width=\linewidth]{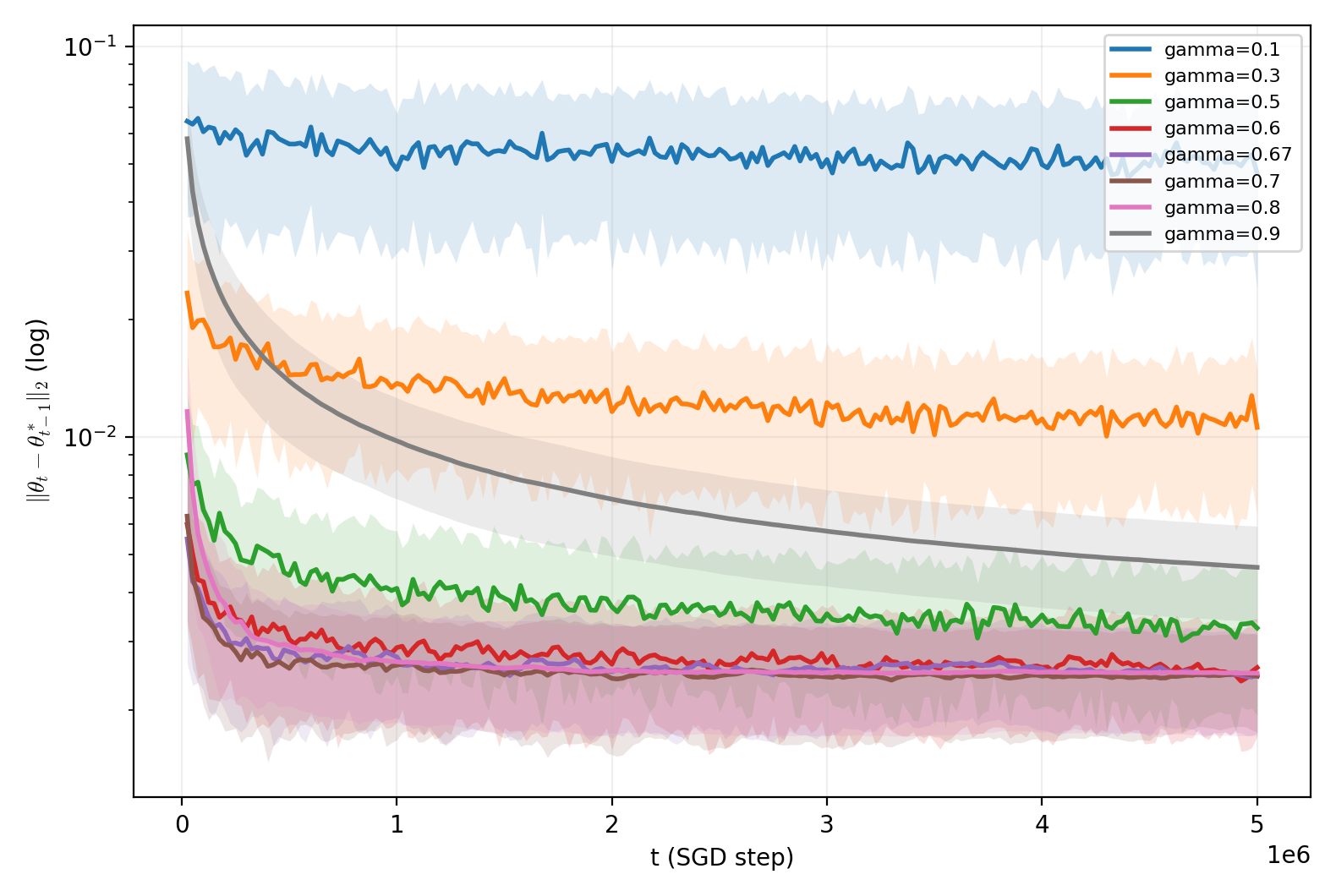}
        \caption{Varying $\gamma$}
        \label{fig: Linear SGD gamma}
    \end{subfigure}
    \vspace{-1ex}
    \caption{Numerical validation of theoretical properties for FSGD. The results exhibit dimension and step-size scaling that aligns with the derived theoretical rates in Section \ref{sec: FSGD}. }
    \label{fig: Linear SGD combined}
    \vspace{-1ex}
\end{figure}

\vspace{-1ex}





\subsection{Real-Data Experiment: MNIST}
\label{subsec:real-data}

We evaluate FSGD on MNIST using one-pass image classification.
For each of 30 random splits, the official training set is partitioned into 320 unlabeled warm-up images, 54,680 supervised training images and 5,000 validation images.
Each split is evaluated using five training seeds, with the official 10,000-image test set fixed.
Images are flattened into $d=784$ dimensions.
Projected methods use $\hat{k}=64$ and a $64\to128\to10$ ReLU network, while Vanilla-NN uses $784\to128\to10$.
All methods minimize cross-entropy loss using plain mini-batch SGD with batch size 32 and learning rate $\eta_t=t^{-0.3}$.

\begin{table}[ht]
\centering
\small
\setlength{\tabcolsep}{6pt}
\caption{MNIST test accuracy (\%) after one pass. Accuracy is mean (SD) across 30 split means, each averaging five training seeds. Gains are Factor-NN minus each baseline, in percentage points.}
\label{tab:mnist_main}
\begin{tabular}{lcc}
\toprule
Method & Test accuracy & Gain [95\% CI] \\
\midrule
Factor-NN
& \textbf{96.24 (0.07)}
& --- \\
Factor-NN-F0
& 95.71 (0.10)
& 0.535 [0.509, 0.561] \\
NN-Stiefel
& 95.96 (0.09)
& 0.284 [0.264, 0.303] \\
Vanilla-NN
& 95.82 (0.05)
& 0.418 [0.397, 0.440] \\
Random Projection
& 93.44 (0.10)
& 2.801 [2.763, 2.840] \\
\bottomrule
\end{tabular}
\end{table}

Table~\ref{tab:mnist_main} compares Factor-NN, which continues Oja updates throughout training, with four baselines: Factor-NN-F0 freezes the projection after warm-up; NN-Stiefel jointly learns the projection and classifier from the supervised loss while maintaining orthonormal projection columns; Vanilla-NN trains directly on the raw pixels; and Random Projection uses a fixed random orthogonal projection.
In particular, the superior performance of Factor-NN supports the benefit of continued covariance-based representation learning for classification.
These results provide empirical evidence for the effectiveness of FSGD on real image data beyond the strongly convex setting covered by our theory.
Complete experimental settings, learning-rate and freezing-time comparisons, and additional CIFAR-10 and NCEP results are provided in Appendix~\ref{app:real_data_details}.

\vspace{-1ex}
\section{Conclusion}
\vspace{-0.5ex}
We develop FSGD, a new optimization framework that integrates online latent factor learning with stochastic gradient descent. 
Our analysis explicitly characterizes how representation estimation error and idiosyncratic noise propagate through the optimization dynamics, revealing a fundamental trade-off between factor estimation accuracy, optimization error and minimizer shift induced by online factor learning, which is a phenomenon that does not arise in classical SGD theory with fixed covariates.
Empirical studies further confirm the strong empirical performance of FSGD.
Overall, this work advances the theoretical understanding of high-dimensional SGD and opens new directions for integrating representation learning with scalable optimization.

\section{Limitations}
\label{sec:limitations}

Despite its convergence guarantees and promising empirical results, FSGD has limitations in its modeling assumptions and theoretical scope.

First, the linear factor representation in \eqref{eq:factor model} may fail to capture predictive structure on nonlinear manifolds, despite allowing a nonlinear regression function and full-dimensional idiosyncratic noise $u_i$. Nevertheless, linear PCA remains effective in many scientific applications \citep{greenacre2022principal, zhang2024fast}. Extending FSGD to nonlinear representations is an important future direction.

Second, our theoretical guarantees assume that the true factor dimension $k$ is known and used by the algorithm. Extending the analysis to an estimated or misspecified dimension would require reformulating both the factor-estimation comparison (Proposition~\ref{prop: Factor Estimation}) and the Oja subspace analysis (Lemma~\ref{lemma: Lq moment bound for oja update - main}). This entails substantial additional arguments beyond substituting the selected dimension for $k$ in the existing bound. A rigorous treatment of these effects, together with adaptive rank selection for streaming data, remains an important direction for future research.

\clearpage
\newpage

\bibliography{reference}
\bibliographystyle{icml2026}

\newpage

\appendix

\section*{\Large Appendices}

This supplement consists of four sections. Appendix~\ref{app:notation} summarizes the main notation. Appendix~\ref{app: simulation} reports detailed simulation studies, supplementing Section~\ref{subsec:simulation}. Appendix~\ref{app:real_data_details} presents the evaluation on real-world datasets, supplementing Section~\ref{subsec:real-data}. Appendix~\ref{sec: app proofs} includes all technical lemmas and complete proofs of the propositions and theorems in Sections~\ref{sec: Factor Augmented SGD} and~\ref{sec: FSGD}.

\section{Summary of Notation}
\label{app:notation}

Table~\ref{tab:notation} summarizes the main notation.
Data-defined quantities describe the data-generating model,
the observed samples, or properties of the population objective
for a fixed loss.
Algorithmic quantities specify or are computed by FSGD.
Analysis-only quantities are introduced to state and prove
the theoretical results and need not be computed by the algorithm.
In our theoretical results, the factor dimension used by FSGD
equals the true factor dimension $k$.

{\small
\setlength{\tabcolsep}{4pt}
\renewcommand{\arraystretch}{1.15}

\begin{longtable}{
    @{}p{0.18\textwidth}
    p{0.59\textwidth}
    p{0.16\textwidth}@{}
}
\caption{Summary of the main notation.}
\label{tab:notation}\\
\toprule
Symbol & Definition & Role \\
\midrule
\endfirsthead

\multicolumn{3}{l}{\textit{Table~\ref{tab:notation} continued}}\\
\toprule
Symbol & Definition & Role \\
\midrule
\endhead

\midrule
\multicolumn{3}{r}{\textit{Continued on the next page}}\\
\endfoot

\bottomrule
\endlastfoot

$d$
& Ambient dimension of the observed covariates.
& Data-defined \\

$k$
& True factor dimension; also the factor dimension used by FSGD
in the theoretical results.
& Data-defined \\

$x_{t,i},\,y_{t,i}$
& Observed covariate and response for sample $i$ in mini-batch $t$.
& Data-defined \\

$f_{t,i},\,u_{t,i},\,\varepsilon_{t,i}$
& Latent factor, idiosyncratic component, and response noise
in the factor model.
& Data-defined \\

$B$
& Factor loading matrix in $\mathbb{R}^{d\times k}$,
normalized by $d^{-1}B^\top B=I_k$.
& Data-defined \\

$\Sigma,\,\Sigma_f$
& Covariance matrices of the observed covariate and latent factor,
respectively.
& Data-defined \\

$V$
& Matrix of the leading $k$ eigenvectors of $\Sigma$.
& Data-defined \\

$\lambda_i$
& $i$-th largest eigenvalue of $\Sigma$.
& Data-defined \\

$\rho_k$
& Population eigengap $\lambda_k-\lambda_{k+1}$.
& Data-defined \\

$K_u$
& Upper bound on the sub-Gaussian norm of the
idiosyncratic component.
& Data-defined \\

$\lambda_k(\Sigma_f)$
& Minimum eigenvalue of the factor covariance matrix.
& Data-defined \\

$M$
& Covariance-fluctuation bound in
Assumption~\ref{asp: ojs_assump}(i).
& Data-defined \\

$M_A$
& Almost-sure upper bound on $\|A_t\|_{\mathrm{op}}$
in Assumption~\ref{asp:ojas_step_size}.
& Data-defined \\

$M_f$
& Factor moment $\bigl\|\,|f|_2\,\bigr\|_q$.
& Data-defined \\

$\mu$
& Strong-convexity constant in
Assumption~\ref{asp:Strongly Convex}.
& Data-defined \\

$L_{s,\theta},\,L_{s,f}$
& Gradient Lipschitz constants for the parameter and factor inputs,
respectively, in
Assumptions~\ref{asp:Lipschitz Continuity for theta}
and~\ref{asp:Lipschitz Continuity for f}.
& Data-defined \\

$B_t$
& Observed mini-batch $\{(x_{t,i},y_{t,i})\}_{i=1}^{m}$.
& Data-defined \\

$X_t$
& Mini-batch covariate matrix
$[x_{t,1},\ldots,x_{t,m}]\in\mathbb{R}^{d\times m}$.
& Data-defined \\

$A_t$
& Mini-batch matrix $m^{-1}X_tX_t^\top$.
Its product with the subspace estimate can be evaluated
without explicitly forming $A_t$.
& Data-defined \\

\midrule

$p$
& Dimension of the model parameter $\theta$.
& Algorithmic \\

$m$
& Mini-batch size.
& Algorithmic \\

$T_0,\,T$
& Number of warm-up updates and number of Stage-II iterations,
respectively.
& Algorithmic \\

$Q_t$
& Subspace estimate after $t$ Stage-II Oja updates;
$Q_0$ is the estimate obtained after warm-up.
& Algorithmic \\

$\widehat f_{t,i}$
& Estimated factor $d^{-1/2}Q_{t-1}^\top x_{t,i}$.
& Algorithmic \\

$l(\theta;f,y)$
& Downstream loss function.
& Algorithmic \\

$\theta_t$
& FSGD parameter iterate.
& Algorithmic \\

$g_t$
& Mini-batch stochastic gradient computed using the
estimated factors.
& Algorithmic \\

$\eta_t$
& SGD step size.
& Algorithmic \\

$\eta^{\mathrm{oja},(0)}$
& Constant Oja step size during Stage-I warm-up.
& Algorithmic \\

$\eta_t^{\mathrm{oja}}$
& Oja step size during Stage II.
& Algorithmic \\

$\alpha,\,\beta$
& Parameters in the theoretical Oja schedule
$\eta_t^{\mathrm{oja}}=\alpha/[(\beta+t)\rho_k]$.
& Algorithmic \\

$\gamma$
& Decay exponent of the polynomial SGD step-size schedule.
& Algorithmic \\

$t_1$
& Iteration after which the subspace estimate is frozen
in the frozen variant.
& Algorithmic \\

\midrule

$s,\,q$
& Exponent of the parameter-space $\ell^s$ norm and
order of the $L^q$ moment, respectively.
& Analysis-only \\

$\bar\rho_k$
& Normalized eigengap
$\min\{\rho_k/M,\rho_k/\|\Sigma\|_{\mathrm{op}},1\}$.
& Analysis-only \\

$R_t$
& Population-based orthogonal rotation used to compare
estimated factors with rotated latent factors.
& Analysis-only \\

$L(\theta;R)$
& Rotated population risk
$\mathbb{E}[l(\theta;Rf,y)]$.
& Analysis-only \\

$\theta^*$
& Oracle minimizer $\arg\min_\theta L(\theta;I)$.
& Analysis-only \\

$\theta_t^*$
& Moving population minimizer
$\arg\min_\theta L(\theta;R_t)$.
& Analysis-only \\

$C$
& Absolute constant, unless stated otherwise.
& Analysis-only \\

$C_{\alpha,\beta}$
& Constant summarizing the warm-start transient
in the Oja error bound.
& Analysis-only \\

$A_1,\ldots,A_4$
& Coefficients collecting the remaining quantities
in the convergence bound of
Theorem~\ref{thm:Moment_Convergence_of_SGD}.
& Analysis-only \\

$F_{t_1},\,\Xi_{t_1}$
& Factor-estimation and gradient-error quantities
used in the frozen-variant convergence bound.
& Analysis-only \\

\end{longtable}
}

\section{Simulation}
\label{app: simulation}
The synthetic experiments serve two purposes: (i) under a linear factor model satisfying the assumptions of our theory, they verify the predicted scaling behavior in Theorem \ref{thm:Moment_Convergence_of_SGD}; and (ii) under a nonlinear neural network model, they assess practical predictive performance and computational efficiency. 
All synthetic experiments, including both the linear and neural network simulations, were run on a computational cluster powered by Intel Xeon Gold 6342 processors. The experiments used up to 100 CPU cores in parallel, with approximately 4GB of memory allocated per core. The total compute budget was on the order of \(10^4\) CPU core-hours.

\subsection{Synthetic Experiments: Linear Model}

We first conduct linear model experiments under a factor model that exactly matches our theoretical assumptions. The goal is to verify the scaling behavior stated by Theorem~\ref{thm:Moment_Convergence_of_SGD}.

\paragraph{Data-generating process.}
We strictly follow the factor model \eqref{eq:factor model}.  Specifically, let $Z \in \mathbb{R}^{d \times k}$ have i.i.d.\ standard Gaussian entries, and let $Q$ be the orthonormal factor from the QR decomposition of $Z$. We set the loading matrix to be $B=\sqrt{d}\,Q$. The latent factor $f$ and idiosyncratic component $u$ are independent and have i.i.d.\ $\mathrm{Unif}[-0.5,0.5]$ entries. The response is generated by
$y = f^\top \theta^* + \varepsilon$,
where $\theta^* \sim \mathrm{Unif}(0,1)$ and $\varepsilon \sim \mathcal N(0,0.3)$ are both independent from anything else. Each mini-batch contains $m=5$ newly generated samples.

\paragraph{Optimization details.}
In the Stage I of Algorithm \ref{alg:FASGD}, we set constant learning rate $0.01$ for $T_0=10$ iterations. In Stage II, the SGD step size follows
$\eta_t = 0.5\, t^{-\gamma}$,
and the Oja step size is
$\eta_t^{\text{oja}} = \frac{0.1}{50+t}$.
Unless otherwise specified, we fix $k=3$, run $T=5\times 10^6$ updates, and average over 100 repetitions.

\begin{figure}[H]
    \centering

    \begin{subfigure}[t]{0.45\linewidth}
        \centering
        \includegraphics[width=\linewidth]{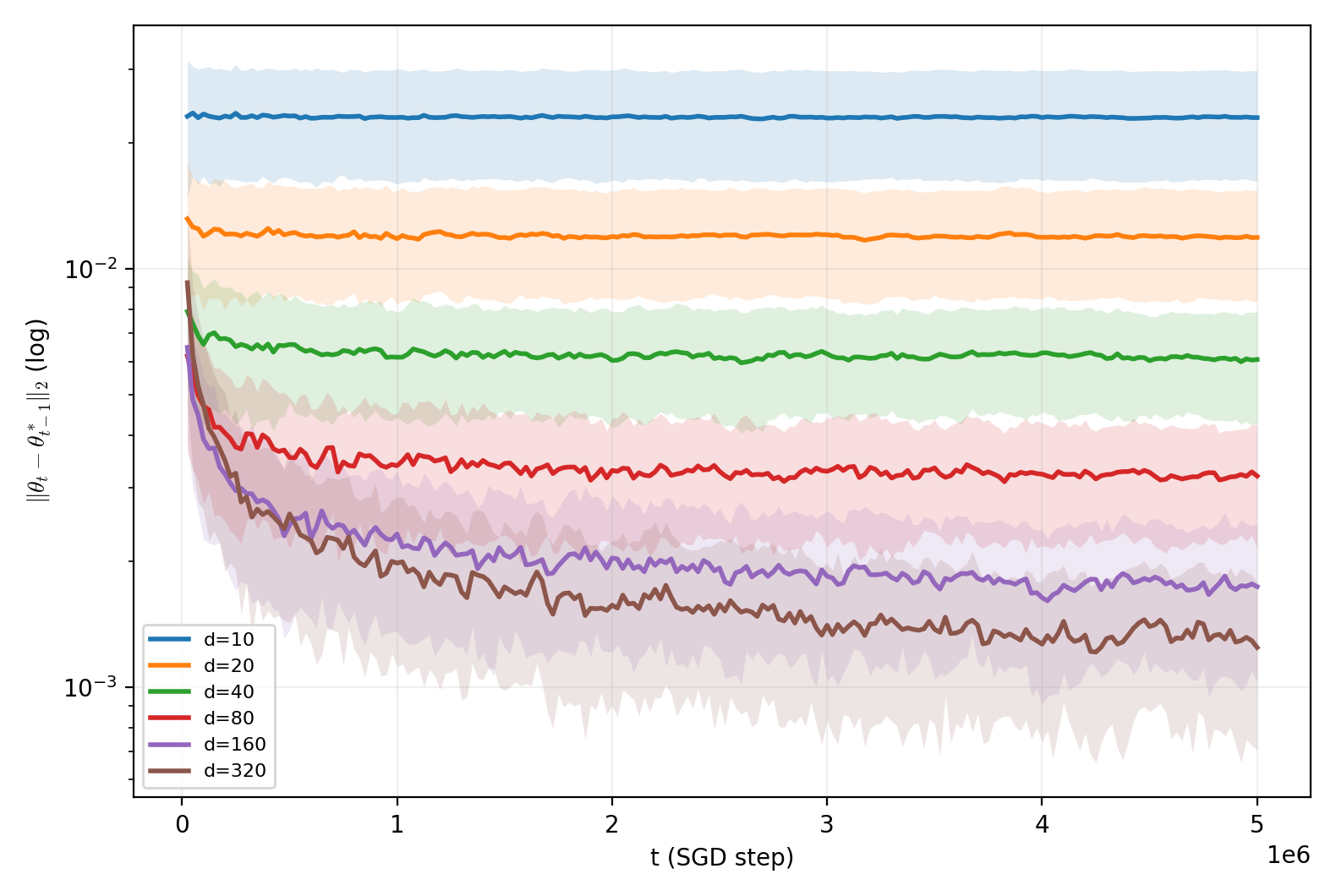}
        \caption{Performance of FSGD over different $d$}
        \label{fig: app Linear SGD d}
    \end{subfigure}
    \hfill
    \begin{subfigure}[t]{0.45\linewidth}
        \centering
        \includegraphics[width=\linewidth]{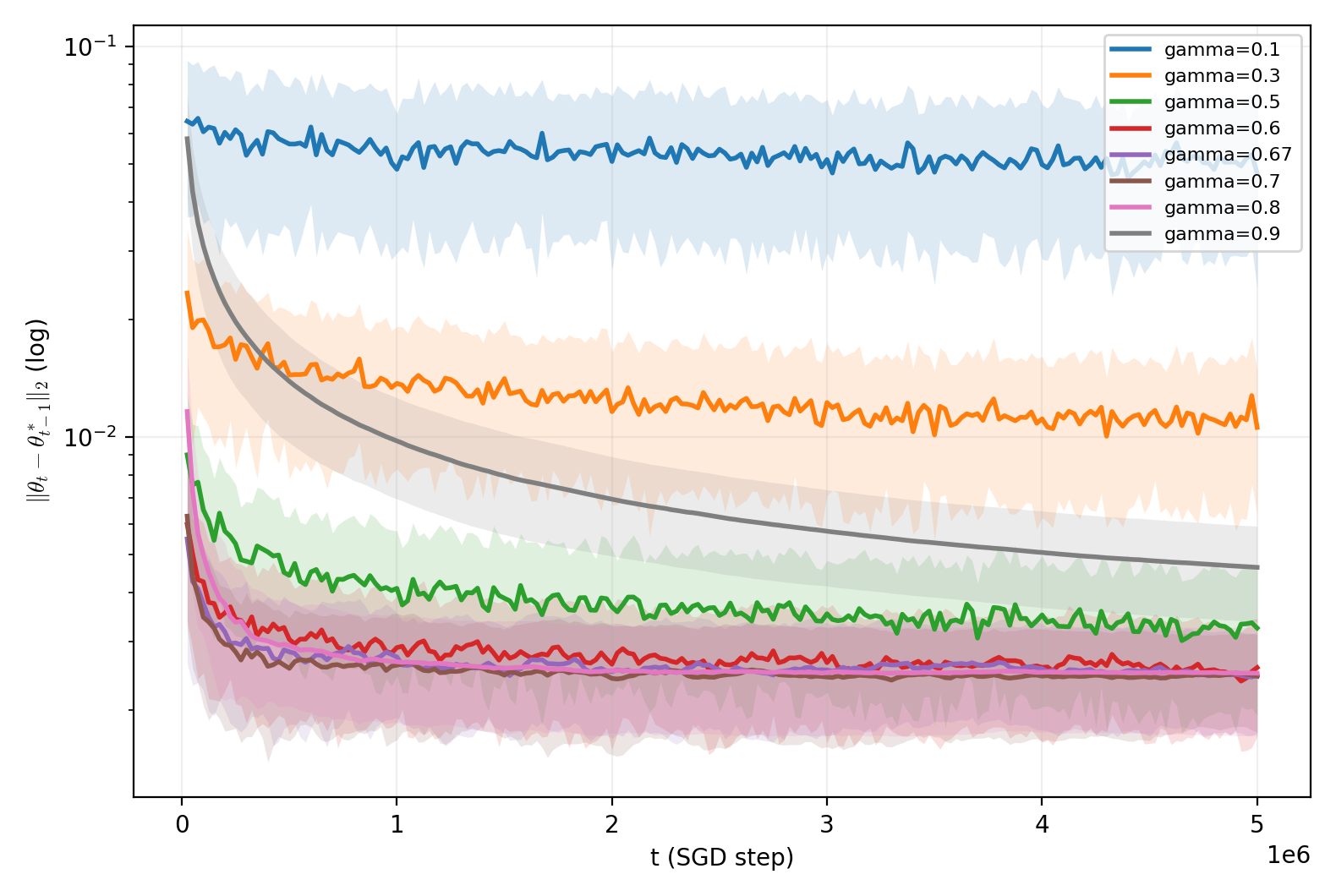}
        \caption{Performance of FSGD over different $\gamma$}
        \label{fig: app Linear SGD gamma}
    \end{subfigure}

    \caption{Performance of FSGD in linear SGD experiments.}
    \label{fig: app Linear SGD combined}
\end{figure}

\paragraph{Scaling with ambient dimension.}
In the first experiment, we fix $\gamma=0.6$ and vary $d \in \{10,20,40,80,160,320\}$. Figure~\ref{fig: app Linear SGD d} shows that the rotated parameter error $|\theta_T-\theta_{T-1}^*|_s$ decreases as $d$ grows, matching the $d^{-1/2}$ scaling predicted by the first term in Theorem~\ref{thm:Moment_Convergence_of_SGD}.

\paragraph{Effect of the SGD decay exponent.}
In the second experiment, we fix $d=100$ and vary
$\gamma \in \{0.1,0.3,0.5,0.6,0.67,0.7,0.8,0.9\}$.
Figure~\ref{fig: app Linear SGD gamma} shows that the best performance is attained near $\gamma \approx 2/3$, which agrees with Remark~\ref{rmk:sgdrate}. Small values of $\gamma$ lead to slower SGD convergence, whereas overly large values increase the effect of minimizer drift.

\subsection{Neural Network}
\label{app:nn_simulation}

We next evaluate FSGD in a nonlinear setting using two-layer ReLU neural networks. 
These experiments are adapted from Section 5.1 of \citet{fan2024factor} and are intended to assess the empirical behavior of FSGD beyond the linear theory.

\paragraph{Data-generating process.}
Under the factor model \eqref{eq:factor model}, the loading matrix $B$ has i.i.d.\ $\mathrm{Unif}[-\sqrt{3},\sqrt{3}]$ entries. 
The latent factor $f$ and idiosyncratic component $u$ are independent and have i.i.d.\ $\mathrm{Unif}[-1,1]$ entries, respectively. 
The response variable is generated as
$y=\mathcal M(f)+\varepsilon$,
where $\varepsilon \sim \mathcal N(0,0.3)$ is independent of $(f,u)$, and
$\mathcal M(f)=\sum_{j=1}^k \mathcal M_j(f_j)$.
In each repetition, the component functions $\mathcal M_j$ are randomly selected from
$\Bigl\{\cos(\pi x),\ \sin(x),\ (1-|x|)^2,\ \frac{1}{1+e^{-x}},\ 2\sqrt{|x|}-1\Bigr\}$.
Unless otherwise specified, the true latent dimension is fixed at $k=5$. 
Depending on the experiment, we vary the ambient dimension $d$ and the estimated factor dimension $\hat r$.

For Stage II of Algorithm~\ref{alg:FASGD}, we generate $n_{\mathrm{II}}=500$ labeled samples. 
For factor-based methods, we additionally use $n_{\mathrm{I}}=50$ unlabeled samples for online PCA warm-up. 
We also generate an independent validation set of size $n_{\mathrm{valid}}=150$. 
Model performance is evaluated by the test $L_2$ loss averaged over $15{,}000$ independent test samples.

\paragraph{Implementation.}
We use fully connected ReLU neural networks with depth $L=2$ and width $N=100$ for all estimators. 
All models are trained using mini-batch SGD with batch size $m=32$, initial learning rate $0.05$, and $500$ epochs. 
The SGD learning rate follows a polynomial decay schedule with exponent $\gamma=0.3$.

For FSGD, the projection matrix is estimated using Oja's algorithm. In stage I, we initialize the factor estimator with a constant learning rate of $0.005$ for $200$ Oja iterations using the warm-up samples. In Stage II, we use
$\eta_t^{\mathrm{oja}} = 0.05(50+t)^{-1}$.

\paragraph{Methods.}
We compare FSGD with several baselines under different NN structures and projection updating strategies:
\begin{itemize}
    \item \textbf{Factor-NN}: the proposed FSGD method, which updates the projection matrix online via Oja's algorithm throughout training;
    \item \textbf{Factor-NN-F0}: the same architecture as Factor-NN, but the projection matrix is frozen immediately after warm-up;
    \item \textbf{Factor-NN-F300}: the same architecture as Factor-NN, but the projection matrix is frozen after 300 epochs;
    \item \textbf{Oracle-NN}: uses the true latent factors as inputs and serves as an ideal benchmark without factor estimation error;
    \item \textbf{Vanilla-NN}: trains the neural network directly on the raw covariates $x$ without representation learning;
    \item \textbf{PCA-A-NN}: augments the raw input with the online factor projection, i.e.,
    \[
    z = \bigl[d^{-1/2}Q_{t-1}^\top x,\; x\bigr];
    \]
    \item \textbf{NN-Joint}: jointly learns the projection matrix and network weights from data, initialized from the warm-up phase;
    \item \textbf{NN-Stiefel}: jointly learns the projection matrix and network weights from the supervised loss, with the projection initialized by the Oja warm-up and constrained to have orthonormal columns;
    \item \textbf{Random Proj}: replaces Oja's online update with a random projection matrix sampled once at initialization and kept fixed thereafter;
    \item \textbf{PPCA}: periodically recomputes the projection matrix via offline PCA using the most recent $W$ mini-batches, and performs this refresh every $M$ mini-batches.
\end{itemize}

\paragraph{NN-Stiefel implementation.}
NN-Stiefel jointly optimizes the network parameters $\theta$ and the projection matrix $Q$, subject to $Q^\top Q=I_k$.
For mini-batch $B_t$, define the empirical loss
\[
\widehat L_t(\theta,Q)
=
\frac{1}{m}\sum_{i=1}^m
\ell\bigl(\theta;d^{-1/2}Q^\top x_{t,i},y_{t,i}\bigr),
\]
where $\ell$ is the squared loss used in these experiments.
Starting from the Euclidean gradient
\[
G_t=\nabla_Q\widehat L_t(\theta_{t-1},Q_{t-1}),
\]
we project it onto the tangent space of the Stiefel manifold:
\[
G_{t,\mathrm{tan}}
=
G_t-Q_{t-1}\operatorname{sym}(Q_{t-1}^\top G_t),
\qquad
\operatorname{sym}(A)=\frac{A+A^\top}{2}.
\]
The projection is then updated by a tangent step followed by QR retraction,
\[
Q_t
=
\operatorname{qf}\bigl(
Q_{t-1}-\eta_QG_{t,\mathrm{tan}}
\bigr),
\]
where $\operatorname{qf}(A)$ denotes the orthonormal factor in the thin QR decomposition of $A$, with diagonal signs chosen so that the triangular factor has nonnegative diagonal entries.
The network parameters are updated by ordinary SGD on the same mini-batch loss, with both gradients evaluated at $(\theta_{t-1},Q_{t-1})$.
The tangent projection and QR retraction cost $O(dk^2)$ per update.

The projection learning rate $\eta_Q$ is selected using validation loss from $\{0.001,0.01,0.1,1.0\}$, yielding $\eta_Q=0.1$.
The network architecture and downstream SGD settings are the same as those used for Factor-NN.

\begin{figure*}[h]
    \begin{subfigure}[b]{0.6\textwidth}
        \centering
        \includegraphics[height=1.8in, width=\linewidth]{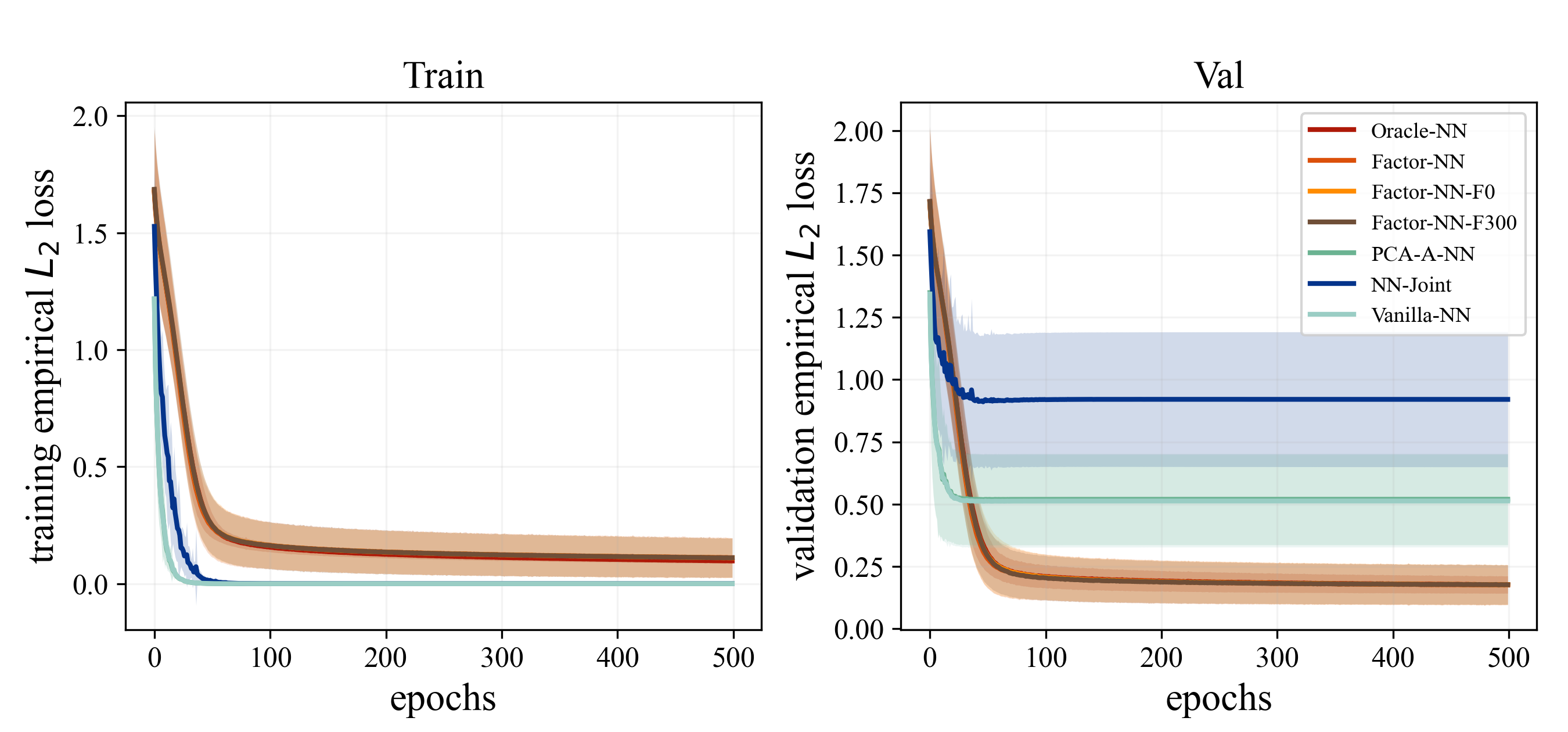}
         \vspace{-3ex}
         \caption{Train and Validation Curves}
         \label{fig: app compare with other estimators}
    \end{subfigure}%
    \begin{subfigure}[b]{0.4\textwidth}
        \centering
        \includegraphics[height=1.6in, width=\linewidth]{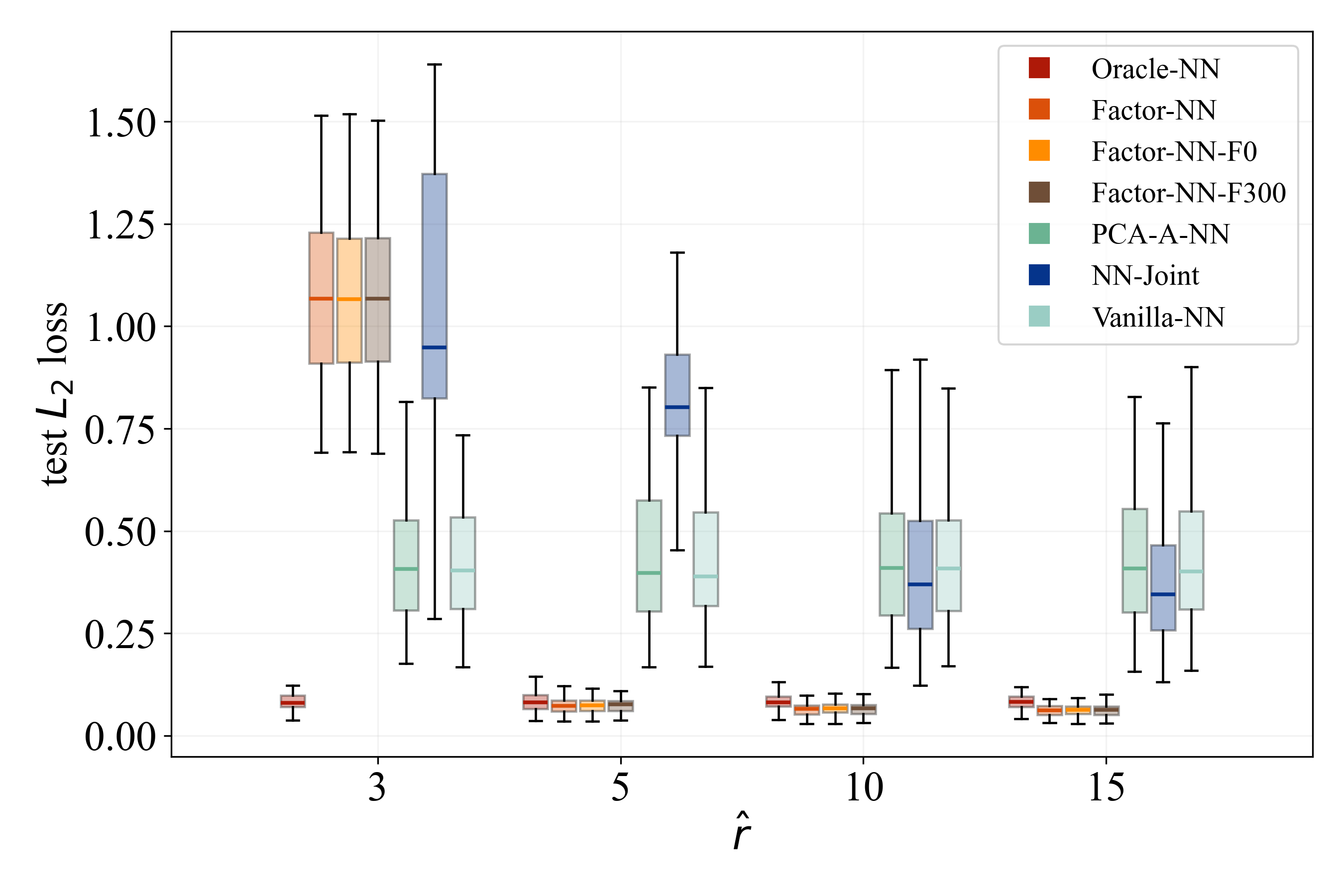}
         \vspace{-3ex}
         \caption{Performance over Different Methods}
         \label{fig: app compare with estimated rank}
    \end{subfigure}%
    \caption{Evaluation of Empirical Performance of FSGD}
    \label{fig: app simu-compare}
\end{figure*}

\paragraph{Main comparison.}
Figure~\ref{fig: app compare with other estimators} plots the training and validation curves when $\hat r=5$. 
FSGD-based methods consistently outperform all practical baselines and achieve performance close to the oracle benchmark. 
Figure~\ref{fig: app compare with estimated rank} further reports the test loss under different estimated factor dimensions $\hat r$. 
The results show that Factor-NN is robust to moderate overestimation of the factor dimension, whereas underestimation leads to clear degradation. 
The ``NN-Joint'' baseline is unstable when $\hat r$ is small, but improves when the representation is over-specified. 
Moreover, the differences among Factor-NN, Factor-NN-F0, and Factor-NN-F300 are small when $\hat r$ is sufficiently large, suggesting that minimizer drift has only a limited effect in this regime. 

\paragraph{Effect of ambient dimension.}
We also evaluate the effect of increasing ambient dimension 
$d \in \{100,200,400,600,800,1000,1500,2000,3000\}$,
while fixing the true factor dimension $k=5$ and the SGD decay exponent $\gamma=0.3$. 
The results are summarized in Table~\ref{tab: app dvary_results}. 
Across all settings, training for 500 epochs yields stable optimization, so the remaining error is primarily driven by factor estimation. 
Both PCA-A-NN and Vanilla-NN deteriorate as $d$ increases, whereas Factor-NN and its frozen-projection variants generally improve and approach the oracle benchmark.
NN-Stiefel also outperforms PCA-A-NN and Vanilla-NN across all evaluated dimensions, supporting the benefit of learning a low-dimensional representation in this setting.
However, its test loss remains consistently higher than that of Factor-NN, indicating that the advantage of Factor-NN is not solely due to correcting the scaling and conditioning issues of unconstrained joint optimization.
The comparison suggests an advantage of covariance-based representation learning in this small-sample, high-dimensional setting.
The overall improvement of Factor-NN with increasing $d$ is qualitatively consistent with our factor-estimation analysis, in which a larger ambient dimension facilitates recovery of the latent factor structure.

\begin{table}[H]
\centering
\caption{Mean empirical test $L_2$ loss under varying ambient dimension $d$. Numbers in parentheses are standard deviations over repetitions.}
\label{tab: app dvary_results}
\begin{tabular}{lccc}
\toprule
Method & $d{=}100$ & $d{=}200$ & $d{=}400$ \\
\midrule
Oracle-NN       & 0.087 (0.021) & 0.083 (0.022) & 0.082 (0.019) \\
Factor-NN       & 0.150 (0.047) & 0.105 (0.021) & 0.086 (0.018) \\
Factor-NN-F0    & 0.147 (0.045) & 0.105 (0.022) & 0.085 (0.019) \\
Factor-NN-F300  & 0.148 (0.045) & 0.105 (0.021) & 0.086 (0.019) \\
NN-Stiefel      & 0.248 (0.040) & 0.196 (0.033) & 0.143 (0.045) \\
PCA-A-NN        & 0.342 (0.060) & 0.350 (0.092) & 0.394 (0.131) \\
Vanilla-NN      & 0.347 (0.064) & 0.360 (0.106) & 0.387 (0.129) \\
\midrule
Method & $d{=}600$ & $d{=}800$ & $d{=}1000$ \\
\midrule
Oracle-NN       & 0.084 (0.023) & 0.085 (0.023) & 0.085 (0.022) \\
Factor-NN       & 0.081 (0.019) & 0.077 (0.017) & 0.086 (0.090) \\
Factor-NN-F0    & 0.080 (0.019) & 0.078 (0.017) & 0.087 (0.092) \\
Factor-NN-F300  & 0.080 (0.017) & 0.077 (0.017) & 0.086 (0.091) \\
NN-Stiefel      & 0.137 (0.060) & 0.111 (0.027) & 0.111 (0.035) \\
PCA-A-NN        & 0.406 (0.149) & 0.420 (0.171) & 0.430 (0.168) \\
Vanilla-NN      & 0.403 (0.147) & 0.417 (0.162) & 0.433 (0.171) \\
\midrule
Method & $d{=}1500$ & $d{=}2000$ & $d{=}3000$ \\
\midrule
Oracle-NN       & 0.085 (0.023) & 0.085 (0.024) & 0.084 (0.024) \\
Factor-NN       & 0.072 (0.019) & 0.071 (0.018) & 0.073 (0.019) \\
Factor-NN-F0    & 0.073 (0.018) & 0.071 (0.018) & 0.070 (0.015) \\
Factor-NN-F300  & 0.074 (0.018) & 0.072 (0.017) & 0.069 (0.015) \\
NN-Stiefel      & 0.104 (0.036) & 0.105 (0.047) & 0.097 (0.037) \\
PCA-A-NN        & 0.443 (0.179) & 0.447 (0.184) & 0.453 (0.193) \\
Vanilla-NN      & 0.444 (0.178) & 0.447 (0.182) & 0.454 (0.190) \\
\bottomrule
\end{tabular}
\end{table}

Interestingly, FSGD-based methods can outperform Oracle-NN when $d$ is large.
Our diagnostic experiments attribute this observed advantage to the interaction between factor rotation and the neural network's default first-layer initialization.
PyTorch initializes the first-layer weights independently from
$\operatorname{Unif}(-1/\sqrt{k},1/\sqrt{k})$, which is not rotationally invariant.
We compare this initialization with the rotation-invariant Gaussian distribution
$\mathcal{N}(0,1/(3k))$, which has the same entrywise variance.

We conducted 50 paired runs at $d=1000$.
Let $Q_0$ denote the projection obtained after warm-up, $R$ the corresponding orthogonal alignment such that
$d^{-1/2}Q_0^\top x \approx R^\top f$, and $W_0$ the initial first-layer weight matrix.
We compare four configurations, specified by their inputs and initial first-layer weights in Table~\ref{tab:oracle_initialization}.
The rotation-only configuration changes the factor coordinates while keeping $W_0$ fixed.
The rotation-with-matched-weights configuration also replaces $W_0$ by $W_0R$, preserving the initial predictions because
\(
(W_0R)(R^\top f)=W_0f.
\)
The estimated-factor configuration additionally includes factor-estimation error.
Within each run, all configurations share the data, mini-batch order, remaining initialization, and other training settings.

\begin{table}[ht]
\centering
\small
\setlength{\tabcolsep}{5pt}
\caption{Test MSE at $d=1000$ under uniform and Gaussian first-layer initialization, reported as mean (SD) over 50 paired runs.}
\label{tab:oracle_initialization}
\begin{tabular}{lllcc}
\toprule
Configuration & Input & Weights & Uniform & Gaussian \\
\midrule
Oracle
& $f$ & $W_0$
& 0.0850 (0.0220) & 0.0662 (0.0146) \\
Rotation only
& $R^\top f$ & $W_0$
& 0.0670 (0.0156) & 0.0684 (0.0154) \\
Rotation + matched weights
& $R^\top f$ & $W_0R$
& 0.0850 (0.0221) & 0.0662 (0.0146) \\
Estimated factors
& $d^{-1/2}Q_0^\top x$ & $W_0$
& 0.0742 (0.0174) & 0.0754 (0.0166) \\
\bottomrule
\end{tabular}
\end{table}

Matching the first-layer weights to the factor rotation reproduces the oracle performance under both initializations.
Under Gaussian initialization, the difference between the rotation-only and oracle configurations is not statistically significant
(paired 95\% confidence interval: $[-0.00018,0.00447]$), while the estimated-factor configuration performs worse.
These comparisons support an explanation based on the interaction between factor coordinates and initialization, rather than a benefit from factor-estimation error.

\paragraph{Time-memory trade-off.}
To further quantify the computational cost of different online learning schemes, we compare the per-step runtime, memory usage, and test loss at $d=4000$, averaged over 50 repetitions.
Both the per-step runtime and memory usage include the additional overhead from Oja updates.

We summarize the main findings from Table~\ref{tab: app nn_time_memory_full}. 
First, FSGD-based methods achieve the best test loss among the neural network-based methods, including PCA-A-NN and NN-Joint, and substantially outperform Random Projection and the periodic online projection update scheme PPCA. 
In terms of efficiency, FSGD is considerably cheaper than Vanilla SGD in both per-step runtime and memory usage (see comparison between Factor-NN and Vanilla-NN), although it is more expensive than NN-Joint and Oracle-NN due to the additional Oja update. 
However, after freezing the learned subspace, the Factor-NN-F300 achieves a computational cost comparable to the NN-Joint and Oracle-NN baselines while maintaining the lowest test loss.

In contrast, PPCA can be computationally expensive when the projection is recomputed frequently or when the PCA window size $W$ is large, yet it still suffers from high prediction error. 
One possible explanation is that each periodic SVD recomputation induces an abrupt change in the learned representation, making it difficult for the downstream neural network to readapt, especially under a decaying learning rate schedule. 
PPCA also incurs a larger memory cost because it needs to store the sliding window of mini-batch samples for PCA recomputation.

\begin{table}[ht]
\centering
\caption{Neural-network time-memory trade-off at $d=4000$, averaged over 50 repetitions. Per-step time includes Oja/SVD overhead.}
\label{tab: app nn_time_memory_full}
\begin{tabular}{lccc}
\toprule
Method & SGD Per-step (ms) & SGD Memory (MB) & Test Loss \\
\midrule
Factor-NN           & 1.08 & 1.15 & 0.0687 \\
Factor-NN-F0        & 0.48 & 0.62 & 0.0694 \\
Factor-NN-F300      & 0.50 & 1.15 & \textbf{0.0677} \\
PCA-A-NN        & 2.61 & 4.21 & 0.4814 \\
NN-Joint          & 0.57 & 0.23 & 0.6233 \\
Oracle-NN         & \textbf{0.39} & \textbf{0.08} & 0.0876 \\
Vanilla-NN    & 1.79 & 3.13 & 0.4746 \\
Random Proj    & 0.48 & 0.62 & 1.6147 \\
PPCA $(M=20,W=2)$   & 1.17 & 4.33  & 1.0936 \\
PPCA $(M=20,W=5)$   & 3.98 & 10.35 & 1.0363 \\
PPCA $(M=20,W=10)$  & 8.77 & 20.70 & 1.0588 \\
PPCA $(M=100,W=2)$  & 0.62 & 4.33  & 1.2450 \\
PPCA $(M=100,W=5)$  & 1.18 & 10.35 & 1.2811 \\
PPCA $(M=100,W=10)$ & 2.14 & 20.70 & 1.1816 \\
PPCA $(M=500,W=2)$  & 0.51 & 4.33  & 1.5810 \\
PPCA $(M=500,W=5)$  & 0.62 & 10.35 & 1.6352 \\
PPCA $(M=500,W=10)$ & 0.81 & 20.70 & 1.6684 \\
\bottomrule
\end{tabular}
\end{table}

\paragraph{Warm-up sensitivity.}
We consider a deliberately challenging setting with $d=1000$ and $n_{\rm warm}=5$, averaged over $50$ repetitions. 
The oracle test MSE is $0.087$. 
Table~\ref{tab: app warmup_sensitivity_f0} reports the results for Factor-NN-F0 where the projection is frozen immediately after warm-up, while Table~\ref{tab: app warmup_sensitivity_cont} reports the results for Factor-NN where Oja updates continue throughout Stage II.

Overall, FSGD (Factor-NN) is insensitive to the warm-up hyperparameters as long as  Oja updates are continued in Stage II. 
The reason is that online representation learning can correct an inaccurate subspace estimate obtained from a poor warm-up phase. 
In Table~\ref{tab: app warmup_sensitivity_f0} where the projection is frozen immediately after warm-up (Factor-NN-F0), performance degrades severely with MSE ranging from 0.53 to 1.49. 
In contrast, Factor-NN with continued Oja updates achieves MSE $0.074$-$0.097$ even at $T_0=1$, demonstrating that Stage II online updates can correct a poor initialization without relying on warm-up.
These results indicate that a long or carefully tuned warm-up phase is not necessary in practice.

\begin{table}[H]
\centering
\caption{Warm-up sensitivity for Factor-NN-F0 at $d=1000$ with $n_{\rm warm}=5$ (50 repetitions). Rows correspond to warm-up length $T_0$ and columns correspond to the warm-up learning rate. Entries are test MSE.}
\label{tab: app warmup_sensitivity_f0}
\begin{tabular}{lccccc}
\toprule
$T_0$ & 0.0005 & 0.001 & 0.005 & 0.01 & 0.05 \\
\midrule
1  & 1.487 & 1.455 & 1.423 & 1.426 & 1.435 \\
5  & 1.362 & 1.318 & 1.243 & 1.233 & 1.222 \\
10 & 1.231 & 1.140 & 1.027 & 1.014 & 0.999 \\
20 & 1.055 & 0.952 & 0.831 & 0.812 & 0.800 \\
50 & 0.791 & 0.686 & 0.568 & 0.548 & 0.532 \\
\bottomrule
\end{tabular}
\end{table}

\begin{table}[H]
\centering
\caption{Warm-up sensitivity for Factor-NN at $d=1000$ with $n_{\rm warm}=5$ (50 repetitions). Rows correspond to warm-up length $T_0$ and columns correspond to the warm-up learning rate. Entries are test MSE.}
\label{tab: app warmup_sensitivity_cont}
\begin{tabular}{lccccc}
\toprule
$T_0$ & 0.0005 & 0.001 & 0.005 & 0.01 & 0.05 \\
\midrule
1  & 0.083 & 0.084 & 0.097 & 0.074 & 0.076 \\
5  & 0.077 & 0.076 & 0.075 & 0.079 & 0.075 \\
10 & 0.078 & 0.075 & 0.077 & 0.077 & 0.083 \\
20 & 0.085 & 0.076 & 0.075 & 0.075 & 0.075 \\
50 & 0.088 & 0.076 & 0.076 & 0.081 & 0.075 \\
\bottomrule
\end{tabular}
\end{table}

\paragraph{Sensitivity to Stage-II Oja hyperparameters.}
Finally, we study the sensitivity to the Stage-II Oja step-size schedule
$\eta_t^{\mathrm{oja,II}} = c_{\mathrm{oja}}(C_{\mathrm{oja}}+t)^{-1}.$
We vary $c_{\mathrm{oja}} \in \{0.005,0.01,0.05,0.1,0.5\}$ and
$C_{\mathrm{oja}} \in \{500,100,50,10,5\}$ at $d=1000$, using 50 repetitions and no alignment.

Tables~\ref{tab: app stage2_oja_fsgd} and \ref{tab: app stage2_oja_fsgd_f300} report the results for Factor-NN and Factor-NN-F300, respectively. 
Across all configurations, the performance remains remarkably stable. 
For Factor-NN, the test loss ranges roughly from $0.0719$ to $0.0954$, and for Factor-NN-F300 from $0.0706$ to $0.0844$, both close to the oracle benchmark. 
Thus, the method is not particularly sensitive to the choice of Stage-II Oja hyperparameters, which further supports the practical robustness of the online subspace update.

\begin{table}[H]
\centering
\caption{Sensitivity to Stage-II Oja hyperparameters for Factor-NN at $d=1000$ (50 repetitions). Entries are test MSE under $\eta_t^{\mathrm{oja,II}} = c_{\mathrm{oja}}(C_{\mathrm{oja}}+t)^{-1}$.}
\label{tab: app stage2_oja_fsgd}
\begin{tabular}{lccccc}
\toprule
$C_{\mathrm{oja}} \backslash c_{\mathrm{oja}}$ & 0.005 & 0.01 & 0.05 & 0.1 & 0.5 \\
\midrule
500 & 0.0799 & 0.0902 & 0.0836 & 0.0771 & 0.0804 \\
100 & 0.0858 & 0.0954 & 0.0752 & 0.0768 & 0.0801 \\
50  & 0.0914 & 0.0910 & 0.0735 & 0.0739 & 0.0803 \\
10  & 0.0910 & 0.0820 & 0.0722 & 0.0722 & 0.0806 \\
5   & 0.0778 & 0.0759 & 0.0734 & 0.0719 & 0.0806 \\
\bottomrule
\end{tabular}
\end{table}

\begin{table}[H]
\centering
\caption{Sensitivity to Stage-II Oja hyperparameters for Factor-NN-F300 at $d=1000$ (50 repetitions). Entries are test MSE under $\eta_t^{\mathrm{oja,II}} = c_{\mathrm{oja}}(C_{\mathrm{oja}}+t)^{-1}$.}
\label{tab: app stage2_oja_fsgd_f300}
\begin{tabular}{lccccc}
\toprule
$C_{\mathrm{oja}} \backslash c_{\mathrm{oja}}$ & 0.005 & 0.01 & 0.05 & 0.1 & 0.5 \\
\midrule
500 & 0.0844 & 0.0843 & 0.0715 & 0.0716 & 0.0718 \\
100 & 0.0844 & 0.0730 & 0.0714 & 0.0724 & 0.0718 \\
50  & 0.0770 & 0.0716 & 0.0720 & 0.0718 & 0.0710 \\
10  & 0.0722 & 0.0710 & 0.0721 & 0.0709 & 0.0714 \\
5   & 0.0715 & 0.0706 & 0.0716 & 0.0710 & 0.0714 \\
\bottomrule
\end{tabular}
\end{table}

\paragraph{Benefits of Stage-II representation updates.}
The similar performance of Factor-NN and Factor-NN-F0 in Table~\ref{tab: app dvary_results} can be explained by the informative standard warm-up, which uses 50 samples and 200 Oja iterations.
The benefit of continued representation learning becomes clearer when the initial projection is less accurate.
With only five warm-up samples, Factor-NN-F0 retains the inaccurate projection and incurs test MSEs of $0.532$--$1.487$ (Table~\ref{tab: app warmup_sensitivity_f0}), whereas Factor-NN achieves $0.074$--$0.097$ by continuing to refine it (Table~\ref{tab: app warmup_sensitivity_cont}).
Thus, freezing can be effective when warm-up already provides an accurate projection, while Stage-II updates reduce reliance on the quality of the initial representation.

\paragraph{Summary.}
Taken together, these experiments support three conclusions. 
First, factor-based learning substantially improves predictive performance over training directly in the ambient space. 
Second, the online update in FSGD achieves an attractive statistical-computational trade-off, combining low test loss with moderate runtime and memory. 
Third, the method is robust to both warm-up design and Stage-II Oja hyperparameters, indicating that its empirical performance does not rely on delicate tuning.

\section{Real-Data Experiment}
\label{app:real_data_details}

\subsection{MNIST}
\label{app:mnist}

We evaluate FSGD on MNIST to assess its empirical performance in image classification.

\paragraph{Data and preprocessing.}
For each of 30 random splits, the official training set is partitioned into 320 unlabeled warm-up images, 5,000 validation images, and 54,680 supervised training images.
The official test set of 10,000 images remains fixed.
Each split is evaluated using five training seeds.
Images are flattened into $d=784$ dimensions, divided by 255, and centered using the pixelwise mean estimated from the warm-up images.

\paragraph{Models and training.}
The selected projection dimension is $\hat{k}=64$.
Projected methods use a fully connected ReLU network with architecture $64\to128\to10$, while Vanilla-NN uses $784\to128\to10$.
All methods minimize cross-entropy loss using plain mini-batch SGD without momentum, with batch size 32 and weight decay $10^{-4}$.
Training makes one pass over the supervised samples, giving 1,709 updates.
The classifier learning rate is
\[
\eta_t=\eta_0t^{-0.3},\qquad
\eta_0\in\{0.75,1.00,1.25\},
\]
where $t$ starts at 1.

Stage I makes one pass over the warm-up images using Oja's algorithm with batch size 8 and constant step size $0.1$, giving 40 updates.
During Stage II, Factor-NN updates the projection after each supervised mini-batch using
\[
\eta_u^{\mathrm{oja}}=\frac{5}{500+u},
\qquad u=0,1,\ldots.
\]
After each Stage-I or Stage-II Oja update, the basis obtained by thin QR is aligned to the preceding basis using Procrustes alignment.

\paragraph{Comparison methods.}
We compare five methods:
Factor-NN continues the Oja updates throughout supervised training;
Factor-NN-F0 freezes the projection immediately after Stage I;
NN-Stiefel jointly learns the projection and classifier from the supervised loss;
Vanilla-NN uses the raw covariates;
and Random Projection uses a fixed random orthogonal projection.
In these image experiments, NN-Stiefel updates the projection using a Euclidean gradient step with learning rate $0.01$, followed by thin QR retraction to maintain orthonormal columns.
Factor-NN, Factor-NN-F0, and NN-Stiefel are initialized with the same projection matrix obtained from the Oja warm-up.
Within each split and training seed, projected methods share the downstream network initialization, and all methods use the same supervised mini-batch order.

\paragraph{Evaluation and statistical analysis.}
For both image datasets, we report test accuracy at the end of the single supervised pass.
We first average the five training seeds within each split and then report the mean and sample standard deviation across the 30 split means.
The full learning-rate comparisons are reported below; $\eta_0=1$ serves as a common reference setting for the two datasets.
At this setting, we compare Factor-NN with each of the four baselines using paired differences between split means.
We use two-sided $t$-tests with 29 degrees of freedom and apply Holm correction jointly to the eight comparisons across the two datasets.
The reported 95\% confidence intervals are unadjusted paired $t$ intervals.
These comparisons quantify variability under the repeated-split evaluation with the official test set fixed.

\paragraph{Predictive performance.}
Table~\ref{tab:app_mnist_accuracy} reports the complete learning-rate comparison.
Factor-NN achieves the highest mean test accuracy at every evaluated learning rate.
At $\eta_0=1$, its mean accuracy is $96.2426\%$, compared with $95.7078\%$ for Factor-NN-F0, $95.9590\%$ for NN-Stiefel, and $95.8245\%$ for Vanilla-NN.
The paired gains and confidence intervals are reported in Table~\ref{tab:app_image_paired}.
All four comparisons remain significant after Holm correction. 
These results provide empirical evidence that FSGD is effective on real image data, without imposing an explicit linear factor model on the data-generating process.

\begin{table}[ht]
\centering
\small
\setlength{\tabcolsep}{5pt}
\caption{MNIST test accuracy (\%) after one pass, reported as mean (SD) across 30 split means.}
\label{tab:app_mnist_accuracy}
\begin{tabular}{lccc}
\toprule
Method & $\eta_0=0.75$ & $\eta_0=1.00$ & $\eta_0=1.25$ \\
\midrule
Factor-NN
& \textbf{95.8541 (0.0593)}
& \textbf{96.2426 (0.0710)}
& \textbf{96.4695 (0.0780)} \\
Factor-NN-F0
& 95.3087 (0.0954)
& 95.7078 (0.0998)
& 95.9494 (0.0929) \\
NN-Stiefel
& 95.5961 (0.0840)
& 95.9590 (0.0880)
& 96.1737 (0.0952) \\
Vanilla-NN
& 95.2636 (0.0493)
& 95.8245 (0.0533)
& 96.1847 (0.0628) \\
Random Projection
& 92.9389 (0.0842)
& 93.4412 (0.0992)
& 93.7388 (0.0960) \\
\bottomrule
\end{tabular}
\end{table}

\paragraph{Effect of freezing time.}
To examine the benefit of continued representation learning, we additionally freeze the projection after the first
$\lceil\alpha T\rceil$ supervised updates, where
$\alpha\in\{1/4,1/2,3/4\}$ and $T=1709$ is the total number of supervised updates.
The corresponding freezing times are 428, 855, and 1,282 updates.
The downstream classifier continues training after the projection is frozen, and all other settings remain unchanged.

Table~\ref{tab:app_mnist_freezing} shows that mean test accuracy increases with the coupling duration at all three learning rates.
Freezing immediately after warm-up incurs the largest loss, while freezing after $3T/4$ gives accuracy very close to full coupling.
This pattern illustrates the trade-off reflected in Corollary~\ref{cor:Moment_Convergence_of_SGD-freeze_Oja}: freezing eliminates subsequent minimizer drift but retains the representation error at the freezing time.
The empirical results show that freezing too early can reduce predictive performance.
Full coupling also avoids the need to choose a freezing time in advance.

\begin{table}[ht]
\centering
\small
\setlength{\tabcolsep}{5pt}
\caption{MNIST test accuracy (\%) under different freezing times, reported as mean (SD) across 30 split means.}
\label{tab:app_mnist_freezing}
\begin{tabular}{lccc}
\toprule
Method & $\eta_0=0.75$ & $\eta_0=1.00$ & $\eta_0=1.25$ \\
\midrule
Factor-NN-F0
& 95.3087 (0.0954)
& 95.7078 (0.0998)
& 95.9494 (0.0929) \\
Factor-NN-$T/4$
& 95.7549 (0.0672)
& 96.1515 (0.0766)
& 96.3741 (0.0891) \\
Factor-NN-$T/2$
& 95.8251 (0.0667)
& 96.2125 (0.0749)
& 96.4371 (0.0831) \\
Factor-NN-$3T/4$
& 95.8461 (0.0609)
& 96.2365 (0.0760)
& 96.4581 (0.0803) \\
Factor-NN
& \textbf{95.8541 (0.0593)}
& \textbf{96.2426 (0.0710)}
& \textbf{96.4695 (0.0780)} \\
\bottomrule
\end{tabular}
\end{table}

\subsection{CIFAR-10}
\label{app:cifar10}

We further evaluate the same five methods on CIFAR-10 using a one-pass classification protocol.

\paragraph{Data and preprocessing.}
For each of 30 random splits, the official training set is partitioned into 1,000 unlabeled warm-up images, 5,000 validation images, and 44,000 supervised training images.
Each split is evaluated using five training seeds, and the official test set of 10,000 images remains fixed.
Images are flattened into $d=3072$ dimensions.
All pixels are divided by 255 and centered using pixelwise means estimated from the warm-up images.

\paragraph{Models and training.}
The selected projection dimension is $\hat{k}=128$.
Projected methods use a fully connected ReLU network with architecture $128\to256\to256\to10$, while Vanilla-NN uses $3072\to256\to256\to10$.
All methods minimize cross-entropy loss using plain SGD without momentum, with batch size 128 and weight decay $10^{-4}$.
One pass over the supervised samples gives 344 updates.
The classifier learning rate is
\[
\eta_t=\eta_0t^{-0.3},\qquad
\eta_0\in\{0.3,0.5,1.0,1.5\},
\]
where $t$ starts at 1.

Stage I makes one pass over the warm-up images using Oja's algorithm with batch size 32 and constant step size $0.1$, giving 32 updates.
During Stage II, Factor-NN updates the projection after each supervised mini-batch using
\[
\eta_u^{\mathrm{oja}}
=\frac{5}{500+u},
\qquad u=0,1,\ldots.
\]
The Oja alignment procedure and comparison methods follow Section~\ref{app:mnist}.
NN-Stiefel again uses a projection learning rate of $0.01$ and thin QR retraction.
Factor-NN, Factor-NN-F0, and NN-Stiefel are initialized with the same projection matrix obtained from the Oja warm-up.
Evaluation and statistical analysis also follow Section~\ref{app:mnist}.

\paragraph{Predictive performance.}
Table~\ref{tab:app_cifar10_accuracy} reports the complete learning-rate comparison.
Factor-NN achieves the highest mean test accuracy at every evaluated learning rate.
At the common setting $\eta_0=1$, it improves over Factor-NN-F0, NN-Stiefel, Vanilla-NN, and Random Projection by $0.600$, $0.471$, $0.547$, and $2.532$ percentage points, respectively.
All four comparisons remain significant after Holm correction; the paired confidence intervals are reported in Table~\ref{tab:app_image_paired}.
Together with the MNIST results, these comparisons provide empirical evidence for the predictive benefit of continued representation learning in one-pass classification.

\begin{table}[ht]
\centering
\small
\setlength{\tabcolsep}{4pt}
\caption{CIFAR-10 test accuracy (\%) after one pass, reported as mean (SD) across 30 split means.}
\label{tab:app_cifar10_accuracy}
\begin{tabular}{lcccc}
\toprule
Method & $\eta_0=0.3$ & $\eta_0=0.5$ & $\eta_0=1.0$ & $\eta_0=1.5$ \\
\midrule
Factor-NN
& \textbf{40.2748 (0.1450)}
& \textbf{42.9919 (0.2033)}
& \textbf{45.2353 (0.3625)}
& \textbf{45.6488 (0.4436)} \\
Factor-NN-F0
& 39.8564 (0.1608)
& 42.4607 (0.1958)
& 44.6351 (0.3429)
& 45.0264 (0.4238) \\
NN-Stiefel
& 40.0243 (0.1741)
& 42.6857 (0.2117)
& 44.7638 (0.3831)
& 45.1144 (0.4161) \\
Vanilla-NN
& 37.9615 (0.1574)
& 41.9580 (0.2046)
& 44.6887 (0.5061)
& 45.2200 (0.5833) \\
Random Projection
& 37.6999 (0.1870)
& 40.4037 (0.1796)
& 42.7037 (0.2985)
& 43.2419 (0.3561) \\
\bottomrule
\end{tabular}
\end{table}

\begin{table}[ht]
\centering
\small
\setlength{\tabcolsep}{5pt}
\caption{Paired test-accuracy gains of Factor-NN at $\eta_0=1$, in percentage points.}
\label{tab:app_image_paired}
\begin{tabular}{llcc}
\toprule
Dataset & Baseline & Mean gain & 95\% CI \\
\midrule
MNIST & Factor-NN-F0 & 0.535 & [0.509, 0.561] \\
& NN-Stiefel & 0.284 & [0.264, 0.303] \\
& Vanilla-NN & 0.418 & [0.397, 0.440] \\
& Random Projection & 2.801 & [2.763, 2.840] \\
\midrule
CIFAR-10 & Factor-NN-F0 & 0.600 & [0.562, 0.638] \\
& NN-Stiefel & 0.471 & [0.391, 0.552] \\
& Vanilla-NN & 0.547 & [0.433, 0.660] \\
& Random Projection & 2.532 & [2.457, 2.606] \\
\bottomrule
\end{tabular}
\end{table}

\subsection{NCEP/NCAR Reanalysis}
\label{app:ncep}
We further evaluate FSGD on the NCEP/NCAR Reanalysis 1 dataset \citep{kalnay1996ncep}, a global atmospheric reanalysis dataset containing monthly Z500 anomalies on a $2.5^\circ \times 2.5^\circ$ grid, publicly available from NOAA Physical Sciences Laboratory.
The dataset consists of $T=887$ monthly observations from 1950 to 2023 over $d=10{,}511$ spatial grid points. 
Our prediction task is to forecast the next-month anomaly at a target location using the current global field. 
This setting is well suited to our framework because large-scale climate variability induces a strong low-dimensional factor structure in the high-dimensional covariates. 
The first 120 months are used for warm-up, the next 240 months for online learning and hyperparameter selection, and the held-out period 1980--2023 (527 months) is used for test evaluation.

\paragraph{Linear prediction model.}
To isolate the effect of representation learning, all methods use the same linear predictor. 
Specifically, each method predicts the response using either
$y_t = x_t^\top \theta$ or 
$y_t = f_t^\top \theta, \quad f_t = Q^\top x_t$,
depending on whether the method operates in the original ambient space or in a learned low-dimensional factor space.

\paragraph{Methods.}
We compare the following estimators.

\textbf{FSGD.}
FSGD first estimates a projection matrix $Q \in \mathbb{R}^{d\times k}$ during warm-up via streaming Oja updates, and then initializes the regression coefficient by ordinary least squares on the warm-up period:
\[
\hat{\theta}_{\mathrm{warm}}
=
(F_{\mathrm{warm}}^\top F_{\mathrm{warm}})^{-1}F_{\mathrm{warm}}^\top y_{\mathrm{warm}},
\qquad
F_{\mathrm{warm}} = X_{\mathrm{warm}} Q.
\]
During online learning, FSGD jointly updates the projection matrix $Q$ using Oja's algorithm and the regression coefficient $\theta$ using SGD in the $k$-dimensional factor space. 
We search over
\[
k \in \{5,10,20\},\quad
c_{\mathrm{sgd}} \in \{10^{-4},\ldots,1\},\quad
\gamma_{\mathrm{sgd}} \in \{0.1,0.3,0.67,0.8\},
\]
\[
c_{\mathrm{oja}} \in \{10^{-5},\ldots,10^{-2}\},\quad
\gamma_{\mathrm{oja}} = 1,\quad
c_{\mathrm{warm}} \in \{0.5,1.0\}.
\]
Its storage cost is $O(dk)$, which is about $0.40$ MB when $k=5$.

\textbf{Periodic PCA + SGD (PPCA).}
PPCA first computes an offline PCA projection matrix on the warm-up period and initializes $\theta$ by OLS in the projected space. 
During online learning, it updates $\theta$ by SGD, and every $M$ months recomputes the projection matrix by offline PCA on the most recent window of size $W$. 
After each PCA refresh, the regression coefficient is rotated accordingly to align with the new subspace. 
We search over
\[
k \in \{5,10,20\},\quad
M \in \{3,6,12,24\},\quad
W \in \{30,60,120\},
\]
together with SGD hyperparameters
\[
c_{\mathrm{sgd}} \in \{10^{-5},\ldots,10^{-3}\},\qquad
\gamma_{\mathrm{sgd}} \in \{0.1,0.3,0.67,0.8\}.
\]
Its storage cost is $O(Wd)$, since the window data must be retained for repeated SVD computations; for $W=120$, this is approximately $9.62$ MB when $k=5$.

\textbf{Vanilla SGD.}
Vanilla SGD fits a linear model directly in the original $d=10{,}511$ dimensional space. 
It is initialized using the minimum-norm warm-up OLS solution
\[
\hat{\theta}_{\mathrm{warm}} = (X_{\mathrm{warm}}^\top X_{\mathrm{warm}})^+ X_{\mathrm{warm}}^\top y_{\mathrm{warm}},
\]
and is then updated by SGD online. 
We search over
\[
c_{\mathrm{sgd}} \in \{10^{-7},\ldots,10^{-4}\},\qquad
\gamma_{\mathrm{sgd}} \in \{0.1,0.3,0.67,0.8\}.
\]
Its storage cost is $O(d)$, about $0.08$ MB.

\textbf{Random Projection + SGD (RP).}
RP replaces the learned projection matrix with a fixed random orthogonal matrix
$R \in \mathbb{R}^{d\times k}$,
sampled once at initialization and never updated thereafter. 
The regression coefficient is initialized by OLS on the projected warm-up data and then updated by SGD online in the $k$-dimensional projected space. 
We search over
\[
k \in \{5,10,20\},\qquad
c_{\mathrm{sgd}} \in \{10^{-5},\ldots,10^{-3}\},\qquad
\gamma_{\mathrm{sgd}} \in \{0.1,0.3,0.67,0.8\}.
\]
Its storage cost is also $O(dk)$, about $0.40$ MB when $k=5$.

\textbf{Persistence.}
The persistence baseline predicts
$y_{t+1}=y_t.$

\textbf{Prevailing Mean.}
As an additional naive baseline, we also consider the prevailing mean predictor
$y_{t+1} = \frac{1}{t}\sum_{s=1}^t y_s.$

\paragraph{Hyperparameter selection.}
All hyperparameters are selected by maximizing $R^2$ on the validation period 1960--1979. 
The held-out test period 1980--2023 is never used for model selection. 

\paragraph{Predictive performance.}
Table~\ref{tab: app ncep_results_full} reports the test $R^2$ on the held-out period 1980--2023 for four response locations, which were randomly selected from those satisfying a common stability criterion: all compared methods achieved test $R^2>-1$.
This criterion was applied uniformly to exclude severe numerical or predictive failures.
FSGD achieves the best performance in three of the four selected locations, outperforming both Vanilla SGD and Random Projection in all four cases. 
Compared with PPCA, FSGD is slightly better on three locations, while PPCA performs better on one low-persistence location. 
These results suggest that a low-dimensional factor structure is useful in the NCEP task. FSGD further provides an online strategy  to track this structure, achieving performance comparable to or better than PPCA while avoiding repeated offline SVD and large window storage.
The prevailing mean baseline performs poorly across all locations.

\begin{table}[ht]
\centering
\small
\setlength{\tabcolsep}{5pt}
\caption{Test $R^2$ on the held-out period 1980--2023 for four selected response variables.}
\begin{tabular}{lcccccc}
\toprule
Response & FSGD & PPCA & Persistence & Vanilla SGD & Random Proj & Prev.\ Mean \\
\midrule
P06\_TropSPac4 & \textbf{0.641} & 0.631 & 0.601 & 0.549 & 0.542 & -0.965 \\
P05\_TropNPac2 & \textbf{0.568} & 0.553 & 0.500 & 0.459 & 0.487 & -0.774 \\
P05\_TropSPac8 & \textbf{0.592} & 0.576 & 0.502 & 0.449 & 0.375 & -0.655 \\
P01\_SubNAtl10 & 0.379 & \textbf{0.418} & 0.104 & 0.175 & 0.329 & -0.581 \\
\bottomrule
\end{tabular}
\label{tab: app ncep_results_full}
\end{table}

\paragraph{Computational cost.}
We also compare total runtime, per-step runtime, and storage cost on the NCEP task. 
The results are summarized in Table~\ref{tab: app ncep_cost}.
For FSGD, runtime increases with the projection dimension $k$ due to the additional online Oja updates. 
Vanilla SGD and Random Projection have lower runtimes, whereas FSGD achieves higher test $R^2$ at all four selected locations in the reported run. 
PPCA ($k=5$) can become significantly more expensive when projection updates are frequent or the window size is large, since it repeatedly performs SVD on high-dimensional data and must store the window observations.

We note that runtime measurements depend on the hardware, BLAS/LAPACK backend, and Python/numpy environment; therefore, the reported computational times should be interpreted as machine-dependent wall-clock benchmarks rather than hardware-independent quantities.


\begin{table}[H]
\centering
\small
\setlength{\tabcolsep}{5pt}
\caption{Computational cost on the NCEP real-data task.}
\begin{tabular}{lccc}
\toprule
Method & Total Time (s) & Per-step (ms) & Storage (MB) \\
\midrule
FSGD $(k=5)$  & \textbf{0.61} & \textbf{0.79} & \textbf{0.40} \\
FSGD $(k=10)$ & 1.34 & 1.75 & 0.80 \\
FSGD $(k=15)$ & 2.18 & 2.84 & 1.20 \\
\midrule
PPCA (range over $M\in\{3,6,12,24\}$, $W\in\{30,60,120\}$) & 0.30--11.11 & 0.39--14.48 & 2.41--9.62 \\
PPCA $(M=3,W=120)$   & 11.11 & 14.48 & 9.62 \\
PPCA $(M=12,W=120)$  & 2.78 & 3.63 & 9.62 \\
PPCA $(M=24,W=30)$   & 0.30 & 0.39 & 2.41 \\
\midrule
Vanilla SGD         & 0.04 & 0.05 & 0.08 \\
Random Proj $(k=5)$ & 0.02 & 0.03 & 0.40 \\
Random Proj $(k=10)$& 0.04 & 0.05 & 0.80 \\
Persistence         & 0.00 & 0.00 & 0.00 \\
\bottomrule
\end{tabular}
\label{tab: app ncep_cost}
\end{table}

\paragraph{Summary.}

The NCEP results illustrate the usefulness of factor-based representations for the selected forecasting tasks, while the relative predictive performance of FSGD and PPCA varies across locations.
The main computational benefit of FSGD relative to periodic PCA is that it avoids repeated offline SVD and the storage of a sliding data window.
The four selected locations provide illustrative comparisons rather than evidence of an average predictive advantage across all NCEP targets.

\section{Proofs}
\label{sec: app proofs}
\subsection{Auxiliary Lemmas}
\begin{lemma}[Lemma 2 of \citet{li2025statistical}]
\label{lemma: lemma2}
    For any $q\geq 2$, any even integer $s\geq2$, and any two $d$-dimensional random vectors $w,v$, we have
    \[\Big\||{w}+{v}|_s\Big\|_q^2\leq\Big\||{w}|_s\Big\|_q^2+2\Big\||{w}|_s\Big\|_q^{2-q}\mathbb{E}\left(|{w}|_s^{q-s}\sum_{j=1}^dw_j^{s-1}v_j\right)+\left(\max\{q,s\}-1\right)\Big\||{v}|_s\Big\|_q^2.\]
\end{lemma}

\begin{lemma}[Norm of sub-Gaussian vectors]
\label{lemma: Norm of sub-Gaussian vectors}
If $X$ is a mean-zero, sub-Gaussian random vector in $\R^d$ with $\| X \|_{\psi_2} = K$. Let $B$ be an $d \times k$ matrix. Then for any $t>0$, we have
\[
\PP( |B^\top X |_{2} \geq CK\| B\|_F + t ) \leq \exp \Big( -\frac{ct^2}{K^2 \|B\|_{\rm op}^2} \Big),
\]
where $\| \cdot \|_F$ and $\|\cdot\|_{\op}$ denote the Frobenius norm, and operator norm respectively, $c,C$ are absolute constants.
\end{lemma}
\begin{proof}
    See Exercises 6.2.6 and 6.3.5 in \citet{vershynin2018high}.
\end{proof}

\begin{lemma}[Rosenthal Inequality]
\label{lemma:Rosenthal Inequality}
    Let $X_1, X_2, \cdots, X_n$ independent random variables with $\EE X_i = 0$, and $\EE |X_i|^q < \infty$ for $q \geq 2$. Then
    \[
    \EE | \sum_{i=1}^n X_i |^q \leq C_q \Big( \sum_{i=1}^n \EE |X_i|^q + (\sum_{i=1}^n \EE |X_i|^2)^{\frac{q}{2}}  \Big),
    \]
    where $C_q$ is constant only depending on $q$.
\end{lemma}

\begin{lemma}
\label{lemma: exp_moment_from_tail}
Let $Y\ge 0$ be a random variable. Let $B>0$ and $\beta>0$. Assume that for every $\delta\in(0,1)$ satisfying $\beta \delta < 1$,
\begin{equation}
\label{eq:hp_bound_param_delta}
\mathbb P\!\left(
Y > B\sqrt{\log\!\Big(\frac{1}{\beta\delta}\Big)}
\right)\le \delta .
\end{equation}
Then for any $q\ge 2$,
\begin{equation}
\label{eq:Lq_bound_simplified}
\|Y\|_q
\le
B\left(
\sqrt{\log_+\!\Big(\frac{1}{\beta}\Big)}
+
\sqrt{\frac{q}{2}}\,
\beta^{-1/q}
\right).
\end{equation}
\end{lemma}

\begin{proof}
Define $t_0 := B\sqrt{\log_+\!\Big(\frac{1}{\beta}\Big)}$ and 
\[
\delta(t):=\beta^{-1}\exp\!\left(-\frac{t^2}{B^2}\right) \in (0,1), \text{ for } t \geq t_0.
\]
We also note that $\beta \delta(t)= \exp(-\frac{t^2}{B^2}) < 1$ when $t\ge t_0$. Thus $\delta(t)$ lies in the admissible range.

Plugging $\delta(t)$ into
\eqref{eq:hp_bound_param_delta} yields the tail bound
\begin{equation}
\label{eq:tail_piecewise}
\mathbb P(Y>t)\le \delta(t)
=
\beta^{-1}\exp\!\left(-\frac{t^2}{B^2}\right),
\qquad t\ge t_0.
\end{equation}
For $t<t_0$, we trivially have $\mathbb P(Y>t)\le 1$. 
Now we consider the $L_q$-norm,
\[
\mathbb E[Y^q]
=
\int_0^\infty q t^{q-1}\mathbb P(Y>t)\,dt
\le
\int_0^{t_0} q t^{q-1}\,dt
+
\beta^{-1}\int_{t_0}^{\infty} q t^{q-1}e^{-t^2/B^2}\,dt .
\]
The first integral equals $t_0^q$. For the second one, change variables $u=t^2/B^2$:
\begin{align*}
\beta^{-1}\int_{t_0}^{\infty} q t^{q-1}e^{-t^2/B^2}\,dt
&\le
\beta^{-1}\int_{0}^{\infty} q t^{q-1}e^{-t^2/B^2}\,dt \\
&=
\beta^{-1}\, q\,\frac{B^q}{2}\int_0^\infty u^{\frac{q}{2}-1}e^{-u}\,du\\
&=
\beta^{-1}\,\frac{q}{2}\,B^q\,\Gamma\!\left(\frac{q}{2}\right).
\end{align*}
Hence,
\[
\mathbb E[Y^q]\le t_0^q + \beta^{-1}\,\frac{q}{2}\,B^q\,\Gamma\!\left(\frac{q}{2}\right).
\]
Taking $q$-th roots and using $(x+y)^{1/q}\le x^{1/q}+y^{1/q}$ for $q\geq 1$ yields
\begin{equation}
\label{eq: exp-decay-moment}
    \|Y\|_q
\le
t_0
+
B\left(
\beta^{-1}\,\frac{q}{2}\,\Gamma\!\left(\frac{q}{2}\right)
\right)^{1/q} \leq t_0+B\sqrt{\frac{q}{2}}\,
\beta^{-1/q}
\end{equation}
where the second inequality is due to $\Gamma(x)\le x^{x-1}$ for $x\ge 1/2$.
We complete the proof  by substituting the definition of $t_0$ in \eqref{eq: exp-decay-moment}.
\end{proof}

\begin{lemma}[Embedding between $l_s$ and $l_2$]
\label{lemma:Embedding between ls and l2}
Let $k\in\mathbb N$ and $z\in\mathbb R^k$. For any $s\in[1,\infty]$,
\begin{equation}
\label{eq:lps-l2}
|z|_s \;\le\; k^{\left(\frac{1}{s}-\frac{1}{2}\right)_+}\,|z|_2,
\qquad x_+ := \max\{x,0\}.
\end{equation}
Specifically, when $s \geq 2$, we have $|z|_s \;\le\; |z|_2$
\end{lemma}

Let $V,Q \in \RR^{d \times k}$ to be matrices with orthogonal columns. Suppose the singular values of $V^\top Q$ are $\sigma_1 \geq \sigma_2 \geq \cdots \geq \sigma_k \geq 0$. Then we call
\begin{equation}
\label{eq: def of principal angles}
    \Theta(Q, V)
=
\operatorname{diag}
\big(
\cos^{-1}(\sigma_1),
\cos^{-1}(\sigma_2),
\ldots,
\cos^{-1}(\sigma_k)
\big)
\end{equation}
as the principal angles. Lemma \ref{lemma: sin theta distance} provides some convenient expressions of $\sin \Theta$ distance, and shows $\operatorname{dist}(Q,V):=\| V V^\top - Q Q^\top \|$ distance is equivalent to $\sin \Theta$ distance.
\begin{lemma}[$\sin \Theta$ Distance  (Lemma 1 of \citet{cai2018rate})]
Let $V,Q \in \RR^{d \times k}$ to be matrices with orthogonal columns. Then
\label{lemma: sin theta distance}
    \[\|\sin\Theta(Q ,V)\|_{\op}=\sqrt{1-\sigma_{\min}^2(V^\top Q)},\]
    and
    \[
    \bigl\|\sin\Theta(Q, V)\bigl\|_{\op} \leq \operatorname{dist}(Q,V)= \bigl\| V V^\top - Q Q^\top \bigl\|_{\op} \leq2 \bigl\| \sin\Theta(Q, V) \bigl\|_{\op}.
    \]
Let $R =  \operatorname{arg}\min_{R \in \mathbb{O}_k} \big\| Q - V R \big\|_F$, and $D_{F}(Q,V) = \|Q-VR\|_{F}$, then
\[
\bigl\| \sin \Theta(Q, V) \bigr\|_{F}
\;\le\;
D_{F}(Q, V)
\;\le\;
\sqrt{2}\,\bigl\| \sin \Theta(Q, V) \bigr\|_{F}.
\]
\end{lemma}

Now, let $A \in \RR^{k \times k}$ be a non-singular matrix, then there exist a unique $Q$ with orthonormal column vectors and semi-definite $H$ such that 
\begin{equation}
\label{eq:polar decomposition of A}
    A = QH
\end{equation}
The decomposition \eqref{eq:polar decomposition of A} is called the polar decomposition of $A$. In fact, we have
\[
Q = A(A^\top A)^{-\frac{1}{2}}, \quad H = (A^\top A)^{\frac{1}{2}}.
\]

\begin{lemma}[Theorem 1 of \citet{li1995new}]
\label{lemma: Lipthitz of polar factor}
Let $A_1$ and $A_2$ be two $n \times n$ non-singular complex matrices whose
polar decompositions are given by \eqref{eq:polar decomposition of A}. Then
\begin{equation}
\label{eq:polar_Q_bound}
\| Q_1 - Q_2 \|_{\op}
\le
\frac{2}{\sigma_{min}(A_1) + \sigma_{min}(A_2)}
\,
\| A_1 - A_2 \|_{\op}.
\end{equation}
\end{lemma}

\subsection{Proofs for Proposition \ref{prop: Factor Estimation} in  Section \ref{subsec: Latent Factor Model}}

Suppose the factor model in \eqref{eq:factor model} holds. Define $\Sigma := \operatorname{Cov}(x)$, we write $V \in \RR^{d \times k}$ to be the orthogonal matrix whose columns are the leading $k$ eigenvectors of $\Sigma$. The following proposition shows that the column space of $V$ is close to column space of $B$.
\begin{lemma}
\label{lemma: V is close to B}
Define the normalized loading matrix
\[
B_0 := d^{-1/2}B \in \R^{d\times k},
\quad\text{such that } B_0^\top B_0 = I_k.
\]
Suppose factor model \eqref{eq:factor model} satisfies Assumption \ref{asp: factor model regularity}, then there exists $O_\star\in \mathbb{O}_k$ such that
\begin{equation}
\label{eq:prop_B0V_procrustes}
\|B_0 - V O_\star\|_{F}
\le \sqrt{2}\,\|\sin\Theta(B_0,V)\|_{F}
\le C\,\frac{K_u^2}{\lambda_k(\Sigma_f)} d^{-\frac{1}{2}},
\end{equation}
where $C$ is an absolute constant.
\end{lemma}
\begin{proof}
Denote $\Sigma_u := \operatorname{Cov}(u)$ and $\Sigma_f := \operatorname{Cov}(f)$. From $x=Bf+u$, $\EE f=0$, and the independence of $f$ and $u$, we have
\[
\Sigma 
= B\Sigma_f B^\top+\Sigma_u, 
\]
We denote the signal covariance as $\Sigma_s:=B\Sigma_fB^\top = d\,B_0\Sigma_f B_0^\top$, so that
$\Sigma=\Sigma_s+\Sigma_u$.

Since $\Sigma_f$ is symmetric positive definite with $\lambda_k(\Sigma_f)>0$, the matrix
$B_0\Sigma_fB_0^\top$ has rank $k$ and its nonzero eigenvalues are exactly
$\{\lambda_j(\Sigma_f)\}_{j=1}^k$, while the remaining
eigenvalues are $0$. Consequently,
\[
\lambda_k(\Sigma_s)= d\,\lambda_k(\Sigma_f),\qquad \lambda_{k+1}(\Sigma_s)=0,
\]
and the eigengap equals
\[
\delta_k := \lambda_k(\Sigma_s)-\lambda_{k+1}(\Sigma_s)=d\,\lambda_k(\Sigma_f)>0.
\]

Let $V\in\R^{d\times k}$ collect the leading $k$ eigenvectors of $\Sigma$.
By the Davis--Kahan $\sin\Theta$ theorem (Theorem 2 in \cite{yu2015useful}), it holds that 
\begin{equation}
    \label{eq: Davis-Kahan Main Bound}
    \|\sin\Theta(B_0,V)\|_{F}
\le \frac{\|\Sigma-\Sigma_s\|_{F}}{\delta_k}
= \frac{\|\Sigma_u\|_{F}}{d\,\lambda_k(\Sigma_f)}.
\end{equation}
By definition of the sub-Gaussian norm of a random vector,
$\|u\|_{\psi_2}=\sup_{\|v\|_2=1}\|v^\top u\|_{\psi_2}=K_u$.
Thus, for any unit vector $v$, the random variable $v^\top u$ is also 
sub-Gaussian with $\|v^\top u\|_{\psi_2}\le K_u$. The standard moment bound for
sub-Gaussian random variables implies 
\begin{equation}
    \|\Sigma_u\|_{\op}
=\sup_{\|v\|_2=1} v^\top \Sigma_u v
\le \sup_{\|v\|_2=1}\EE[(v^\top u)^2]
\le C K_u^2,
\end{equation}
It then implies
\begin{equation}
    \label{eq:Sigma_u bound}
    \|\Sigma_u\|_{F} \leq d^{\frac{1}{2}} \|\Sigma_u\|_{\op} \leq C K_u^2  d^{\frac{1}{2}}
\end{equation}

At the same time, Lemma \ref{lemma: sin theta distance} implies
\begin{equation}
\label{eq: sp <= sin}
    \inf_{O\in  \mathbb{O}_k }\|B_0-VO\|_{F}
\le \sqrt{2}\,\|\sin\Theta(B_0,V)\|_{F}.
\end{equation}
We complete the proof by combining \eqref{eq: Davis-Kahan Main Bound}, \eqref{eq:Sigma_u bound} and \eqref{eq: sp <= sin}.
\end{proof}

\begin{lemma}
    \label{lemma: error is p-1/2}
    For any $Q \in \RR^{d \times k}$ consisting of orthonormal columns, suppose random vector $u_i \in \RR^d$ is sub-Gaussian with $\| u_i \|_{\psi_2} \leq K_u$, and $s \geq 2$, then
    \[
    \Big \| | d^{-\frac{1}{2}} Q^\top u_i |_s \Big\|_q \leq \, K_u( k^{\frac{1}{2}} + C  \, q^{\frac{1}{2}}) \cdot d^{-\frac{1}{2}} ,
    \]
    where $c, C$ are absolute constants.
\end{lemma}

\begin{proof}
    Since $Q \in \R^{d\times k}$ has orthonormal columns, it satisfies $\|Q\|_F = k^{\frac{1}{2}}$ and $\|Q\|_{\op} = 1$.  Applying Lemma \ref{lemma: Norm of sub-Gaussian vectors}, it holds that
\begin{equation}
    \PP\Big( |Q^\top u_i|_2 \geq C \,K_u\, k^{\frac{1}{2}} + t \Big) \leq \exp \Big( -\frac{ct^2}{K_u^2} \Big) 
\end{equation}
So that random variable $|Q^\top u_i|_2$ is sub-Gaussian. By the property of sub-Gaussian, the moment satisfies
\begin{equation}
    \Big \| |Q^\top u_i|_2 \Big \|_q \leq c \, K_u k^{\frac{1}{2}} + C K_u q^{\frac{1}{2}},
\end{equation}
where $c, C$ are absolute constants. Next, followed by Lemma \ref{lemma:Embedding between ls and l2} and $s\geq 2$, we have
\[
\Big \| |Q^\top u_i|_s \Big \|_q \leq c \, K_u k^{\frac{1}{2}} + C K_u   \, q^{\frac{1}{2}},
\]
The proof can be finished by multiplying factor $d^{-\frac{1}{2}}$ on both sides.
\end{proof}

\begin{proof}[Proof of Proposition \ref{prop: Factor Estimation}]
Let $O_*:= \operatorname{arg}\min_{O \in \mathbb{O}_k } \| B_0-V O\|_F$ as defined in Lemma \ref{lemma: V is close to B}, where $B_0:=d^{-\frac{1}{2}}B$. We then define $R:= \operatorname{arg}\min_{O \in \mathbb{O}_k } \| VO_*-QR\|_F$.

 Using $x=d^{1/2}B_0f+u$ and $\hat f=d^{-1/2}Q^\top x$,
\[
\hat f
= d^{-1/2}Q^\top(d^{1/2}B_0f+u)
= Q^\top B_0f + d^{-1/2}Q^\top u,
\]
Hence
\[
\hat f - Rf
= d^{-1/2}Q^\top u + (Q^\top B_0-R)f = d^{-1/2}Q^\top u + (Q^\top B_0-Q^\top VO_\star)f + (Q^\top VO_\star-R)f
\]
Applying triangular inequality gives
\begin{equation}
    \label{eq: factor estimation decomposition - full}
    \Big\|\,|\hat f-Rf|_s\,\Big\|_{q}
\le
\Big\|\,|d^{-1/2}Q^\top u|_s\,\Big\|_{q}
+
\Big\|\,|(Q^\top B_0-Q^\top VO_\star)f|_s\,\Big\|_{q} 
+
\Big\|\,|(Q^\top VO_\star-R)f|_s\,\Big\|_{q} 
.
\end{equation}
Lemma \ref{lemma: error is p-1/2} provides the error bound of the first term of \eqref{eq: factor estimation decomposition - full}, i.e.
\begin{equation}
    \label{eq: factor estimation decomposition - term1}
    \Big \| | d^{-\frac{1}{2}} Q^\top u_i |_s \Big\|_q \leq K_u d^{-\frac{1}{2}} \cdot ( c \, k^{\frac{1}{2}}+C \, q^{\frac{1}{2}}),
\end{equation}
where $c, C$ are absolute constants. 

Note that $Q$ is defined to have orthonormal columns. Applying Lemma \ref{lemma: V is close to B}, Lemma \ref{lemma:Embedding between ls and l2} and the relationship $\| \cdot\|_{\op} \leq \| \cdot \|_F$, we have
\begin{equation}
    \label{eq: factor estimation decomposition - term3}
    \Big\|\,|(Q^\top B_0-Q^\top VO_\star)f|_s\,\Big\|_{q}  \leq \Big\|\,\|(B_0-VO_\star) \|_{\op} |f|_2\,\Big\|_{q} \leq C \frac{K_u^2} { \lambda_k(\Sigma_f)} \Big\||f|_2\,\Big\|_{q} d^{-\frac{1}{2}}
\end{equation}

It remains to bound the second term $\Big\|\,|(Q^\top VO_\star-R)f|_s\,\Big\|_{q}$ on RHS of \eqref{eq: factor estimation decomposition - full}. By Lemma \ref{lemma:Embedding between ls and l2}, we only need to focus on $\Big\|\,|(Q^\top VO_\star-R)f|_2\,\Big\|_{q}$.

Utilizing Lemma \ref{lemma: sin theta distance}, the independence of $(Q,R)$ and $f$, and the relationship $\|A\|_{\op} \leq \|A\|_{F} \leq k^{\frac{1}{2}}\|A\|_{\op}$ for $A\in\RR^{d\times k}, k \leq d$, we have
\begin{align}
\label{eq: factor estimation decomposition - dist}
    \Big\|\,|(Q^\top VO_\star-R)f|_s\,\Big\|_{q} &\leq  
    \Big\|\,\|Q^\top VO_\star-R\|_{\op}\,\Big\|_{q} \Big\|\,|f|_2\,\Big\|_{q} 
    \leq 
     \Big\|\,\|Q^\top VO_\star-R\|_{F}\,\Big\|_{q} \Big\|\,|f|_2\,\Big\|_{q}  \nonumber \\
     & \leq  \Big\|\,\|VO_\star-QR\|_{F}\,\Big\|_{q} \Big\|\,|f|_2\,\Big\|_{q} \leq\sqrt{2}  \,\Big\| \|\sin \Theta(Q,VO_*)\|_F \Big\|_q \Big\|\,|f|_2\,\Big\|_{q} \nonumber \\
     & \leq\sqrt{2}k^{\frac{1}{2}}\Big\|\|\sin \Theta(Q,VO_*)\|_{\op} \Big\|_q \Big\|\,|f|_2\,\Big\|_{q} \leq   \sqrt{2} k^{\frac{1}{2}} \Big\| \operatorname{dist}(Q,VO_\star) \Big\|_{q}\Big\|\,|f|_2\,\Big\|_{q} \nonumber \\
     &= \sqrt{2} k^{\frac{1}{2}} \Big\| \operatorname{dist}(Q,V) \Big\|_{q}\Big\|\,|f|_2\,\Big\|_{q}
\end{align}
where $\operatorname{dist}(Q,VO_\star) = \operatorname{dist}(Q,V) $ since $O_k \in \RR^{k \times k}$ is orthonormal matrix.

Combining \eqref{eq: factor estimation decomposition - term1}, \eqref{eq: factor estimation decomposition - term3}, \eqref{eq: factor estimation decomposition - dist} and \eqref{eq: factor estimation decomposition - full}, we obtain
\begin{align}
    \Big\| |\hat{f} -Rf|_s \Big\|_q \leq \Big( C \frac{K_u^2}{ \lambda_k(\Sigma_f)} \Big\||f|_2\,\Big\|_{q}
+
K_u   ( c \, k^{\frac{1}{2}}+C \, q^{\frac{1}{2}}) \Big) \cdot d^{-\frac{1}{2}} + \sqrt{2}k^{\frac{1}{2}}  \Big\| \operatorname{dist}(Q,V) \Big\|_{q}\Big\|\,|f|_2\,\Big\|_{q}
\end{align}

Now we complete the proof.
\end{proof}

\subsection{Proofs for Section \ref{subsec: Factor Augmented SGD}}
\label{subsec: Proofs for Factor Augmented SGD}
In this section, we show that the additional alignment step introduced in Stage II of Algorithm~\ref{alg:FASGD} does not affect the theoretical bounds established in Theorem~3 of \citet{huang2021streaming}. 

Now let's recall the Oja's iteration. Let $\{A_t\}_{t\ge1}\subset\R^{d\times d}$ be a sequence of matrices and
$\{\eta^o_t\}_{t\ge1}$ be step sizes. Initialize with any $Z_0\in\R^{d\times k}$ of full column rank
and set $Q_0=\mathrm{QR}[Z_0]$, where $\mathrm{QR}[\cdot]$ denotes the $Q$-factor of the thin QR
decomposition. For $t\ge1$, we set
\begin{equation}
\label{eq:oja_update_align}
Q_t' := \mathrm{QR}[(I+\eta_t^{\rm oja} A_t)\,Q_{t-1}].
\end{equation}
Once we have $Q_t'$, we perform an orthogonal Procrustes alignment to fix the rotational ambiguity of $Q$'s. Specifically, we compute $S_t:=Q_{t}'^\top Q_{t-1}$, and take SVD $S_t:=U_t \Lambda_t V_t^\top$. We then set $Q_t:=Q_t'U_tV_t^\top$. Equivalently, $Q_t$ is the unique solution to the orthogonal Procrustes problem
\begin{equation}
    \label{eq: align-update}
    Q_t:= Q_t'O_t, \quad O_t = \operatorname{arg}\min_{O\in \mathbb{O}_{k}} \|Q_{t-1} - Q_{t}'O\|_F,
\end{equation}
where $O_t$ has closed form solution, $O_t=S_t (S_t^\top S_t)^{-\frac{1}{2}}=U_tV_t^\top$.

\begin{lemma}
\label{lemma: Factor subspace -- Our}
    If $\sup_{P\in\mathbb{O}_{d,k}}\| P^{\top}(A_t - \Sigma) \|_{\op}\leq M$ almost surely, Let $\lambda_1 \geq \cdots \geq \lambda_d$ be the eigenvalues of $\Sigma$ , and let $\rho_k = \lambda_k - \lambda_{k+1}$. For any $\delta\in (0,1)$, define
    \begin{align*}
    T_0=\Theta&\left(\frac{kM^2}{\delta^2\rho_k^2}{\log^4(\frac{dM}{\rho_k\delta})}\right),
    \;
    \beta=\Theta\left(\frac{M^2}{\rho_k^2}\log(\frac{Mk}{\rho_k\delta})\right),
    \;\\
    &\quad  \eta^{\rm oja}_t=\left\{\begin{array}{ll}\Theta\left(\frac{\log(d/\delta)}{\rho_kT_0}\right),&t\leq T_0\\
    \Theta\left(\frac{1}{\rho_k(\beta+t-T_0)}\right),&t>T_0.\end{array}\right.
    \end{align*}
    Suppose $Q_t$  follows the updating rule \eqref{eq:oja_update_align} and \eqref{eq: align-update}, then for any $T > T_0$, $Q_T$ satisfies 
    \[
    \operatorname{dist}(Q_T,V)\leq C^{\prime}\frac{M}{\rho_k}\sqrt{\frac{\log(Mk/\rho_k\delta)}{T-T_0}}
    \]
    with probability at least $1-\delta$, where $C'$ is an absolute positive constant.
\end{lemma}
\begin{proof}
    Using the same notation of \citet{huang2021streaming}, we write $V \in \RR^{d \times k}$ for the orthogonal matrix whose columns are the $k$ leading eigenvectors of $\Sigma$, and $U \in \RR^{d\times (d-k)}$ for the orthogonal matrix whose columns are the remaining eigenvectors. Let $Z_t := Y_t Y_{t-1} \cdots Y_1 Z_0$, where $Y_i = I_k+\eta_{i}A_i$, it suffices to prove 
    \[
    \operatorname{dist}(Q_t, V) \leq \|W_t \|_{\op}, \quad W_t := U^\top Z_t (V^\top Z_t)^{-1}
    \]
    The remaining proof are identical to \citet{huang2021streaming}.
    Define $B_t:=(R_t')^{-1} O_t$, we first claim that
    \begin{equation}
    \label{eq: claim for Q_t}
        Q_t = Y_t Y_{t-1} \cdots Y_1 Q_0 B_{1}\cdots B_{t-1}B_t, t\geq 0,
    \end{equation}
    Equation~\eqref{eq: claim for Q_t} holds for $t=0$. We prove the claim by mathematical induction on $t$. Assume that~\eqref{eq: claim for Q_t} holds at time $t-1$, i.e., it holds for $Q_{t-1}$. We now show that~\eqref{eq: claim for Q_t} also holds for $Q_t$.

    Since $Q_t'=QR[Y_t Q_{t-1}]$ has full column rank, there exists $R_t'\in\RR^{k\times k}$ upper triangular and non-singular, such that $Q_t' R_t' = Y_t Q_{t-1}$, then
    \begin{equation}
        Q_t' = Y_t Q_{t-1} (R_t')^{-1}
    \end{equation}
    Followed by \eqref{eq: align-update}, it follows that
    \begin{equation}
        Q_t = Q_t'O_t = Y_t Q_{t-1}(R_t')^{-1}O_t =Y_t Q_{t-1}B_t = Y_tY_{t-1} \cdots Y_1 Q_0 B_{1}\cdots B_{t-1}B_t,
    \end{equation}
    Therefore, the claim holds for all $Q_t$.
    Recall that $Q_0 = QR[Z_0]$, there exist $R_0$ such that $Q_0R_0 = Z_0$, we can write \eqref{eq: claim for Q_t} as
    \begin{equation}
        Q_t = Z_t  P_t, \quad P_t:= R_0^{-1}B_1 \cdots B_t \in \R^{k \times k} \text{ non-singular}
    \end{equation}
    
    We also note that $\|V^\top Q_t\|_{\op} \leq 1$, then we complete the proof by showing that
    \begin{align}
        \operatorname{dist}(Q_t,V)&=\|U^\top Q_{t} \|_{\op} \leq \|U^\top Q_{t} (V^\top Q_t)^{-1} \|_{\op} \cdot \| V^\top Q_t\|_{\op} \leq \|U^\top Q_{t} (V^\top Q_t)^{-1} \|_{\op} \nonumber \\
        &=\|U^\top Z_tP_t (V^\top Z_tP_t)^{-1} \|_{\op} \nonumber\\
        &=\| U^\top Z_t (V^\top Z_t)^{-1}\|_{\op}
    \end{align}
\end{proof}


\subsection{Proof for Lemma \ref{lemma: Lq moment bound for oja update - main} in Section \ref{subsec: Factor Augmented SGD}}

By adapting the proof of \cite{huang2021streaming}, we first establish an \(L_q\)-moment bound for the Oja update.
    Let
\[
Y_i:=I+\eta_i^{\rm oja}A_i,
\qquad
Z_t:=Y_tY_{t-1}\cdots Y_1Q_0.
\]
We write 
\(V\in\mathbb R^{d\times k}\) and \(U\in\mathbb R^{d\times(d-k)}\) be the
orthogonal matrices whose columns are respectively the top \(k\) and remaining
\(d-k\) eigenvectors of \(\Sigma\). 
Following \cite{huang2021streaming}, we define
\[
W_t:=U^\top Z_t(V^\top Z_t)^{-1}
\]
We then define the good events as
\begin{equation}
    \label{eq: good events}
    G_0:=\{\|W_0\|_{\op}\le1\},
\qquad
G_t:=G_{t-1}\cap\{\|W_t\|_{\op}\le \sqrt{2e} \},
\qquad t\ge1.
\end{equation}

The following lemma gives the complete statement of Lemma~\ref{lemma: Lq moment bound for oja update - main}.
\begin{lemma}[$L^q$-Moment Bound for Oja Update]
\label{lemma: Lq moment bound for oja update}
    Suppose Assumption \ref{asp: ojs_assump} holds. Set $\eta_t^{\rm oja}
=
\frac{\alpha}{(\beta+t)\rho_k}$ and define the normalized gap
    \[
\bar\rho_k
:=
\min\left\{
\frac{\rho_k}{M},
\frac{\rho_k}{\|\Sigma\|_{\op}},
1
\right\}.
\]
If
\[
\alpha\ge 8,
\qquad
\beta\ge
\max \big\{ \frac{4(1+\sqrt{2e})\alpha}{\bar\rho_k}, \,
C\cdot \frac{\alpha^2}{\bar\rho_k^2} \big\},
\]
where $C$ is an absolute constant. Then for every $q \geq 2$ and $t \geq 1$, it holds that
\[
\left\|\operatorname{dist}(Q_t,V)\right\|_q
\le
k^{1/q}
\left[
\left(\frac{\beta+1}{\beta+t}\right)^{\alpha/2}
+
C\cdot\frac{\alpha q^{1/2}}{\bar\rho_k}\cdot
\frac{t^{1/2}}{\beta+t}
+
C^{2/q} \cdot \alpha^{2/q}\bar\rho_k^{-2/q}
\exp\left\{
-\frac{\beta\bar\rho_k^2}{\alpha^2 C^2q}
\right\}
\right].
\]
Additionally, for any $t \geq 1$ ,we have
\begin{equation}
    \mathbb P(G_t^c)
\le
k C^2 \alpha^2\bar\rho_k^{-2}
\exp\left\{
-\frac{\beta\bar\rho_k^2}{C^2\alpha^2}
\right\}.
\end{equation}
\end{lemma}

\begin{proof}
By Lemma 6 of \cite{huang2021streaming}, it holds that
\begin{equation}
    \operatorname{dist}(Q_t,V)
=
\|U^\top Q_t\|_{\op}
\le
\|U^\top Q_t(V^\top Q_t)^{-1}\|_{\op}
=
\|U^\top Z_t(V^\top Z_t)^{-1}\|_{\op}
=
\|W_t\|_{\op}.
\label{eq:huang_lemma6_reduction}
\end{equation}

We now verify that \(\|W_0\|_{\op}
<1\) holds almost surely.
Since
\[
\operatorname{dist}(Q_0,V)
=
\|U^\top Q_0\|_{\op}
\le \frac12,
\]
we have
\[
(V^\top Q_0)^\top(V^\top Q_0)
=
Q_0^\top V V^\top Q_0
=
I_k-Q_0^\top U U^\top Q_0.
\]
Therefore
\[
\sigma_{\min}(V^\top Q_0)^2
=
1-\|U^\top Q_0\|_{\op}^2
\ge
1-\frac14
=
\frac34,
\]
and hence
\[
\sigma_{\min}(V^\top Q_0)\ge \frac{\sqrt3}{2}.
\]
It follows that
\[
\|W_0\|_{\op}
=
\|U^\top Q_0(V^\top Q_0)^{-1}\|_{\op}
\le
\|U^\top Q_0\|_{\op}
\cdot
\|(V^\top Q_0)^{-1}\|_{\op}
\le
\frac12\cdot\frac{2}{\sqrt3}
=
\frac1{\sqrt3}
<1.
\label{eq:warm_start_to_huang}
\]
By Proposition 11 of \cite{huang2021streaming}, for every $p\geq 2$, we have
\begin{equation}
    \|W_t\mathbf 1_{G_t}\|_{p,p}^2
\le
k^{2/p}
\left(\frac{\beta+1}{\beta+t}\right)^\alpha
+
p k^{2/p}
\left(\frac{C\alpha}{\bar\rho_k}\right)^2
\frac{t}{(\beta+t)^2},
\label{eq:huang_prop11_exact}
\end{equation}
where, for a random matrix \(X\), 
\[
\|X\|_{p,p}
:=
\left(\mathbb E\|X\|_{S_p}^p\right)^{1/p},
\qquad
\|X\|_{S_p}
:=
\left(\sum_{i\ge1}\sigma_i(X)^p\right)^{1/p},
\]
and \(\{\sigma_i(X)\}_{i\ge1}\) are the singular values of \(X\). Here \(\|\cdot\|_{S_p}\) denotes the Schatten-\(p\) norm, and \(\|\cdot\|_{p,p}\) denotes the corresponding \(L_p\)-norm of this random matrix norm.

Now we decompose
\begin{equation}
    \|\operatorname{dist}(Q_t,V)\|_q
\le
\|\operatorname{dist}(Q_t,V)\mathbf 1_{G_t}\|_q
+
\|\operatorname{dist}(Q_t,V)\mathbf 1_{G_t^c}\|_q.
\label{eq:split_good_bad}
\end{equation}
First consider the good-event term. By \eqref{eq:huang_lemma6_reduction} and $\| \cdot \|_{\op} \leq \| \cdot \|_{Sq}$,
\begin{align}
\label{eq:good_part_schatten}
\|\operatorname{dist}(Q_t,V)\mathbf 1_{G_t}\|_q \leq
\|W_t\mathbf 1_{G_t}\|_{\op,L_q}  \leq
\|W_t\mathbf 1_{G_t}\|_{S_q,L_q} =
\|W_t\mathbf 1_{G_t}\|_{q,q}.
\end{align}
Applying \eqref{eq:huang_prop11_exact} with \(p=q\), we get
\[
\|W_t\mathbf 1_{G_t}\|_{q,q}^2
\le
k^{2/q}
\left(\frac{\beta+1}{\beta+t}\right)^\alpha
+
q k^{2/q}
\left(\frac{C\alpha}{\bar\rho_k}\right)^2
\frac{t}{(\beta+t)^2}.
\]
Taking square roots and using \(\sqrt{x+y}\le \sqrt{x}+\sqrt{y}\),
\begin{equation}
    \|\operatorname{dist}(Q_t,V)\mathbf 1_{G_t}\|_q
\le
k^{1/q}
\left[
\left(\frac{\beta+1}{\beta+t}\right)^{\alpha/2}
+
\frac{C\alpha\sqrt{q\,t}}{\bar\rho_k(\beta+t)}
\right].
\label{eq:good_part_final}
\end{equation}
Next consider the bad-event term. Since
\[
0\le \operatorname{dist}(Q_t,V)\le1,
\]
we have
\begin{equation}
    \|\operatorname{dist}(Q_t,V)\mathbf 1_{G_t^c}\|_q
\le
\mathbb P(G_t^c)^{1/q}.
\label{eq:bad_part_dist_bound}
\end{equation}
Following the proof of Theorem 12 of \cite{huang2021streaming}, it holds that 
\begin{equation}
    \mathbb P(G_t^c)
\le
k C^2 \alpha^2\bar\rho_k^{-2}
\exp\left\{
-\frac{\beta\bar\rho_k^2}{C^2\alpha^2}
\right\}.
\label{eq:bad_prob_final}
\end{equation}
By \eqref{eq:bad_part_dist_bound} and \eqref{eq:bad_prob_final}, we then have
\begin{equation}
    \|\operatorname{dist}(Q_t,V)\mathbf 1_{G_t^c}\|_q
\le
 C^{2/q}  k^{1/q} \alpha^{2/q}\bar\rho_k^{-{2/q}}
\exp\left\{
-\frac{\beta\bar\rho_k^2}{C^2\alpha^2q}
\right\}
\label{eq:bad_part_final}
\end{equation}
Finally, combining \eqref{eq:split_good_bad}, \eqref{eq:good_part_final}, and
\eqref{eq:bad_part_final}, we obtain
\[
\left\|\operatorname{dist}(Q_t,V)\right\|_q
\le
k^{1/q}
\left[
\left(\frac{\beta+1}{\beta+t}\right)^{\alpha/2}
+
C\cdot\frac{\alpha q^{1/2}}{\bar\rho_k}\cdot
\frac{t^{1/2}}{\beta+t}
+
C^{2/q} \cdot \alpha^{2/q}\bar\rho_k^{-2/q}
\exp\left\{
-\frac{\beta\bar\rho_k^2}{\alpha^2 C^2q}
\right\}
\right].
\]
\end{proof}

\begin{remark}
    By taking 
    \[
\beta \ge
\frac{\alpha^2 C^2 q}{\bar\rho_k^2}
\left[
\frac{1}{2}\log d
+
\frac{1}{q}
\log\left(
\frac{k C^2 \alpha^2}{\bar\rho_k^2}
\right)
\right] =  \Theta\Big( \bar \rho_k^{-2} \big( q\log d+ \log \frac{k}{\bar \rho_k^2} \big) \Big),
\]
the last term has the order of $\Ocal(d^{-1/2})$.
\end{remark}

\subsection{Proofs for 
Proposition \ref{prop: Stability of theta* in Oja's Update} in 
Section \ref{sec: FSGD}}
\label{sec: proofs for theta^* stability}

\begin{lemma}[Stability of $\theta^*$]
\label{lemma: theta_star_rotation_stability}
Let $R_1,R_2\in\R^{k\times k}$ be orthogonal matrices (possibly random), and suppose $(f,y)$ is independent of $(R_1,R_2)$. For $i=1,2$, define $\theta^*_i$ as
\[
F_i(\theta):=\EE\!\big[l(\theta;R_i f,y)\,\big|\,R_i\big],
\qquad 
\theta_i^*:=\arg\min_{\theta} F_i(\theta).
\]
Under Assumption~\ref{asp:Strongly Convex} and Assumption~\ref{asp:Lipschitz Continuity for f}, we have
\[
\Big\|\,\big|\theta_1^*-\theta_2^*\big|_s\,\Big\|_q
\;\le\;
\frac{L_{s,f}}{\mu}\;
\Big\|\,\big|(R_1-R_2)f\big|_s\,\Big\|_q.
\]
\end{lemma}

\begin{proof}
Note that that $F_1$ is $l^s$-strong convex with parameter $\mu$.
i.e. for all $\theta_1,\theta_2$,
\begin{equation}
\label{eq:F1_strongly_convex}
\big\langle (\theta_1-\theta_2)^{\circ(s-1)},\,
\nabla F_1(\theta_1)-\nabla F_1(\theta_2)\big\rangle
\ge \mu\,|\theta_1-\theta_2|_s^s.
\end{equation}

Then, we apply \eqref{eq:F1_strongly_convex} with $(\theta_1,\theta_2)=(\theta_1^*,\theta_2^*)$. Note that $\nabla F_1(\theta_1^*)=0$, it holds that
\[
\mu\,|\theta_1^*-\theta_2^*|_s^s
\le
\big\langle (\theta_1^*-\theta_2^*)^{\circ(s-1)},\,\nabla F_1(\theta_1^*)-\nabla F_1(\theta_2^*)\big\rangle
=
-\big\langle (\theta_1^*-\theta_2^*)^{\circ(s-1)},\,\nabla F_1(\theta_2^*)\big\rangle.
\]
Applying H\"older's inequality with the parameters $s$ and $s/(s-1)$ yields
\[
\mu\,|\Delta|_s^s
\le
\big|\big\langle \Delta^{\circ(s-1)},\,\nabla F_1(\theta_2^*)\big\rangle\big|
\le
\big|\Delta^{\circ(s-1)}\big|_{s/(s-1)}\;\big|\nabla F_1(\theta_2^*)\big|_s
=
|\Delta|_s^{s-1}\;\big|\nabla F_1(\theta_2^*)\big|_s,
\]
where $\Delta:=\theta_1^*-\theta_2^*$ for notation brevity.
We only consider $|\Delta|_s \neq 0$, otherwise it's trivial. Dividing both sides by $|\Delta|_s^{s-1}$ gives
\begin{equation}
\label{eq:delta_grad_bound}
|\theta_1^*-\theta_2^*|_s
\le
\frac{1}{\mu}\,\big|\nabla F_1(\theta_2^*)\big|_s 
=
\frac{1}{\mu}\,\big|\nabla F_1(\theta_2^*) - \nabla F_2(\theta_2^*) \big|_s,
\end{equation}
where the last equality holds by definition of $\theta^*_2$.

Applying Jensen's inequality, it holds that
\[
\big|\nabla F_1(\theta_2^*) - \nabla F_2(\theta_2^*) \big|_s
\le
\EE\!\big[\big|\nabla_\theta l(\theta_2^*;R_1 f,y)-\nabla_\theta l(\theta_2^*;R_2 f,y)\big|_s\,\big|\,R_1,R_2\big],
\]
where we can condition on $(R_1,R_2)$, since $F_1$ depends only on $R_1$ not on $R_2$, and vice versa.

By applying condition Jensen inequality, $\nabla_\theta F_i(\theta_i^*)=0$, and utilizing Assumption~\ref{asp:Lipschitz Continuity for f}, we obtain
\begin{align}
\label{eq:grad-q-lip}
    \Big\| \big|\nabla F_1(\theta_2^*) - \nabla F_1(\theta_1^*) \big|_s \Big\|_q &\leq \EE^{\frac{1}{q}}\Big[  \EE^q\big( \big|\nabla_\theta l(\theta_2^*;R_1 f,y)-\nabla_\theta l(\theta_2^*;R_2 f,y)\big|_s\,\big|\,R_1,R_2  \big)\Big] \nonumber \\
    &\leq \EE^{\frac{1}{q}}\Big[  \EE\big( \big|\nabla_\theta l(\theta_2^*;R_1 f,y)-\nabla_\theta l(\theta_2^*;R_2 f,y)\big|_s^q\,\big|\,R_1,R_2  \big)\Big] \nonumber\\
    &=\EE^{\frac{1}{q}}\Big[  \big|\nabla_\theta l(\theta_2^*;R_1 f,y)-\nabla_\theta l(\theta_2^*;R_2 f,y)\big|_s^q\Big] \nonumber \\ 
    & = \Big\| \big|\nabla_\theta l(\theta_2^*;R_1 f,y)-\nabla_\theta l(\theta_2^*;R_2 f,y)\big|_s \Big\|_q \nonumber\\
    &\leq L_{s,f} \Big\| |(R_1-R_2)f|_s \Big\|_q 
\end{align}

Finally, we complete the proof by combining \eqref{eq:delta_grad_bound} and \eqref{eq:grad-q-lip}:
\[
\Big\|\,\big|\theta_1^*-\theta_2^*\big|_s\,\Big\|_q \leq
\frac{1}{\mu} \Big\| \big|\nabla F_1(\theta_2^*) - \nabla F_1(\theta_1^*) \big|_s \Big\|_q \
\le
\frac{L_{s,f}}{\mu}\;
\Big\|\,|(R_1-R_2)f|_s\,\Big\|_q.
\]
\end{proof}



\begin{lemma}[Procrustes Lipschitz Bound]
\label{lemma: Lipthitz of Rotation R}
Let \(V,U\) be the matrices of the leading and remaining eigenvectors of
\(\Sigma\). For \(Q\in\mathcal O_{d,k}\), define
\[
W(Q):=U^\top Q(V^\top Q)^{-1}
\]
whenever \(V^\top Q\) is invertible. For \(i=1,2\), let
\[
R_i
:=
\arg\min_{R\in\mathbb O_k}\|Q_i-VR\|_F .
\]
If
\[
\|W(Q_i)\|_{\op}\le \kappa,\qquad i=1,2,
\]
then
\[
\|R_1-R_2\|_{\op}
\le
C_\kappa \|V^\top(Q_1-Q_2)\|_{\op}
\le
C_\kappa \|Q_1-Q_2\|_{\op},
\]
where \(C_\kappa\) depends only on \(\kappa\). In particular, for
\(\kappa=\sqrt{2e}\), \(C_\kappa\) is an absolute constant.
\end{lemma}

\begin{proof}
Let
\[
A_i:=V^\top Q_i,\qquad B_i:=U^\top Q_i.
\]
Since \(Q_i\) has orthonormal columns, i.e. $Q_i^\top Q_i = I_k$, it holds that
\[
A_i^\top A_i+B_i^\top B_i=I_k.
\]
Moreover, \(B_i=W(Q_i)A_i\) by definition of $W(Q_i)$. Hence
\[
A_i^\top(I_k+W(Q_i)^\top W(Q_i))A_i=I_k.
\]
For any \(x\in\mathbb R^k\), we note that
\[
\|x\|_2^2
=
\|(I_k+W(Q_i)^\top W(Q_i))^{1/2}A_i x\|_2^2
\le
(1+\kappa^2)\|A_i x\|_2^2.
\]
Therefore
\[
\sigma_{\min}(A_i)
\ge
(1+\kappa^2)^{-1/2},
\qquad i=1,2.
\]
The Procrustes solution is the polar factor of \(A_i\):
\[
R_i=A_i(A_i^\top A_i)^{-1/2}.
\]
By the polar-factor perturbation bound (Lemma \ref{lemma: Lipthitz of polar factor}),
\[
\|R_1-R_2\|_{\op}
\le
\frac{2}{\sigma_{\min}(A_1)+\sigma_{\min}(A_2)}
\|A_1-A_2\|_{\op}.
\]
Using the lower bound on \(\sigma_{\min}(A_i)\),
\[
\|R_1-R_2\|_{\op}
\le
\sqrt{1+\kappa^2}\|A_1-A_2\|_{\op}
=
\sqrt{1+\kappa^2}\|V^\top(Q_1-Q_2)\|_{\op}.
\]
This proves the lemma with \(C_\kappa=\sqrt{1+\kappa^2}\).
\end{proof}


\begin{lemma}[Theorem 3.1 of \citet{stewart1977perturbation}]
\label{Lemma: QR-Perturbation --- Q-Factor}
    Let $A = QR$, where $A \in \RR^{d \times k}$ has rank $k$ and $Q \in \RR^{d \times k}$ with $Q^\top Q = I_k$. Let $E \in \RR^{d \times k}$ satisfies $\|A^\dagger\|_F\|E\|_F<\frac{1}{2}$. Let the linear operator $T:\mathbb{R}^{k\times k}\to\mathbb{R}^{k\times k}$ that maps the space of upper triangular matrices into the space of symmetric matrices be defined by
\[
T(F) := RF + F^\top R,
\]
and define
\[
F_{1}
:= T^{-1}\!\left[ R^\top (Q^\top E) + (E^\top Q)R \right].
\]
Under the condition
\[
\tau \|F_{1,t}\|_F < \frac14,
\qquad
\tau = k \|R^{-1}\|_F \Big(1+\frac{1}{\sqrt{2}}\kappa(R) \Big)
\]
Then there exist matrices $W \in \R^{d\times k}, F\in \RR^{k \times k}$, such that $A+E = (Q+W)(R+F),\; (Q+W)^\top (Q+W)=I_{k}$ is the QR decomposition of $A$, with 
\[
\|W\|_F \leq \frac{3\kappa(A)\frac{\|E\|_F}{\|A\|_{\op}} }{1-2\kappa(A)\frac{\|E\|_F}{\|A\|_{\op}}},
\]
where $\kappa(A):=\|A\|_{\op} \|A^\dagger \|_{\op}$.
\end{lemma}

The next Lemma controls the one-step difference for Oja's iteration. Note that, we adopt the same notation as introduced above Lemma~\ref{lemma: Factor subspace -- Our}.

\begin{lemma}[One-step Difference Bound for Online PCA iteration]
\label{lemma:oja_onestep_Qdiff}
Consider the update \eqref{eq:oja_update_align} and \eqref{eq: align-update}. Assume that $A_t$ is symmetric with $\|A_t\|_{\op}\le M_A$ a.s. , and $\eta_t^{\rm oja} < \frac{1}{4(\sqrt2+1)M_A}\,k^{-2}$. 
Then for all $t\ge1$,
\begin{equation}\label{eq:Qt_Qtm1_bound}
\|Q_t - Q_{t-1}\|_F  \leq 6\eta_t^{\rm oja} k^{\frac{1}{2}}M_A, \quad  \|Q_t'-Q_{t-1}\|_F \leq 6\eta_t^{\rm oja} k^{\frac{1}{2}}M_A
\end{equation}
The same upper bounds also hold for $\|\cdot \|_{\op}$-norm since $\|\cdot\|_{\op} \leq \| \cdot\|_F$. Moreover, the same argument applies to the standard Oja iteration
\eqref{eq:oja_update}.
\end{lemma}
\begin{proof}
It suffices to consider the update \eqref{eq:oja_update_align} and \eqref{eq: align-update} and prove the perturbation bound for $Q_t' -Q_{t-1}$, since $\|Q_t - Q_{t-1}\|_F \leq \|Q_t'-Q_{t-1}\|_F$.
    Note that $Q_{t-1} = Q_{t-1} I_{k}$, is the Q-factor of QR decomposition itself. We first verify the condition of Lemma \ref{Lemma: QR-Perturbation --- Q-Factor} with $A=Q_{t-1}, Q=Q_{t-1}, R=I_{k}, E = \eta_t^{\rm oja}A_t Q_{t-1}$ and $W=Q'_t-Q_{t-1}$. Now the linear operator $T$ can be written as
    \[
    T(F) = R^\top F + F^\top R = F + F^\top,
    \]

    Then, by definition,
\begin{align}
\label{eq: F1 definition}
F_{1}
&= T^{-1}\!\left[
Q_{t-1}^\top E + E^\top Q_{t-1}
\right] \nonumber \\
&= \eta_t^{\rm oja}\,
T^{-1}\!\left[
Q_{t-1}^\top A_t Q_{t-1}
+
\big(Q_{t-1}^\top A_t Q_{t-1}\big)^\top
\right]
\end{align}
Let
\[
S := Q_{t-1}^\top A_t Q_{t-1},
\qquad
B := S + S^\top.
\]
Restricting $F$ to be upper triangular, $T$ admits the closed-form inverse $T^{-1}(B) = \operatorname{triu}(B)-\frac{1}{2}\operatorname{diag}(B)$ for any symmetric $B$, i.e.
    \[
(T^{-1}B)_{ij} =
\begin{cases}
B_{ij}, & i < j, \\[4pt]
\dfrac{1}{2} B_{ii}, & i = j, \\[6pt]
0, & i > j .
\end{cases}
\]
Then we have
\begin{equation}
    \|T^{-1}B\|_F^2
=
\sum_{i<j} B_{ij}^2 + \frac14 \sum_i B_{ii}^2
\le \frac12 \|B\|_F^2 \leq \|S\|_F^2 + \|S^\top\|_F^2
\le 2k\,\|A_t\|_{\mathrm{op}}^2, \nonumber
\end{equation}
Consequently,  
\begin{equation}
\label{eq: T-1B F norm bound}
    \|F_1\|_F = \eta_t^{\rm oja} \|T^{-1}B\|_F \leq \sqrt{2}\eta_t^{\rm oja} k^{\frac{1}{2}}\|A_t\|_{\op} \leq \sqrt{2} M_A k^{\frac{1}{2}} \eta_t^{\rm oja}
\end{equation}
By combining \eqref{eq: F1 definition}, \eqref{eq: T-1B F norm bound} and under the assumption $\eta_t^{\rm oja}
<
\frac{1}{4(\sqrt2+1)M_A}\,k^{-2}$, we obtain that
\[
\tau \|F_{1}\|_F =k^{\frac{3}{2}}\!\left(1+\frac{1}{\sqrt2}\right) \|F_1\|_F < \frac{1}{4},
\]
where $\tau$ is defined in Lemma \ref{Lemma: QR-Perturbation --- Q-Factor}.
Additionally, it follows that 
\[
\|A^\dagger\|_F \|E\|_F \leq \eta_t^{\rm oja} k M_A < \frac{1}{4(\sqrt{2}+1)}k^{-1} < \frac{1}{2}
\]

Now we apply Lemma \ref{Lemma: QR-Perturbation --- Q-Factor} to get   
\[
\|Q_{t}'-Q_{t-1}\|_F =\|W\|_F\leq \frac{3\kappa(A)\frac{\|E\|_F}{\|A\|_{\op}} }{1-2\kappa(A)\frac{\|E\|_F}{\|A\|_{\op}}} \leq 6 \eta_t^{\rm oja} k^{\frac{1}{2}} M_A,
\]
where the last inequality holds after plugging in the definitions \[
\kappa(A) = 1, \quad \|A\|_{\op}:=\|Q_{t-1}\|_{\op}=1, \quad \|E\|_F\leq\eta_t^{\rm oja} k^{\frac{1}{2}} M_A  < \frac{1}{2}
\]
Lastly, since $Q_t$ satisfies \eqref{eq: align-update}, 
it immediately follows that
\[
\|Q_t - Q_{t-1}\|_F \leq \|Q_t'-Q_{t-1}\|_F \leq 6\eta_t^{\rm oja} k^{\frac{1}{2}}M_A
\]

\end{proof}

By combining Lemma \ref{lemma: theta_star_rotation_stability}, \ref{lemma: Lipthitz of Rotation R}, and Lemma \ref{lemma:oja_onestep_Qdiff}, we are now able to prove a complete statement of proposition \ref{prop: Stability of theta* in Oja's Update}.

\begin{proposition}[Stability of \(\theta^*\) under Oja's update]
\label{prop: Stability of theta* in Oja's Update -- app version}
Suppose Assumptions \ref{asp: ojs_assump},
\ref{asp:Strongly Convex}, \ref{asp:Lipschitz Continuity for f} and  \ref{asp:ojas_step_size} hold.
Fix a finite horizon \(T\) with $\log T = \Ocal(\log d)$, and choose the Stage-II Oja parameter
\begin{equation}
    \label{eq: beta condition without T in Prop}
    \beta
\ge
C\bar\rho_k^{-2}
\left[
q\log d+\log\left(\frac{k}{\bar\rho_k^2}\right)
\right], \qquad \alpha \geq 8
\end{equation}
Then, Lemma \ref{lemma: Lq moment bound for oja update} gives that
\[
\mathbb P(G_T^c)\le d^{-q/2}T^{-q},
\]
where Oja good event \(G_T\) is defined in \eqref{eq: good events}.
For every \(1\le t\le T\), it holds that
\[
\Big\|\,\big|\theta_t^*-\theta_{t-1}^*\big|_s\,\Big\|_q
\le
C\frac{L_{s,f}}{\mu}M_f
\left(
M_Ak^{1/2}\eta_t^{\rm oja}
+
d^{-1/2}T^{-1}
\right),
\]
where \(M_f:=\||f|_2\|_q\).
In particular, if \(\eta_t^{\rm oja}\le c_k/t\) for some $c_k$ independent of $t$, then
\[
\Big\|\,\big|\theta_t^*-\theta_{t-1}^*\big|_s\,\Big\|_q
\le
C\frac{L_{s,f}}{\mu}M_f(M_Ak^{1/2}c_k+d^{-1/2}) t^{-1}.
\]
\end{proposition}

\begin{proof} 
Let $\E_z[ \cdot]$ denote the expectation taken only w.r.t the random vectors $(f,y)$. Recall that $\theta^*_{t}$ is the minimizer of $\EE_z \Big[ l(\theta; R_{t}f, y) \big| R_{t} \Big]$. Lemma \ref{lemma: theta_star_rotation_stability} provides the following inequality
\begin{equation}
\label{eq: theta difference bound: R,f-1--prop}
    \Big\|\,\big|\theta_{t}^*-\theta_{t-1}^*\big|_s\,\Big\|_q
\;\le\;
\frac{L_{s,f}}{\mu}\;
\Big\|\,\big|(R_{t}-R_{t-1})f\big|_s\,\Big\|_q 
\leq 
\frac{L_{s,f}}{\mu}  \Big\| \| R_{t}-R_{t-1} \|_{\op} \Big\|_q \Big\| |f|_2 \Big\|_q
\end{equation}
On the good event $G_T$, where $G_T$ defined in \eqref{eq: good events}, we have 
\[
\|W_t\|_{\op}\le \kappa,\qquad
\|W_{t-1}\|_{\op}\le \kappa, \qquad \kappa:=\sqrt{2e}.
\]
Therefore, by Lemma \ref{lemma: Lipthitz of Rotation R}, 
\[
\|R_t-R_{t-1}\|_{\op}
\le
C_\kappa \|Q_t-Q_{t-1}\|_{\op}.
\]

Followed by Lemma \ref{lemma:oja_onestep_Qdiff}, it holds on event $G_T$ that
\begin{equation}
\nonumber
     \| R_{t}-R_{t-1} \|_{\op}  \leq C M_A k^{\frac{1}{2}} \eta_{t}^{oja},
\end{equation}
where $C$ is an absolute constant.

On \(G_T^c\), since \(R_t,R_{t-1}\in\mathbb O_k\),
\[
\|R_t-R_{t-1}\|_{\op}\le 2.
\]
Therefore,
\[
\|R_t-R_{t-1}\|_{\op}
\le
CM_Ak^{1/2}\eta_t^{\rm oja}
+
2 \cdot \mathbf 1_{G_T^c}.
\]
Taking \(L_q\)-norm gives
\[
\Big\|\|R_t-R_{t-1}\|_{\op}\Big\|_q
\le
CM_Ak^{1/2}\eta_t^{\rm oja}
+
2\mathbb P(G_T^c)^{1/q}.
\]
By Lemma \ref{lemma: Lq moment bound for oja update}, we have
\begin{equation}
    \mathbb P(G_t^c)
\le
k C^2 \alpha^2\bar\rho_k^{-2}
\exp\left\{
-\frac{\beta\bar\rho_k^2}{C^2\alpha^2}
\right\}.
\end{equation}
By taking 
\begin{equation}
\label{eq: beta condition without T in proof}
    \beta
\ge
C\bar\rho_k^{-2}
\left[
2q\log d+\log\left(\frac{k}{\bar\rho_k^2}\right)
\right]
\ge
C\bar\rho_k^{-2}
\left[
q\log d+q\log T+\log\left(\frac{k}{\bar\rho_k^2}\right)
\right]
\end{equation}
We have
\[
\mathbb P(G_T^c)^{1/q}
\le
d^{-1/2}T^{-1}.
\]
Hence
\begin{equation}
\label{eq: Rt one step op, q moment bound full}
    \Big\|\|R_t-R_{t-1}\|_{\op}\Big\|_q
\le
CM_Ak^{1/2}\eta_t^{\rm oja}
+
Cd^{-1/2}T^{-1}.
\end{equation}
Combine with \eqref{eq: Rt one step op, q moment bound full} and \eqref{eq: theta difference bound: R,f-1--prop}, we obtain
\[
    \Big\|\,\big|\theta_{t}^*-\theta_{t-1}^*\big|_s\,\Big\|_q
\;\le\; C\frac{L_{s,f}}{\mu}M_f
\left(
M_Ak^{1/2}\eta_t^{\rm oja}
+
d^{-1/2}T^{-1}
\right),
\]

If \(\eta_t^{\rm oja}\le c_k/t\), then for \(1\le t\le T\),
\[
d^{-1/2}T^{-1} \le d^{-1/2} t^{-1}.
\]
Therefore
\[
M_Ak^{1/2}\eta_t^{\rm oja}
+
d^{-1/2}T^{-1}
\le
(M_Ak^{1/2}c_k+d^{-1/2})t^{-1}.
\]
which proves that second claim. By absorbing the absolute constant in \eqref{eq: beta condition without T in proof}, we obtain the same statement as \eqref{eq: beta condition without T in Prop}
\end{proof}
\begin{remark}
    When we set $\eta_t^{oja} = \frac{\alpha}{\rho_k (\beta+t)}$, we can choose $c_k = \frac{\alpha}{\rho_k}$.
\end{remark}

\subsection{Proofs for Theorem \ref{thm:Moment_Convergence_of_SGD} in  Section \ref{sec: FSGD}}
\label{subsec: App Proofs for Section Sec FSGD}

\begin{lemma}[Factor Estimation Error in $L_q$-norm]
\label{lemma: Factor Estimation Error}
    Suppose Assumption \ref{asp: factor model regularity} and Lemma \ref{lemma: Lq moment bound for oja update} holds. Set $\eta_t^{\rm oja}
=
\frac{\alpha}{(\beta+t)\rho_k}$. Define the factor estimator for a new observation $x$ by
\(
\hat f := d^{-1/2} Q_t^\top x,
\)
where $Q_t$ denotes the output of the $t$-th iteration of Oja's algorithm. 
Let $R \in \mathbb{O}_k$ be an orthogonal matrix satisfying the conclusion of 
Proposition~\ref{prop: Factor Estimation}. Then for $q\geq 2, s \geq 2$, we have
\begin{equation}
\label{eq:moment_fhat_error_bound}
\begin{aligned}
    \Big\|\,|\hat f-Rf|_s\,\Big\|_q
    \leq\;&
    \sqrt{2}\,k^{1/2+1/q}
    \Big\|\,|f|_2\,\Big\|_q
    \bigg[
    \left(\frac{\beta+1}{\beta+t}\right)^{\alpha/2}
    +
    C\frac{\alpha q^{1/2}}{\bar\rho_k}
    \frac{t^{1/2}}{\beta+t}  \\
    &\qquad\qquad\qquad\qquad
    +
    C^{2/q}\alpha^{2/q}\bar\rho_k^{-2/q}
    \exp\left\{
    -\frac{\beta\bar\rho_k^2}{\alpha^2 C^2q}
    \right\}
    \bigg] \\
    &+
    \left(
    C\frac{K_u^2}{\lambda_k(\Sigma_f)}
    \Big\|\,|f|_2\,\Big\|_q
    +
    K_u\big(c k^{1/2}+Cq^{1/2}\big)
    \right)d^{-1/2},
\end{aligned}
\end{equation}
where $C>0$ is absolute constant. Furthermore, by taking $\beta \ge
 \Theta\Big( \bar \rho_k^{-2} \big( q\log d+ \log \frac{k}{\bar \rho_k^2} \big) \Big),$ and absorbing the absolute constant, we have
 \begin{equation}
 \label{eq:moment_fhat_error_bound--short}
     \begin{aligned}
         \Big\|\,|\hat f-Rf|_s\,\Big\|_q
\leq\;&
Ck^{1/2+1/q}
\Big\|\,|f|_2\,\Big\|_q
\left[
\left(\frac{\beta+1}{\beta+t}\right)^{\alpha/2}
+
\frac{ q^{1/2}}{\bar\rho_k}
\frac{t^{1/2}}{\beta+t}
\right] \\
&+
C\left(
\frac{K_u^2}{\lambda_k(\Sigma_f)}
\Big\|\,|f|_2\,\Big\|_q
+
K_u(k^{1/2}+q^{1/2})
+
k^{1/2}\Big\|\,|f|_2\,\Big\|_q
\right)d^{-1/2}.
     \end{aligned}
 \end{equation}

\end{lemma}

\begin{proof}

By Lemma \ref{lemma: Lq moment bound for oja update}, we obtain
\begin{equation}
\label{eq: dist(Q_T,V) -- T}
\left\|\operatorname{dist}(Q_t,V)\right\|_q
\le
k^{1/q}
\left[
\left(\frac{\beta+1}{\beta+t}\right)^{\alpha/2}
+
C\cdot\frac{\alpha q^{1/2}}{\bar\rho_k}\cdot
\frac{t^{1/2}}{\beta+t}
+
C^{2/q} \cdot \alpha^{2/q}\bar\rho_k^{-2/q}
\exp\left\{
-\frac{\beta\bar\rho_k^2}{\alpha^2 C^2q}
\right\}
\right].
\end{equation}

Plug \eqref{eq: dist(Q_T,V) -- T} into Proposition \ref{prop: Factor Estimation} and absorb absolute constant, we obtain
\begin{equation}
\begin{aligned}
    \Big\|\,|\hat f-Rf|_s\,\Big\|_q
    \leq\;&
    \sqrt{2}\,k^{1/2+1/q}
    \Big\|\,|f|_2\,\Big\|_q
    \bigg[
    \left(\frac{\beta+1}{\beta+t}\right)^{\alpha/2}
    +
    C\frac{\alpha q^{1/2}}{\bar\rho_k}
    \frac{t^{1/2}}{\beta+t}  \\
    &\qquad\qquad\qquad\qquad
    +
    C^{2/q}\alpha^{2/q}\bar\rho_k^{-2/q}
    \exp\left\{
    -\frac{\beta\bar\rho_k^2}{\alpha^2 C^2q}
    \right\}
    \bigg] \\
    &+
    \left(
    C\frac{K_u^2}{\lambda_k(\Sigma_f)}
    \Big\|\,|f|_2\,\Big\|_q
    +
    K_u\big(c k^{1/2}+Cq^{1/2}\big)
    \right)d^{-1/2},
\end{aligned}
\end{equation}
By taking
\[
\beta \ge
\frac{\alpha^2 C^2 q}{\bar\rho_k^2}
\left[
\frac{1}{2}\log d
+
\frac{1}{q}
\log\left(
\frac{k C^2 \alpha^2}{\bar\rho_k^2}
\right)
\right]
=
\Theta\left(
\bar\rho_k^{-2}
\left(
q\log d+\log\frac{k}{\bar\rho_k^2}
\right)
\right),
\]
where \(\alpha\) and \(C\) are absolute constants, we have
\[
k^{1/q} C^{2/q}\alpha^{2/q}\bar\rho_k^{-2/q}
\exp\left\{
-\frac{\beta\bar\rho_k^2}{\alpha^2 C^2q}
\right\}
\le d^{-1/2}.
\]
Therefore, the exponential residual term in
Lemma~\ref{lemma: Lq moment bound for oja update} can be absorbed into the
\(d^{-1/2}\) factor-estimation error term

\[
\begin{aligned}
\Big\|\,|\hat f-Rf|_s\,\Big\|_q
\leq\;&
\sqrt{2}\,k^{1/2+1/q}
\Big\|\,|f|_2\,\Big\|_q
\left[
\left(\frac{\beta+1}{\beta+t}\right)^{\alpha/2}
+
C\frac{\alpha q^{1/2}}{\bar\rho_k}
\frac{t^{1/2}}{\beta+t}
\right] \\
&+
C
\left(
\frac{K_u^2}{\lambda_k(\Sigma_f)}
\Big\|\,|f|_2\,\Big\|_q
+
K_u(k^{1/2}+q^{1/2})
+
k^{1/2}\Big\|\,|f|_2\,\Big\|_q
\right)d^{-1/2}.
\end{aligned}
\]
\end{proof}

To simplify the term $(\frac{\beta+1}{\beta+t})^{\alpha/2}$, we provide the following lemma.
\begin{lemma}
\label{lem:decay-beta-alpha-half}
Let $\beta\ge 0$ and $\alpha\ge 2$. Then, for all $t\ge 1$,
\[
    \left(\frac{\beta+1}{\beta+t}\right)^{\alpha/2}
    \le C_{\alpha,\beta} t^{-1/2},
\]
where
\begin{equation}
    \label{eq: def of C alpha beta}
    C_{\alpha,\beta}
=
\begin{cases}
1,
& 0\le \beta \le \alpha-1, \\[6pt]
\dfrac{(\beta+1)^{\alpha/2}\beta^{(1-\alpha)/2}
(\alpha-1)^{(\alpha-1)/2}}
{\alpha^{\alpha/2}},
& \beta>\alpha-1.
\end{cases}
\end{equation}
Moreover, this is the smallest possible constant for the inequality to hold
uniformly over $t\ge 1$. In particular, when $\alpha=8$,
\[
C_{8,\beta}
=
\begin{cases}
1,
& 0\le \beta\le 7, \\[6pt]
\dfrac{7^{7/2}}{8^4}
\dfrac{(\beta+1)^4}{\beta^{7/2}},
& \beta>7.
\end{cases}
\]
\end{lemma}

\begin{proof}
It suffices to find the smallest constant $C_{\alpha,\beta}$ such that
\[
    \left(\frac{\beta+1}{\beta+t}\right)^{\alpha/2}
    \le C_{\alpha,\beta} t^{-1/2}
\]
for all $t\ge 1$. Equivalently,
\[
    C_{\alpha,\beta}
    =
    \sup_{t\ge 1}
    t^{1/2}
    \left(\frac{\beta+1}{\beta+t}\right)^{\alpha/2}.
\]
Define
\[
    f(t)
    =
    t^{1/2}
    \left(\frac{\beta+1}{\beta+t}\right)^{\alpha/2},
    \qquad t\ge 1.
\]
Then
\[
    \log f(t)
    =
    \frac12\log t
    +
    \frac{\alpha}{2}\log(\beta+1)
    -
    \frac{\alpha}{2}\log(\beta+t).
\]
Therefore,
\[
    \frac{d}{dt}\log f(t)
    =
    \frac{1}{2t}
    -
    \frac{\alpha}{2(\beta+t)}
    =
    \frac{\beta-(\alpha-1)t}{2t(\beta+t)}.
\]

If $0\le \beta\le \alpha-1$, then for all $t\ge 1$,
\[
    \beta-(\alpha-1)t
    \le
    \beta-(\alpha-1)
    \le 0.
\]
Hence $f(t)$ is non-increasing on $[1,\infty)$. Thus
\[
    \sup_{t\ge 1} f(t)
    =
    f(1)
    =
    1.
\]
Therefore,
\[
    C_{\alpha,\beta}=1.
\]

If $\beta>\alpha-1$, then the derivative changes sign at
\[
    t_*=\frac{\beta}{\alpha-1}>1.
\]
Thus $f(t)$ is increasing on $[1,t_*]$ and decreasing on
$[t_*,\infty)$. Hence
\[
    \sup_{t\ge 1} f(t)=f(t_*).
\]
A direct calculation gives
\[
\begin{aligned}
    f(t_*)
    &=
    \left(\frac{\beta}{\alpha-1}\right)^{1/2}
    \left(
        \frac{\beta+1}
        {\beta+\frac{\beta}{\alpha-1}}
    \right)^{\alpha/2} =
    \frac{(\beta+1)^{\alpha/2}\beta^{(1-\alpha)/2}
    (\alpha-1)^{(\alpha-1)/2}}
    {\alpha^{\alpha/2}}.
\end{aligned}
\]
Therefore,
\[
    C_{\alpha,\beta}
    =
    \frac{(\beta+1)^{\alpha/2}\beta^{(1-\alpha)/2}
    (\alpha-1)^{(\alpha-1)/2}}
    {\alpha^{\alpha/2}}.
\]
Combining the two cases proves the desired inequality. 
\end{proof}

\begin{lemma}[Moment decay of $\nabla_\theta l$]
\label{lemma:Randomness of nabla_l}
Let \(q\ge 2\) and \(s\ge 2\). For i.i.d. samples
\((f_i,y_i),\, i=1,\ldots,m\), suppose
\[
    \big\|\,|\nabla_\theta l(\theta;f_i,y_i)|_q\,\big\|_q < \infty .
\]
Then there exists a constant \(C_{s,q}>0\), independent of \(m\), such that
\[
\Big\|
\Big|
\frac{1}{m}\sum_{i=1}^m
\Big(
\nabla_\theta l(\theta;f_i,y_i)
-
\mathbb E_z[\nabla_\theta l(\theta;f_i,y_i)]
\Big)
\Big|_s
\Big\|_q
\le
C_{s,q}m^{-1/2}.
\]
Moreover, one may take
\[
C_{s,q}
=
C_q
\left(
\sum_{j=1}^p
\left[
(\mathbb E X_{1,j}^2)^{1/2}
+
(\mathbb E |X_{1,j}|^q)^{1/q}
\right]^r
\right)^{1/r},
\qquad r:=q\wedge s,
\]
where
\[
X_1 :=
\nabla_\theta l(\theta;f_1,y_1)
-
\mathbb E_z[\nabla_\theta l(\theta;f_1,y_1)].
\]
\end{lemma}

\begin{proof}
For simplicity, define the \(\mathbb R^p\)-valued random vectors
\[
X_i :=
\nabla_\theta l(\theta;f_i,y_i)
-
\mathbb E_z[\nabla_\theta l(\theta;f_i,y_i)],
\qquad
S_m:=\frac{1}{m}\sum_{i=1}^m X_i .
\]
Let
\(r:=q\wedge s.\)
Since \(r\le s\), we have \(|v|_s\le |v|_r\) for all \(v\in\mathbb R^p\).
Therefore,
\[
\big\||S_m|_s\big\|_q
\le
\big\||S_m|_r\big\|_q .
\]
Moreover, since \(r\le q\), we have \(q/r\ge 1\). Hence, by Minkowski's
inequality applied with exponent \(q/r\),
\[
\begin{aligned}
\big\||S_m|_r\big\|_q
&=
\left\|
\left(\sum_{j=1}^p |S_{m,j}|^r\right)^{1/r}
\right\|_q =
\left\|
\sum_{j=1}^p |S_{m,j}|^r
\right\|_{q/r}^{1/r} \\
&\le
\left(
\sum_{j=1}^p
\big\||S_{m,j}|^r\big\|_{q/r}
\right)^{1/r} =
\left(
\sum_{j=1}^p
\|S_{m,j}\|_q^r
\right)^{1/r}.
\end{aligned}
\]
For each \(j\in\{1,\ldots,p\}\), the random variables
\(\{X_{i,j}\}_{i=1}^m\) are i.i.d. and mean-zero. By Rosenthal's inequality (Lemma \ref{lemma:Rosenthal Inequality}),
there exists a constant \(C_q>0\), depending only on \(q\), such that
\[
\begin{aligned}
\left\|\sum_{i=1}^m X_{i,j}\right\|_q^q
&\le
C_q
\left[
\sum_{i=1}^m \mathbb E|X_{i,j}|^q
+
\left(\sum_{i=1}^m \mathbb E X_{i,j}^2\right)^{q/2}
\right] =
C_q
\left[
m\mathbb E|X_{1,j}|^q
+
\left(m\mathbb E X_{1,j}^2\right)^{q/2}
\right].
\end{aligned}
\]
Taking the \(q\)-th root and dividing by \(m\), and absorbing constants into
\(C_q\), we obtain
\[
\|S_{m,j}\|_q
\le
C_q
\left[
m^{-1+1/q}(\mathbb E|X_{1,j}|^q)^{1/q}
+
m^{-1/2}(\mathbb E X_{1,j}^2)^{1/2}
\right].
\]
Since \(q\ge 2\), \(m^{-1+1/q}\le m^{-1/2}\). Hence
\[
\|S_{m,j}\|_q
\le
C_q m^{-1/2}
\left[
(\mathbb E X_{1,j}^2)^{1/2}
+
(\mathbb E|X_{1,j}|^q)^{1/q}
\right].
\]
Substituting this coordinate-wise bound into the previous display gives
\[
\begin{aligned}
\big\||S_m|_s\big\|_q
&\le
\left(
\sum_{j=1}^p
\|S_{m,j}\|_q^r
\right)^{1/r}  \leq
C_q m^{-1/2}
\left(
\sum_{j=1}^p
\left[
(\mathbb E X_{1,j}^2)^{1/2}
+
(\mathbb E|X_{1,j}|^q)^{1/q}
\right]^r
\right)^{1/r}.
\end{aligned}
\]
Define
\[
C_{s,q}
:=
C_q
\left(
\sum_{j=1}^p
\left[
(\mathbb E X_{1,j}^2)^{1/2}
+
(\mathbb E |X_{1,j}|^q)^{1/q}
\right]^r
\right)^{1/r},
\qquad r=q\wedge s.
\]
The assumed \(q\)-th moment condition implies \(C_{s,q}<\infty\). Therefore,
\[
\big\||S_m|_s\big\|_q
\le
C_{s,q}m^{-1/2},
\]
which completes the proof.
\end{proof}

We are now in position to prove Theorem \ref{thm:Moment_Convergence_of_SGD}.

\begin{proof}[Proof of Theorem \ref{thm:Moment_Convergence_of_SGD}]
Let $\mathcal{F}_{t-1}$ denote the $\sigma$-field generated by all random variables revealed up to iterations $0,1,2,\ldots,t-1$. We denote by $\mathbb{E}_z[\cdot]$ the expectation taken only with respect to the random vectors $(f_{t,i}, y_{t,i})$, which are independent of $\Fcal_{t-1}$. Note that in the proof we introduce constants with slightly different notation from that used in the statement of Theorem~\ref{thm:Moment_Convergence_of_SGD}, for clarity.

\noindent\textbf{Step 1.}
Recall 
\[
\theta_{t} = \theta_{t-1} - \eta_{t-1} \nabla_\theta \frac{1}{m}\sum_{i=1}^m l(\theta_{t-1};\hat{f}_{t,i},y_{t,i}),
\]
where $\hat{f}_{t,i} = d^{-\frac{1}{2}}\hat{Q}_{t-1}^\top x_{t,i}$.
It holds that
\begin{align*}
    \Big \| | \theta_t - \theta^*_{t-1} |_s \Big \|_q^2 
    &= \Big \| | \theta_{t-1} - \eta_{t-1}  \nabla_\theta \frac{1}{m}\sum_{i=1}^m l(\theta_{t-1};\hat{f}_{t,i},y_{t,i}) - \theta^*_{t-1} |_s \Big \|_q^2 \\
    &= \Big \| | \theta_{t-1} - 
    \theta^*_{t-1} - \eta_{t-1}  \nabla_\theta \EE_z [ l(\theta_{t-1};R_{t-1}f_{t,i},y_{t,i})] \\
    & \quad \quad+ \eta_{t-1}  \big( \nabla_\theta \EE_z [ l(\theta_{t-1};R_{t-1}f_{t,i},y_{t,i})] -
      \nabla_\theta \frac{1}{m}\sum_{i=1}^m l(\theta_{t-1};\hat{f}_{t,i},y_{t,i})
      \big) |_s \Big \|_q^2
\end{align*}
Applying Lemma \ref{lemma: lemma2}, it follows that
\begin{align}
    \Big \| | \theta_t - \theta^*_{t-1} |_s \Big \|_q^2  &
    \label{eq:theta convergence to theta^*, decompostion, term1-1}
    \leq  \Big \| | \theta_{t-1} - 
    \theta^*_{t-1} - \eta_{t-1}  \nabla_\theta \EE_z [ l(\theta_{t-1};R_{t-1}f_{t,i},y_{t,i})] |_s \Big \|_q^2  \\
    &
    \label{eq:theta convergence to theta^*, decompostion, term2-1}
    \qquad + \eta_{t-1}^2\cdot(\max \{q,s\}-1)\cdot  \Big\| | \nabla_\theta \EE_z [ l(\theta_{t-1};R_{t-1}f_{t,i},y_{t,i})] -
      \nabla_\theta \frac{1}{m}\sum_{i=1}^m l(\theta_{t-1};\hat{f}_{t,i},y_{t,i})
       |_s \Big \|_q^2 \\
    &
    \label{eq:theta convergence to theta^*, decompostion, term3-1}
    \qquad + 2 \Big\| |w|_s \Big \|_q^{2-q} \EE \big( |w|_s^{q-s} \langle w^{\odot (s-1)},v \rangle \big),
\end{align}
where $w:=\theta_{t-1} - 
    \theta^*_{t-1} - \eta_{t-1}  \nabla_\theta \EE_z [ l(\theta_{t-1};R_{t-1}f_{t,i},y_{t,i})]$, and $v:=\eta_{t-1}  \big( \nabla_\theta \EE_z [ l(\theta_{t-1};R_{t-1}f_{t,i},y_{t,i})] -
      \nabla_\theta \frac{1}{m}\sum_{i=1}^m l(\theta_{t-1};\hat{f}_{t,i},y_{t,i})
      \big)$.

\noindent\textbf{Step 2.}
Now we control the first term on the RHS of Equation \eqref{eq:theta convergence to theta^*, decompostion, term1-1}. Note that $\nabla_\theta \EE_z[l(\theta^*_{t-1}; R_{t-1}f_{t,i}, y_{t,i})] = 0$, by applying Lemma \ref{lemma: lemma2} again, we have
\begin{equation}
\label{eq:term1-control-1}
\begin{aligned}
    &\Big \| | \theta_{t-1} - 
    \theta^*_{t-1}  - \eta_{t-1}  \nabla_\theta \EE_z [ l(\theta_{t-1};R_{t-1}f_{t,i},y_{t,i})] |_s \Big \|_q^2  \\
    &= \Big \| | \theta_{t-1} - 
    \theta^*_{t-1}  - \eta_{t-1} \Big( \nabla_\theta \EE_z [ l(\theta_{t-1};R_{t-1}f_{t,i},y_{t,i})]  -\nabla_\theta \EE_z[l(\theta^*_{t-1};R_{t-1}f_{t,i},y_{t,i})]\Big) |_s \Big \|_q^2  \\
    & \leq \Big \| | \theta_{t-1} - 
    \theta^*_{t-1} |_s \Big \|_q^2 + \eta_{t-1}^2 (\max\{q,s \}-1) \Big \| |\nabla_\theta \EE_z [ l(\theta_{t-1};R_{t-1}f_{t,i},y_{t,i})] -\nabla_\theta \EE_z[l(\theta^*_{t-1};R_{t-1}f_{t,i},y_{t,i})]  |_s \Big\|_q^2
    \\
     &\qquad -2\eta_{t-1}\Big \| | \theta_{t-1} - 
    \theta^*_{t-1} |_s \Big \|_q^{2-q} \EE \Big[ |\theta_{t-1} - \theta^*_{t-1}|_s^{q-s} \cdot \\
    &\qquad\quad\langle (\theta_{t-1}-\theta^*_{t-1})^{\odot(s-1)},\EE_z [\nabla_\theta  l(\theta_{t-1};R_{t-1}f_{t,i},y_{t,i})-\nabla_\theta l(\theta^*_{t-1};R_{t-1}f_{t,i},y_{t,i})] \rangle \Big]
\end{aligned}
\end{equation}

Due to Assumption \ref{asp:Lipschitz Continuity for theta}, it yields
\begin{align}
\label{eq:term1-lip_theta}
    \Big \| | \EE_z \big[ \nabla_\theta  l(\theta_{t-1};R_{t-1}f_{t,i},y_{t,i})] -\nabla_\theta l(\theta^*_{t-1};R_{t-1}f_{t,i},y_{t,i}) \big]  |_s \Big\|_q^2 \leq  \Big \| L_{s,\theta} |\theta_{t-1} - \theta^*_{t-1}|_s \Big \|_q^2
\end{align}

Along with Assumption \ref{asp:Strongly Convex}, we can bound
\begin{align}
\label{eq:term1-strong_convex}
    \langle (\theta_{t-1}-\theta^*_{t-1})^{\odot(s-1)},\EE_z [\nabla_\theta  l(\theta_{t-1};R_{t-1}f_{t,i},y_{t,i})-\nabla_\theta l(\theta^*_{t-1};R_{t-1}f_{t,i},y_{t,i})] \rangle \geq \mu|\theta_{t-1}-\theta^*_{t-1}|_s^s
\end{align}

Inserting inequalities \eqref{eq:term1-lip_theta}, \eqref{eq:term1-strong_convex} into Equation \eqref{eq:term1-control-1}, we obtain

\begin{equation}
\begin{aligned}
\label{eq:bound for || |x|_s ||_q-1}
    &\Big \| | \theta_{t-1} - 
    \theta^*_{t-1}  - \eta_{t-1}  \nabla_\theta \EE_z [ l(\theta_{t-1};R_{t-1}f_{t,i},y_{t,i})] |_s \Big \|_q^2  \\
    &\leq \Big \| | \theta_{t-1} - 
    \theta^*_{t-1} |_s \Big \|_q^2+ \eta_{t-1}^2 (\max\{q,s \}-1)L_{s,\theta}^2 \Big\| | \theta_{t-1} - \theta^*_{t-1} |_s \Big \|_q^2 -2\eta_{t-1}\mu\Big\| | \theta_{t-1} - \theta^*_{t-1} |_s \Big \|_q^2 \\
    &=\big( 1-2\mu \eta_{t-1} + \eta_{t-1}^2 (\max\{q,s \}-1)L_{s,\theta}^2  \big) \Big\| | \theta_{t-1} - \theta^*_{t-1} |_s \Big \|_q^2
\end{aligned}
\end{equation}

\noindent\textbf{Step 3.} Now we control the term on the RHS of Equation \eqref{eq:theta convergence to theta^*, decompostion, term2-1}.
\begin{equation}
\label{eq:term2-control}
\begin{aligned}
     &\Big\| | \nabla_\theta \EE_z [ l(\theta_{t-1};R_{t-1}f_{t,i},y_{t,i})] -
      \nabla_\theta \frac{1}{m}\sum_{i=1}^m l(\theta_{t-1};\hat{f}_{t,i},y_{t,i})
       |_s \Big \|_q \\
    &\leq \underbrace{\Big\| | \nabla_\theta \EE_z [ l(\theta_{t-1};R_{t-1}f_{t,i},y_{t,i})] -
      \nabla_\theta \EE_z [ l(\theta^*_{t-1};R_{t-1}f_{t,i},y_{t,i})]
       |_s \Big \|_q  }_{(i)}  \\
       &\qquad + \underbrace{\Big\| | \nabla_\theta \frac{1}{m}\sum_{i=1}^m l(\theta^*_{t-1};R_{t-1}f_{t,i},y_{t,i}) -
      \nabla_\theta \EE_z [ l(\theta^*_{t-1};R_{t-1}f_{t,i},y_{t,i})]
       |_s \Big \|_q }_{(ii)} \\
       &\qquad + \underbrace{\Big\| 
      | \nabla_\theta \frac{1}{m}\sum_{i=1}^m l(\theta^*_{t-1};\hat{f}_{t,i},y_{t,i})
      -\nabla_\theta \frac{1}{m}\sum_{i=1}^m l(\theta^*_{t-1};R_{t-1}f_{t,i},y_{t,i})
       |_s \Big \|_q  }_{(iii)}\\
       &\qquad + \underbrace{\Big\| | \nabla_\theta \frac{1}{m}\sum_{i=1}^m l(\theta^*_{t-1};\hat{f}_{t,i},y_{t,i}) -
      \nabla_\theta \frac{1}{m}\sum_{i=1}^m l(\theta_{t-1};\hat{f}_{t,i},y_{t,i})
       |_s \Big \|_q  }_{(iv)}.
\end{aligned}
\end{equation}

Rely on Assumption \ref{asp:Lipschitz Continuity for theta}, we control $(i),(iv)$ as
\begin{equation}
\label{eq:term2-control-term1,4}
    (i) + (iv) \leq 2L_{s,\theta} \Big\|  |\theta_{t-1} - \theta^*_{t-1}|_s \Big\|_q
\end{equation}

The upper bound of $(ii)$ can be derived using Lemma \ref{lemma:Randomness of nabla_l}:
\begin{equation}
\label{eq:term2-control-term2}
    (ii)=\Big\| |  \frac{1}{m}\sum_{i=1}^m \big ( \nabla_\theta l(\theta^*_{t-1};R_{t-1}f_{t,i},y_{t,i}) -
      \EE_z [ \nabla_\theta l(\theta^*_{t-1};R_{t-1}f_{t,i},y_{t,i})] \big)
       |_s \Big \|_q \leq C_{s,q} m^{-\frac{1}{2}}
\end{equation}

$(iii)$ depends on the estimation of $\hat{f}_i^{t}$. Along with Assumption \ref{asp:Lipschitz Continuity for f}, we obtain
\begin{equation}
\label{eq:term2-control-term3}
\begin{aligned}
    (iii) = \Big\| 
      \nabla_\theta \frac{1}{m}\sum_{i=1}^m l(\theta^*_{t-1};\hat{f}_{t,i},y_{t,i})
      -\nabla_\theta \frac{1}{m}\sum_{i=1}^m l(\theta^*_{t-1};R_{t-1}f_{t,i},y_{t,i})
       |_s \Big \|_q \leq L_{s,f} \Big \| |\hat{f}_{t,i}-R_{t-1}f_{t,i}|_{s} \Big\|_q
\end{aligned}
\end{equation}

Combining Equations \eqref{eq:term2-control-term1,4}, \eqref{eq:term2-control-term2}, \eqref{eq:term2-control-term3} and \eqref{eq:term2-control}, it holds that
\begin{align*}
    & \Big\| | \nabla_\theta \EE_z [ l(\theta_{t-1};R_{t-1}f_{t,i},y_{t,i})] -
      \nabla_\theta \frac{1}{m}\sum_{i=1}^m l(\theta_{t-1};\hat{f}_{t,i},y_{t,i})
       |_s \Big \|_q^2 \\
    &\leq 4 \big(2L_{s,\theta}^2 \Big\|  |\theta_{t-1} - \theta^*_{t-1}|_s \Big\|^2_q + C_{s,q}^2 m^{-1}+ L_{s,f}^2 \Big \| |\hat{f}_{t,i}-R_{t-1} f_{t,i}|_{s} \Big\|^2_q \big)
\end{align*}

An upper bound of \eqref{eq:theta convergence to theta^*, decompostion, term2-1} follows immediately,
\begin{align*}
    &\eta_{t-1}^2\cdot(\max \{q,s\}-1)\cdot  \Big\| | \nabla_\theta \EE_z [ l(\theta_{t-1};R_{t-1} f_{t,i},y_{t,i})] -
      \nabla_\theta \frac{1}{m}\sum_{i=1}^m l(\theta_{t-1};\hat{f}_{t,i},y_{t,i})
       |_s \Big \|_q^2 \\
    &\leq 4 \eta_{t-1}^2\cdot(\max \{q,s\}-1)\cdot\big(2L_{s,\theta}^2 \Big\|  |\theta_{t-1} - \theta^*_{t-1}|_s \Big\|^2_q + C_{s,q}^2 m^{-1}+ L_{s,f}^2 \Big \| |\hat{f}_{t,i}-R_{t-1}f_{t,i}|_{s} \Big\|^2_q \big)
\end{align*}

\textbf{Step 4:} Now we control the term on the RHS of \eqref{eq:theta convergence to theta^*, decompostion, term3-1}. Note that $w=\theta_{t-1} - 
    \theta_{t-1}^* - \eta_{t-1}  \nabla_\theta \EE_z [ l(\theta_{t-1};R_{t-1}f_{t,i},y_{t,i})]\in \Fcal_{t-1}$, we can condition on $\Fcal_{t-1}$ and rewrite the equation as
\begin{align}
    \Big\| |w|_s \Big \|_q^{2-q} \EE \big[ |w|_s^{q-s} \langle w^{\odot (s-1)},v \rangle \big] = \Big\| |w|_s \Big \|_q^{2-q} \EE \Big[ |w|_s^{q-s} \langle w^{\odot (s-1)},\EE(v|\Fcal_{t-1}) \rangle \Big]
\end{align}
By H\"older inequality and Jensen Inequality, 
\[
| \langle w^{\odot(s-1)},\EE(v|\Fcal_{t-1}) \rangle| \leq |w^{\odot(s-1)}|_{s^*} \cdot |\EE(v|\Fcal_{t-1})|_s \leq |w|_s^{s-1} \cdot |\EE(v|\Fcal_{t-1})|_s,
\]
where $s^* = \frac{s}{s-1}$ satisfying $\frac{1}{s^*}+\frac{1}{s}=1$. Consequently,
\begin{equation}
    \label{eq:theta convergence to theta^*-rewrite-term3}
    \Big\| |w|_s \Big \|_q^{2-q} \EE \big[ |w|_s^{q-s} \langle w^{\odot (s-1)},v \rangle \big] \leq  \Big\| |w|_s \Big \|_q^{2-q} \EE \Big[ |w|_s^{q-1} \cdot |\EE(v|\Fcal_{t-1})|_s \Big]
\end{equation}
Again, we apply H\"older inequality, $\EE |ab| \leq \EE^{1/q^*}|a|^{q^*} \cdot \EE^{1/q}|b|^{q}$, with $q^*=\frac{q}{q-1}$ to get
\begin{equation}
\label{eq: interaction term, transformation}
\EE \Big[ |w|_s^{q-1} \cdot |\EE(v|\Fcal_{t-1})|_s \Big] 
\leq \EE^{\frac{q-1}{q}} \Big[ |w|_s^{q} \Big] \cdot \EE^{\frac{1}{q}}\Big[ |\EE(v|\Fcal_{t-1})|_s^q \Big]
=   \Big\| |w|_s \Big \|_q^{q-1} \Big\| |\EE(v|\Fcal_{t-1})|_s  \Big \|_q
\end{equation}
Combining \eqref{eq:theta convergence to theta^*-rewrite-term3} and \eqref{eq: interaction term, transformation}, we actually show that if $w \in \Fcal_{t-1}$, then
\begin{equation}
    \label{eq: Lemma 2 interation term upper bound}
    \Big\| |w|_s \Big \|_q^{2-q} \EE \big[ |w|_s^{q-s} \langle w^{\odot (s-1)},v \rangle \big] \leq \Big\| |w|_s \Big \|_q \cdot \Big\| |\EE(v|\Fcal_{t-1})|_s  \Big \|_q
\end{equation}

We focus on $\EE[v \, |\, \Fcal_{t-1}]$ first. Substitute the definition of $v$, we have
\begin{align}
    \EE[v \, |\, \Fcal_{t-1}] &= \eta_{t-1} \cdot \EE \Big[ \EE_z\big[ \nabla l(\theta_{t-1}; R_{t-1}f_{t,i},y_{t,i})\big] \Big| \Fcal_{t-1} \Big] - \eta_{t-1} \cdot \EE \Big[ \nabla l(\theta_{t-1}; \hat{f}_{t,i},y_{t,i}) | \Fcal_{t-1} \Big]  \nonumber \\
    &=\eta_{t-1} \cdot \Big( \EE_z\big[ \nabla l(\theta_{t-1}; R_{t-1}f_{t,i},y_{t,i})\big]  - \EE_{z}[ \nabla l(\theta_{t-1}; \hat{f}_{t,i},y_{t,i}) ] \Big) \nonumber \\
    &
    \label{eq:E[y|F_t-1] transformation1}
    =\eta_{t-1} \cdot \EE_z\Big[ \nabla l(\theta_{t-1}; R_{t-1}f_{t,i},y_{t,i})  - \nabla l(\theta_{t-1}; \hat{f}_{t,i},y_{t,i})   \Big],
\end{align}
where the second equality is due to $\theta_{t-1},R_{t-1} \in \Fcal_{t-1}$. Taking $l_s$-norm and $L_q$-norm of random variable on both side of \eqref{eq:E[y|F_t-1] transformation1}, we have
\begin{align*}
    \Big\| |\EE(v|\Fcal_{t-1})|_s\Big\|_q &\leq \eta_{t-1} \cdot \EE^{\frac{1}{q}} \Big[ \EE_z^q\big( |\nabla l(\theta_{t-1}; R_{t-1}f_{t,i},y_{t,i})  - \nabla l(\theta_{t-1}; \hat{f}_{t,i},y_{t,i}) |_s \big) \Big]\\
    &\leq \eta_{t-1} \cdot \EE^{\frac{1}{q}} \Big[ \EE_z\big( |\nabla l(\theta_{t-1}; R_{t-1}f_{t,i},y_{t,i})  - \nabla l(\theta_{t-1}; \hat{f}_{t,i},y_{t,i}) |_s^q \big) \Big]\\
    &= \eta_{t-1} \cdot \Big\| | \nabla l(\theta_{t-1}; R_{t-1}f_{t,i},y_{t,i})  - \nabla l(\theta_{t-1}; \hat{f}_{t,i},y_{t,i})|_s \Big\|_q \\
    &\leq \eta_{t-1}  L_{s,f} \Big\| |R_{t-1}f_{t,i}-\hat{f}_{t,i}|_s \Big\|_q,
\end{align*}
where the second inequality holds by applying Jensen's inequality since $q\geq2$.

Replacing this upper bound into inequality \eqref{eq: interaction term, transformation}, we have
\[
 \Big\| |w|_s \Big \|_q^{2-q} \EE \big[ |w|_s^{q-s} \langle w^{\odot (s-1)},v \rangle \big] \leq \eta_{t-1} \Big\| |w|_s \Big \|_q  L_{s,f} \Big\| |R_{t-1}f_{t,i}-\hat{f}_{t,i}|_s\Big\|_q
\]

Note that $w=\theta_{t-1} - 
    \theta^*_{t-1} - \eta_{t-1}  \nabla_\theta \EE_z [ l(\theta_{t-1};R_{t-1}f_{t,i},y_{t,i})]$, Equation \eqref{eq:bound for || |x|_s ||_q-1} already provides a bound for $\Big\| |w|_s \Big \|_q$, i.e.
\begin{equation}
\nonumber
    \Big\| |w|_s \Big \|_q \leq \big( 1-2\mu \eta_{t-1} + \eta_{t-1}^2 (\max\{q,s \}-1)L_{s,\theta}^2  \big)^{\frac{1}{2}} \Big\| | \theta_{t-1} - \theta^*_{t-1} |_s \Big \|_q
\end{equation}

 The term on the RHS of \eqref{eq:theta convergence to theta^*, decompostion, term3-1} can then be controlled by
 \begin{equation}
     2\big( 1-2\mu \eta_{t-1} + \eta_{t-1}^2 (\max\{q,s \}-1)L_{s,\theta}^2  \big)^{\frac{1}{2}} \eta_{t-1}  L_{s,f} \, \Big\| |R_{t-1}f_{t,i}-\hat{f}_{t,i}|_s\Big\|_q \Big\| | \theta_{t-1} - \theta^*_{t-1} |_s \Big \|_q
 \end{equation}

We denote $\alpha_{t-1}:=\big( 1-2\mu \eta_{t-1} + \eta_{t-1}^2 (\max\{q,s \}-1)L_{s,\theta}^2  \big)^{\frac{1}{2}} $ and $C_1^2:=\max \{q,s\}-1$ also for simplicity of notation.  Lemma \ref{lemma: Factor Estimation Error} provides the factor estimation error for the output of the $t$-th iteration of Oja's algorithm, i.e.,
 \begin{equation}
     \begin{aligned}
         \Big\| |R_{t-1}f_{t,i}-\hat{f}_{t,i}|_s\Big\|_q 
\leq \;&
Ck^{1/2+1/q}
\Big\|\,|f|_2\,\Big\|_q
\left[
\left(\frac{\beta+1}{\beta+t}\right)^{\alpha/2}
+
\frac{ q^{1/2}}{\bar\rho_k}
\frac{t^{1/2}}{\beta+t}
\right] \\
&+
C\left(
\frac{K_u^2}{\lambda_k(\Sigma_f)}
\Big\|\,|f|_2\,\Big\|_q
+
K_u(k^{1/2}+q^{1/2})
+
k^{1/2}\Big\|\,|f|_2\,\Big\|_q
\right)d^{-1/2}.
     \end{aligned}
 \end{equation}
Note that $\frac{t^{1/2}}{\beta+t}\leq t^{-1/2}$ and applying Lemma \ref{lem:decay-beta-alpha-half}, it simplifies to 
 \begin{equation}
     \begin{aligned}
         \Big\| |R_{t-1}f_{t,i}-\hat{f}_{t,i}|_s\Big\|_q 
\leq \;&
Ck^{1/2}
\Big\|\,|f|_2\,\Big\|_q (C_{\alpha,\beta}+\frac{q^{1/2}}{\bar \rho_k}) t^{-\frac{1}{2}}\\
&+
C\left(
\frac{K_u^2}{\lambda_k(\Sigma_f)}
\Big\|\,|f|_2\,\Big\|_q
+
K_u(k^{1/2}+q^{1/2})
+
k^{1/2}\Big\|\,|f|_2\,\Big\|_q
\right)d^{-1/2}.
     \end{aligned}
 \end{equation}
Equivalently,
\begin{equation}
    \Big\| |R_{t-1}f_{t,i}-\hat{f}_{t,i}|_s\Big\|_q 
\leq 
C_2 M_f \, t^{-\frac{1}{2}} + C_d,
\end{equation}
where 
\begin{align}
    &C_2:=Ck^{1/2} (C_{\alpha,\beta}+\frac{q^{1/2}}{\bar \rho_k}) , \quad M_f:=\Big\|\,|f|_2\,\Big\|_{q}, \nonumber \\
    & C_d:=C\left(
\frac{K_u^2}{\lambda_k(\Sigma_f)}
 M_f
+
K_u(k^{1/2}+q^{1/2})
+
k^{1/2} M_f
\right)d^{-1/2}. \nonumber
\end{align}
and $C$ is absolute constants.

To conclude the results in Steps 2,3,4, we obtain that $ \Big \| | \theta_t - \theta^*_{t-1} |_s \Big \|_q^2$ admits the following bound:
\begin{equation}
    \begin{aligned}
    \label{eq: pre recursive bound-1}
    \Big \| | \theta_t - \theta^*_{t-1} |_s \Big \|_q^2  &
    \leq  
    \alpha_{t-1}^2 \Big\| | \theta_{t-1} - \theta^*_{t-1} |_s \Big \|_q^2   +  
    4 \eta_{t-1}^2C_1^2\cdot \Big( 2L_{s,\theta}^2 \Big\|  |\theta_{t-1} - \theta^*_{t-1}|_s \Big\|^2_q  \\
    & \qquad + 
    C_{s,q}^2 m^{-1}+ L_{s,f}^2 (C_2 M_f \, t^{-\frac{1}{2}} + C_d)^2 \Big)  \\
    &\quad + 2\alpha_{t-1}   \eta_{t-1}  L_{s,f} \,(C_2 M_f \, t^{-\frac{1}{2}} + C_d) \Big\| | \theta_{t-1} - \theta^*_{t-1} |_s \Big \|_q,
\end{aligned}
\end{equation}

\textbf{Step 5:} Now we have derived that $\Big\| | \theta_{t} - \theta^*_{t-1} |_s \Big \|_q$ can be bounded by a function of $\Big\| | \theta_{t-1} - \theta^*_{t-1} |_s \Big \|_q$. In this step, we consider the minimizer shift $\Big\| | \theta^*_{t-1} - \theta^*_{t-2} |_s \Big \|_q$.

Apply Proposition \ref{prop: Stability of theta* in Oja's Update -- app version}, we obtain that
\begin{equation}
\label{eq: shift difference-1}
    \Big\|\,\big|\theta_{t-1}^*-\theta_{t-2}^*\big|_s\,\Big\|_q
\;\le\;
C\frac{L_{s,f}}{\mu}M_f(M_Ak^{1/2}c_k+d^{-1/2})  t^{-1},
\end{equation}
where $C$ is an absolute constant, $M_f:=\Big\| |f|_2 \Big\|_q$, and $c_k = \frac{\alpha}{\rho_k}$ since $\eta_t^{oja} = \frac{\alpha}{\rho_k (\beta+t)}$.

Apply Lemma \ref{lemma: lemma2} to $\Big\| | \theta_{t-1} - \theta^*_{t-1} |_s \Big \|_q = \Big\| | (\theta_{t-1} - \theta^*_{t-2}) + (\theta_{t-2}^*-\theta_{t-1}^*) |_s \Big \|_q$ with $w = \theta_{t-1} - \theta^*_{t-2}$ and $v =\theta_{t-2}^*-\theta_{t-1}^* $, it holds that
\begin{align}
\label{eq: shift bound}
    \nonumber
    \Big\| | \theta_{t-1} - \theta^*_{t-1} |_s \Big \|_q^2 &\leq \Big\| | w |_s \Big \|_q^2 + 2 \Big\| |w|_s \Big \|_q \Big\| |v|_s  \Big \|_q + (\max \{q,s\}-1)\cdot \Big\| |v|_s  \Big \|_q^2 \\
    &\leq  \Big\| |w|_s \Big \|_q^2 + 2 \, C_4\, \Big\| |w|_s \Big \|_q  t^{-1}  + C_1^2C_4^2\cdot t^{-2},
\end{align}
where $C_4 :=C\frac{L_{s,f}}{\mu}M_f(M_Ak^{1/2}\rho_k^{-1}+d^{-1/2})  $.

\textbf{Step 5.1: Derive the Recursion}

To abbreviate our notation, we denote
\[
F_{t-1}:=C_2M_f t^{-\frac{1}{2}}+C_3d^{-1/2}, \quad 
\Xi^2_{t-1}:=C_{s,q}^2 m^{-1}+ L_{s,f}^2 (C_2 M_f \, t^{-\frac{1}{2}} + C_d)^2,
\]
where $C_3:=C\left(
\frac{K_u^2}{\lambda_k(\Sigma_f)}
 M_f
+
K_u(k^{1/2}+q^{1/2})
+
k^{1/2} M_f
\right)$.
Rewrite shift of minimizer satisfies (as in \eqref{eq: shift difference-1}) as
\begin{equation}\label{eq:ass_shift_v}
\Big\|\,|\theta^*_{t-2}-\theta^*_{t-1}|_s\,\Big\|_{q}
\;\le\; \kappa_{t-1},
\qquad
\kappa_{t-1}:=C_4 t^{-1},
\end{equation}
and that Lemma \ref{lemma: lemma2} implies the inequality \eqref{eq: shift bound} 
\begin{equation}\label{eq:shift_bound_compact}
U^2_{t-1} := \Big\|\,|\theta_{t-1}-\theta^*_{t-1}|_s\,\Big\|_{q}^2
\;\le\;
X^2_{t-1}+2\kappa_{t-1}\,X_{t-1}+C_1^2 \kappa_{t-1}^2,
\qquad t\ge 2,
\end{equation}
where $X_{t-1}:=\|\,|\theta_{t-1}-\theta^*_{t-2}|_s\|_{L_q}^2$. Note that $C_1 \geq 1$ and 
\[
X^2_{t-1}+2\kappa_{t-1}\,X_{t-1}+C_1^2 \kappa_{t-1}^2 \leq (X_{t-1}+C_1 \kappa_{t-1})^2,
\]
we have
\begin{equation}\label{eq:U_expand}
U_{t-1} \le X_{t-1}+ C_1 \kappa_{t-1},
\end{equation}
Then the bound \eqref{eq: pre recursive bound-1} implies, for all $t\ge 2$,
\begin{align}
X^2_t
&\le
\alpha_{t-1}^2 \,U^2_{t-1}
+
8\eta_{t-1}^2 C_1^2 L_{s,\theta}^2\,U^2_{t-1}
+
4\eta_{t-1}^2C_1^2\,\Xi_{t-1}^2
+
2\alpha_{t-1}\eta_{t-1}L_{s,f}F_{t-1}\,U_{t-1},
\label{eq:Zt_pre}
\end{align}
Substituting \eqref{eq:shift_bound_compact} and \eqref{eq:U_expand} into \eqref{eq:Zt_pre} yields
\begin{align}
\label{eq:Zt_after_subst-1}
X^2_t
&\le
\underbrace{\big(\alpha_{t-1}^2+8\eta_{t-1}^2 C_1^2L_{s,\theta}^2\big)}_{=:\beta^2_{t-1}}
\Big(X_{t-1}^2+2\kappa_{t-1} X_{t-1}+C_1^2\kappa_{t-1}^2\Big)
+4\eta_{t-1}^2C_1^2\Xi_{t-1}^2 \nonumber\\
&\quad
+2\alpha_{t-1}\eta_{t-1}L_{s,f}F_{t-1}\,(X_{t-1}+C_1\kappa_{t-1}),
\end{align}
where $\beta^2_{t-1}:=1-2\mu \eta_{t-1} + 9\eta_{t-1}^2 (\max\{q,s \}-1)L_{s,\theta}^2 \leq 1$, by assumption.

Now, we control the cross terms.

\emph{(i) The term $2 \cdot \kappa_{t-1}\beta_{t-1}^2 \cdot X_{t-1}$.}
For any $\varepsilon>0$, Young's inequality gives $2\kappa B\le \varepsilon B^2+\varepsilon^{-1}\kappa^2$.
Hence
\begin{equation}\label{eq:young1}
2 \cdot \kappa_{t-1}\beta_{t-1}^2 \cdot X_{t-1}
\le 
\varepsilon_{t-1} X_{t-1}^2 + \varepsilon_{t-1}^{-1}\kappa_{t-1}^2.
\end{equation}

\emph{(ii) The term $2 \cdot\alpha_{t-1}L_{s,f}F_{t-1}\eta_{t-1}\cdot X_{t-1}$.}
Again, for any $\varepsilon>0$,
\begin{equation}\label{eq:young2-1}
2 \cdot\alpha_{t-1}L_{s,f}F_{t-1}\eta_{t-1}\cdot X_{t-1}
\le
\varepsilon_{t-1} X_{t-1}^2
+\varepsilon_{t-1}^{-1}L_{s,f}^2F_{t-1}^2 \eta_{t-1}^2
\end{equation}

\emph{(iii) The term $2 \cdot\alpha_{t-1}C_1L_{s,f}\kappa_{t-1}\eta_{t-1}\cdot F_{t-1}$.}
Again, for any $\varepsilon>0$,
\begin{equation}\label{eq:young3}
2 \cdot\alpha_{t-1}C_1L_{s,f}\kappa_{t-1}\eta_{t-1}\cdot F_{t-1}
\le
\varepsilon_{t-1} F_{t-1}^2
+\varepsilon_{t-1}^{-1} C_1^2L_{s,f}^2\kappa_{t-1}^2\eta_{t-1}^2
\end{equation}

We set $\varepsilon_{t-1}=\frac{\mu}{8}\eta_{t-1}$ in \eqref{eq:young1}, \eqref{eq:young2-1}. With this choice, the coefficient of $X_t^2$ satisfies
\begin{equation}
\label{eq:a_contraction}
    \beta^2_{t-1} + 2 \varepsilon_{t-1} = 1-2\mu \eta_{t-1} + 9\eta_{t-1}^2 (\max\{q,s \}-1)L_{s,\theta}^2 + \frac{\mu}{4}\eta_{t-1}
\leq 
1- \frac{\mu}{2}\eta_{t-1} 
\end{equation}

Also, we set $\varepsilon_{t-1} = \eta_{t-1}$ in \eqref{eq:young3}, so that 
\begin{equation}
2 \cdot\alpha_{t-1}C_1L_{s,f}\kappa_{t-1}\eta_{t-1}\cdot F_{t-1}
\le
\eta_{t-1} F_{t-1}^2
+C_1^2L_{s,f}^2\kappa_{t-1}^2\eta_{t-1},
\end{equation}
which the factor error $F_{t-1}$ can be absorbed by previous terms.

Now combine \eqref{eq:Zt_after_subst-1} with \eqref{eq:young1}, \eqref{eq:young2-1},
and  using $\alpha_{t-1} \le \beta_{t-1}\le 1$,
we obtain
\begin{equation}
\begin{aligned}
\label{eq:rec_small-1}
    X^2_t
\le&
\Big(1-\frac{\mu}{2}\eta_{t-1}\Big)X^2_{t-1}
+
\frac{8}{\mu}\eta^{-1}_{t-1}\kappa_{t-1}^2
+C_1^2 \kappa_{t-1}^2  \\
&+4C_1^2\eta_{t-1}^2\Xi_{t-1}^2
+(\frac{8}{\mu}L_{s,f}^2+1) \eta_{t-1}F_{t-1}^2
+C_1^2L_{s,f}^2\kappa_{t-1}^2\eta_{t-1}. 
\end{aligned}
\end{equation}

\textbf{Step 5.2: Solving the recursion.}
We consider the recursion of the form
\begin{equation}
\label{eq:gronwall_form-1}
Z_t \le \Big(1-\frac{\mu}{2}\eta_{t-1}\Big)Z_{t-1} + r_{t-1},
\end{equation}
where $Z_t=X_t^2$ and $r_{t-1}$ is the corresponding remainder term.
Let $S_t:=\sum_{j=0}^{t-1}\eta_j$. Multiplying \eqref{eq:gronwall_form-1} by
$\exp\big(\frac{\mu}{2}S_t\big)$ and using that
$\exp(x)\big(1-x)\le 1$ for $x = \frac{\mu}{2}\eta_{t-1} > 0$,
we obtain
\[
\exp\!\Big(\frac{\mu}{2}S_t\Big)Z_t
\le
\exp\!\Big(\frac{\mu}{2}S_{t-1}\Big)Z_{t-1}
+
\exp\!\Big(\frac{\mu}{2}S_t\Big)r_{t-1}.
\]
Iterating from $0$ to $t$ yields
\[
\exp\Big( \frac{\mu}{2}S_t \Big) Z_t \leq Z_0 + \sum_{k=0}^{t-1} \exp\Big( \frac{\mu}{2}S_{k+1} \Big) \, r_k
\]
Reorganizing terms yields
\begin{equation}
\label{eq:weighted_gronwall}
Z_t
\le
\exp\!\Big(-\frac{\mu}{2}S_t\Big)Z_0
+
\sum_{k=0}^{t-1}
\exp\!\Big(-\frac{\mu}{2}(S_t-S_{k+1})\Big)\,r_k.
\end{equation}

\textbf{Step 5.3: Bounding the remainder.}

For \eqref{eq:rec_small-1}, we apply \eqref{eq:weighted_gronwall} with $r_{t-1}=
\frac{8}{\mu}\eta^{-1}_{t-1}\kappa_{t-1}^2
+C_1^2 \kappa_{t-1}^2 
+4C_1^2\eta_{t-1}^2\Xi_{t-1}^2
+(\frac{8}{\mu}L_{s,f}^2+1) \eta_{t-1}F_{t-1}^2
+C_1^2L_{s,f}^2\kappa_{t-1}^2\eta_{t-1}. 
$
Then, it holds that
\begin{equation}
\begin{aligned}
    X_t^2 &\leq \exp\!\Big(-\frac{\mu}{2}S_t\Big)X_0^2 
    + \frac{8}{\mu}C_4^2 \sum_{k=0}^{t-1}\eta_k^{-1}\,\frac{1}{(k+1)^2}\,\exp\!\Big(-\frac{\mu}{2}(S_t-S_{k+1})\Big) \\
    & \quad + C_1^2C_4^2 \sum_{k=0}^{t-1}\frac{1}{(k+1)^2}\,\exp\!\Big(-\frac{\mu}{2}(S_t-S_{k+1}) \Big) \\
    &\quad + 4C_1^2(C_{s,q}^2m^{-1}+2L_{s,f}^2C_d^2) \sum_{k=0}^{t-1}\eta_k^2\,\exp\!\Big(-\frac{\mu}{2}(S_t-S_{k+1}) \Big)
     \\
    &\quad + 8C_1^2L_{s,f}^2C_2^2M_f^2 \sum_{k=0}^{t-1}\eta_k^{2}\,\frac{1}{k+1}\,\exp\!\Big(-\frac{\mu}{2}(S_t-S_{k+1})\Big)  \\
    &\quad + \big(2(\frac{8}{\mu}L_{s,f}^2+1)C_2^2M_f^2\big)  \sum_{k=0}^{t-1}\eta_k\,\frac{1}{k+1}\,\exp\!\Big(-\frac{\mu}{2}(S_t-S_{k+1})\Big)  \\
    &\quad + 2(\frac{8}{\mu}L_{s,f}^2+1)C_d^2  \sum_{k=0}^{t-1}\eta_k
    \exp\!\Big(-\frac{\mu}{2}(S_t-S_{k+1})\Big) \\
    &\quad + L_{s,f}^2C_1^2C_4^2 \sum_{k=0}^{t-1}\eta_k \frac{1}{(k+1)^2}
    \exp\!\Big(-\frac{\mu}{2}(S_t-S_{k+1})\Big)
\end{aligned}
\label{eq: Step5.3}
\end{equation}
Combine with Lemma \ref{lemma:weighted_sum_bounds} and take $\lambda = \frac{\mu}{2}$, we have
\begin{equation}
    \Big \| | \theta_t - \theta^*_{t-1} |_s \Big \|_q^2 \leq \exp\!\Big(-\frac{\mu}{2}S_t\Big)\Big \| | \theta_0 - \theta^* |_s \Big \|_q^2  + \frac{4}{\mu}(\frac{8}{\mu}L_{s,f}^2+1)C_d^2 + o(1)
\end{equation}

Furthermore, we recall the assumption $\eta_t = \Ocal((t+1)^{-\gamma})$ for some $\gamma \in (0,1)$. Then $S_t = \Theta(t^{1-\gamma})$ and Lemma \ref{lemma:Polynomial step size and exponentially weighted sums} gives 
\begin{align}
    X_t^2 &\leq \exp\!\Big(-\frac{\mu}{2}S_t\Big)X_0^2 
    + \frac{2}{\mu} \cdot \Big( \frac{8}{\mu}C_4^2 \Ocal(t^{-(2-2\gamma)}) + C_1^2C_4^2 \Ocal(t^{-(2-\gamma)}) \nonumber \\
    &\quad + 4C_1^2(C_{s,q}^2m^{-1}+2L_{s,f}^2C_d^2)\Ocal(t^{-\gamma})
   + 8C_1^2L_{s,f}^2C_2^2M_f^2 \Ocal(t^{-(1+\gamma)}) \nonumber \\
    &\quad + \big(2(\frac{8}{\mu}L_{s,f}^2+1)C_2^2M_f^2\big)  \Ocal(t^{-1})  + 2(\frac{8}{\mu}L_{s,f}^2+1)C_d^2 \nonumber \\
    &\quad + L_{s,f}^2C_1^2C_4^2\Ocal(t^{-2}) \Big)
\end{align}
Thus,
\begin{equation}
    \begin{aligned}
        \Big \| | \theta_t - \theta^*_{t-1} |_s \Big \|_q^2 &\leq  \frac{4}{\mu}(\frac{8}{\mu}L_{s,f}^2+1)C_d^2 + \frac{8}{\mu}C_1^2(C_{s,q}^2m^{-1}+2L_{s,f}^2C_d^2) \cdot O\!\big(t^{-\gamma}\big) \\
        &\quad + \frac{4}{\mu}(\frac{8}{\mu}L_{s,f}^2+1)C_2^2M_f^2 \cdot \Ocal(t^{-1})+\frac{16}{\mu^2} C_4^2 \cdot \Ocal(t^{-(2-2\gamma)}) + \tilde{\Rcal}_t^2,
    \end{aligned}
\end{equation}
where the remainder term $\tilde{\Rcal}_t^2$ collects all error terms that decay faster in $t$ or grow slower in $d$ than the rates displayed above, i.e.
\begin{equation}
    \label{eq: Rtilde definition}
    \tilde{\Rcal}_t^2:= \exp\!\Big(-\frac{\mu}{2}S_t\Big)X_0^2+ \frac{2}{\mu} \Big(  C_1^2C_4^2 \Ocal(t^{-(2-\gamma)})+ 8C_1^2L_{s,f}^2C_2^2M_f^2 \Ocal(t^{-(1+\gamma)})+ L_{s,f}^2C_1^2C_4^2\Ocal(t^{-2})\Big).
\end{equation}

We keep terms related to $C_2,C_4,C_d$ are they are related to dimension $d$.
Taking square roots on both side, and applying inequality $(\sum a_i)^{\frac{1}{2}} \leq \sum a_i^{\frac{1}{2}}$, we obtain
\begin{equation}
    \begin{aligned}
        \Big \| | \theta_t - \theta^*_{t-1} |_s \Big \|_q &\leq  \sqrt{\frac{4}{\mu}(\frac{8}{\mu}L_{s,f}^2+1)} \, C_d + \sqrt{\frac{8}{\mu}C_1^2(C_{s,q}^2m^{-1}+2L_{s,f}^2C_d^2)} \cdot O\!\big(t^{-\frac{\gamma}{2}}\big) \\
        &\quad + \sqrt{\frac{4}{\mu}(\frac{8}{\mu}L_{s,f}^2+1)} C_2M_f \cdot \Ocal(t^{-\frac{1}{2}})+ \frac{4}{\mu} C_4 \cdot \Ocal(t^{-(1-\gamma)}) + \tilde{\Rcal}_t,
    \end{aligned}
\end{equation}

Substituting the definitions of all constants $C_1, C_2, C_4$ and $C_d$, we obtain
\begin{equation}
    \begin{aligned}
    \Big \| &| \theta_t - \theta^*_{t-1} |_s \Big \|_q \leq \sqrt{\frac{4}{\mu}(\frac{8}{\mu}L_{s,f}^2+1)} \, C\left(
\frac{K_u^2}{\lambda_k(\Sigma_f)}
 M_f
+
K_u(k^{1/2}+q^{1/2})
+
k^{1/2} M_f
\right) \cdot d^{-\frac{1}{2}} \\
&\quad + \sqrt{\frac{8}{\mu} (\max\{q,s \}-1)\Big(C_{s,q}^2m^{-1}+2L_{s,f}^2C^2\big[
\frac{K_u^2}{\lambda_k(\Sigma_f)}
 M_f + K_u(k^{1/2}+q^{1/2}) + k^{1/2} M_f
\big]^2 \cdot d^{-1} \Big)} \cdot O\!\big(t^{-\frac{\gamma}{2}}\big) \\
&\quad + \sqrt{\frac{4}{\mu}(\frac{8}{\mu}L_{s,f}^2+1)} \cdot Ck^{1/2} (C_{\alpha,\beta}+\frac{q^{1/2}}{\bar \rho_k}) \cdot M_f \cdot \Ocal(t^{-\frac{1}{2}}) \\
&\quad + \frac{4}{\mu} \cdot C\frac{L_{s,f}}{\mu}M_f(M_Ak^{1/2}\rho_k^{-1}+d^{-1/2})\cdot \Ocal(t^{-(1-\gamma)}) + \tilde{\Rcal}_t,
    \end{aligned}
\end{equation}
To make the dependence on $d$, $m$, and $t$ explicit, we denote
\begin{equation}
    \begin{aligned}
    \label{eq: Ai definitions}
&A_1:=C\sqrt{\frac{1}{\mu}\left(\frac{1}{\mu}L_{s,f}^2+1\right)}\left(\frac{K_u^2}{\lambda_k(\Sigma_f)}M_f+K_u(k^{1/2}+q^{1/2})+k^{1/2}M_f\right),\\
&A_2:=C\sqrt{\frac{\max\{q,s\}-1}{\mu}}\left(C_{s,q}+L_{s,f}\left[\frac{K_u^2}{\lambda_k(\Sigma_f)}M_f+K_u(k^{1/2}+q^{1/2})+k^{1/2}M_f\right]\right),\\
&A_3:=C\sqrt{\frac{1}{\mu}\left(\frac{1}{\mu}L_{s,f}^2+1\right)}\cdot M_f \, k^{1/2} ,\\
&A_4:=C\frac{L_{s,f}}{\mu^2}M_f,
\end{aligned}
\end{equation}
where $C$ is absolute constant.

Combining the above notations,
we obtain
the stated bound in Theorem~\ref{thm:Moment_Convergence_of_SGD},  i.e.
\begin{align*}
\Big\||\theta_t-\theta_{t-1}^*|_s\Big\|_q&\leq A_1d^{-1/2}+A_2(m^{-1/2}+d^{-1/2})t^{-\gamma/2}\\&+A_3 (C_{\alpha,\beta}+\frac{q^{1/2}}{\bar \rho_k}) \, t^{-1/2}+A_4(M_Ak^{1/2}\rho_k^{-1}+d^{-1/2})t^{-(1-\gamma)}+\widetilde{\mathcal{R}}_t,
\end{align*}

\end{proof}
\begin{remark}
    The terms involving $C_2$ reflect the improvement of subspace distance estimates along the trajectory of the online PCA updates, whereas the terms involving the growing factor $C_4$ arise from the minimizer shift induced by online PCA updates.
\end{remark}


\begin{lemma}
\label{lemma:weighted_sum_bounds}
Let $\{\eta_k\}_{k\ge 0}$ be a non-increasing sequence of nonnegative numbers and define
\(
S_t:=\sum_{j=0}^{t-1}\eta_j,\, t\ge 1.
\)
Fix $\lambda>0$, then for all $t\ge 1$,
\begin{align}
\label{eq:weighted_sum_bounds_lem}
\sum_{k=0}^{t-1}\eta_k
\exp\!\Big(-\lambda(S_t-S_{k+1})\Big)
&\le \frac{1}{\lambda}.
\end{align}
Suppose $\sum_{k=0}^\infty \eta_k=\infty$ and $\eta_k \to 0$, we then have
\begin{align}
\label{eq:weighted_sum_bounds_lem2}
&\sum_{k=0}^{t-1}\eta_k^2
\exp\!\Big(-\lambda(S_t-S_{k+1})\Big)
 \to 0 \text{ , as } t \to \infty.\\
\label{eq:weighted_sum_bounds_lem4}
    &\sum_{k=0}^{t-1}\eta_k \frac{1}{k+1}
\exp\!\Big(-\lambda(S_t-S_{k+1})\Big)
 \to 0 \text{ , as } t \to \infty.
\end{align}
Additionally assuming $(k+1)\eta_k \to \infty$, it then holds that
\begin{equation}
\label{eq:weighted_sum_bounds_lem3}
    \sum_{k=0}^{t-1}\eta_k^{-1} \frac{1}{(k+1)^2}
\exp\!\Big(-\lambda(S_t-S_{k+1})\Big)
 \to 0 \text{ , as } t \to \infty.
\end{equation}
\end{lemma}

\begin{proof}
For any $u\in [S_k,S_{k+1})$, we have $u\le S_{k+1}$, hence
\[
S_t-u \;\ge\; S_t-S_{k+1},
\text{ then} \quad
e^{-\lambda(S_t-S_{k+1})}\;\le\; e^{-\lambda(S_t-u)}.
\]
Consequently
\[
\eta_k\,e^{-\lambda(S_t-S_{k+1})}
=\Big(S_{k+1} - S_k \Big)\,e^{-\lambda(S_t-S_{k+1})} = \int_{S_k}^{S_{k+1}} e^{-\lambda(S_t-S_{k+1})}\,du
\le
\int_{S_k}^{S_{k+1}} e^{-\lambda(S_t-u)}\,du.
\]
Summing over $k=0,\dots,t-1$, it holds that
\[
\sum_{k=0}^{t-1}\eta_k\,e^{-\lambda(S_t-S_{k+1})}
\le
\int_{0}^{S_t} e^{-\lambda(S_t-u)}\,du
=
\int_{0}^{S_t} e^{-\lambda v}\,dv
=
\frac{1-e^{-\lambda S_t}}{\lambda}
\le
\frac{1}{\lambda},
\]
which proves \eqref{eq:weighted_sum_bounds_lem}. 

To show \eqref{eq:weighted_sum_bounds_lem2}, we first fix $K >0$, then
\begin{equation}
\label{eq: sum_divide_2}
    \sum_{k=0}^{t-1}\eta_k^2
\exp\!\Big(-\lambda(S_t-S_{k+1})\Big) = \sum_{k=0}^{K-1}\eta_k^2
\exp\!\Big(-\lambda(S_t-S_{k+1})\Big) 
+ \sum_{k=K}^{t-1}\eta_k^2
\exp\!\Big(-\lambda(S_t-S_{k+1})\Big)
\end{equation}
Since $S_t \to \infty$, the first term on RHS of \eqref{eq: sum_divide_2} goes to $0$ as $t \to \infty$. Note that $\{\eta_k\}$ is non-increasing, for all $k\geq K$ we have $\eta_k^2 \le \eta_k\,\eta_{K}$. Thus followed by \eqref{eq: sum_divide_2}, the second term on the RHS can be controlled as
\[
\sum_{k=K}^{t-1}\eta_k^2
\exp\!\Big(-\lambda(S_t-S_{k+1})\Big) \leq \eta_{K} \sum_{k=K}^{t-1}\eta_k
\exp\!\Big(-\lambda(S_t-S_{k+1})\Big) \leq \frac{\eta_{K}}{\lambda}
\]
We complete the proof of \eqref{eq:weighted_sum_bounds_lem2} holds by taking $K \to \infty$. With the same argument, \eqref{eq:weighted_sum_bounds_lem4} holds since $\frac{1}{K+1} \to 0$.

For \eqref{eq:weighted_sum_bounds_lem3}, similarly, for fixed $K > 0$,
\begin{equation}
\begin{aligned}
    &\sum_{k=0}^{t-1}\eta_k^{-1} \frac{1}{(k+1)^2}
\exp\!\Big(-\lambda(S_t-S_{k+1})\Big) 
\\
& \qquad =
\sum_{k=0}^{K-1}\eta_k^{-1} \frac{1}{(k+1)^2}
\exp\!\Big(-\lambda(S_t-S_{k+1})\Big)
+
\sum_{k=K}^{t-1}\eta_k^{-1} \frac{1}{(k+1)^2}
\exp\!\Big(-\lambda(S_t-S_{k+1})\Big)
\end{aligned}
\end{equation}
The first term goes to $0$ as $t \to \infty$, and the second term
\begin{align}
    \sum_{k=K}^{t-1}\eta_k^{-1} \frac{1}{(k+1)^2}
\exp\!\Big(-\lambda(S_t-S_{k+1})\Big)  &= \sum_{k=K}^{t-1} \frac{1}{\eta_k^2 (k+1)^2}\eta_k\exp\!\Big(-\lambda(S_t-S_{k+1})\Big) \\
&\leq \frac{1}{\eta_K^2 (K+1)^2} \cdot \frac{1}{\lambda}
\end{align}
The last inequality holds since $\eta_k^2 (k+1)^2$ is monotone increasing to infinity. We complete the proof of \eqref{eq:weighted_sum_bounds_lem3} holds by taking $K \to \infty$.

\end{proof}

\begin{lemma}[Polynomial step size and exponentially weighted sums]
\label{lemma:Polynomial step size and exponentially weighted sums}
    Suppose $\eta_k = \Theta((k+1)^{-\gamma})$ for some $0<\gamma<1$, i.e.
    there exist absolute constants $0<a\le b<\infty$ such that for all $k\ge 0$,
\begin{equation}\label{eq:eta_ab_bounds}
a(k+1)^{-\gamma}\le \eta_k\le b(k+1)^{-\gamma}.
\end{equation}
Then, $S_t = \Theta(t^{1-\gamma})$, and the three weighted sums admit the following rates as $t\to\infty$:
\begin{align}
\sum_{k=0}^{t-1}\eta_k^2\,\exp\!\Big(-\lambda(S_t-S_{k+1})\Big)
&=\frac{1}{\lambda} \cdot O\!\big(t^{-\gamma}\big), \label{eq:rate_eta2}\\
\sum_{k=0}^{t-1}\eta_k\,\frac{1}{k+1}\,\exp\!\Big(-\lambda(S_t-S_{k+1})\Big)
&=\frac{1}{\lambda} \cdot O\!\big(t^{-1}\big), \label{eq:rate_eta_over_k}\\
\sum_{k=0}^{t-1}\eta_k^{-1}\,\frac{1}{(k+1)^2}\,\exp\!\Big(-\lambda(S_t-S_{k+1})\Big)
&=\frac{1}{\lambda} \cdot O\!\big(t^{-(2-2\gamma)}\big). \label{eq:rate_inv_eta_over_k}
\end{align}
In particular, all three are $o(1)$ since $\gamma\in(0,1)$.
\end{lemma}
\begin{proof}

Recall
\(
S_t-S_{k+1}=\sum_{j=k+1}^{t-1}\eta_j.
\)
We first derive an uniform lower bound on the tail sum $S_t-S_{k+1}$, by splitting indices into ``early'' $k\le \lfloor t/2\rfloor$ and ``late'' $k>\lfloor t/2\rfloor$.

~\\
\emph{(i) Early indices $0\le k\le \lfloor t/2\rfloor$.}
Then $k+1\le t/2+1$ and
\[
S_t-S_{k+1}\ \ge\ \sum_{j=\lfloor t/2\rfloor}^{t-1}\eta_j
\ \ge\ a\sum_{j=\lfloor t/2\rfloor}^{t-1}(j+1)^{-\gamma}
\ \ge\ a\cdot \frac{t}{2}\cdot t^{-\gamma}
\ =\ \frac{a}{2}\,t^{1-\gamma}.
\]
Hence, for all such $k$,
\begin{equation}\label{eq:early_exp_small}
\exp\!\big(-\lambda(S_t-S_{k+1})\big)\ \le\ \exp\!\Big(-\frac{\lambda a}{2}\,t^{1-\gamma}\Big).
\end{equation}

\emph{(ii) Late indices $\lfloor t/2\rfloor< k\le t-1$.}
For $j\in\{k+1,\dots,t-1\}$ we have $j+1\le t$, hence $(j+1)^{-\gamma}\ge t^{-\gamma}$, and thus
\[
S_t-S_{k+1}
\ \ge\ a\sum_{j=k+1}^{t-1}(j+1)^{-\gamma}
\ \ge\ a\,(t-k-1)\,t^{-\gamma}.
\]
Consequently, for all such $k$,
\begin{equation}\label{eq:late_exp_geo}
\exp\!\big(-\lambda(S_t-S_{k+1})\big)\ \le\ \exp\!\big(-\lambda a (t-k-1)t^{-\gamma}\big).
\end{equation}

We will repeatedly use the elementary bound
\begin{equation}\label{eq:geom_sum_bound}
\sum_{u=0}^{\infty}e^{-c\,u\,t^{-\gamma}}
\ =\ \frac{1}{1-e^{-c t^{-\gamma}}}
\ \le\ \frac{2}{c}\,t^{\gamma}
\qquad \text{for all sufficiently large $t$},
\end{equation}
which follows from $1-e^{-x}\ge x/2$ for small $x>0$. For notation simplicity, we denote $C$ as absolute constants which may differ from line to line.

Now we prove \eqref{eq:rate_eta2}.
Let
\[
A_t := \sum_{k=0}^{t-1}\eta_k^2\,e^{-\lambda(S_t-S_{k+1})}
= \sum_{k=0}^{\lfloor t/2\rfloor}\eta_k^2\,e^{-\lambda(S_t-S_{k+1})} + \sum_{k=\lfloor t/2\rfloor+1}^{t-1}\eta_k^2\,e^{-\lambda(S_t-S_{k+1})}
=: A_t^{\mathrm{early}} + A_t^{\mathrm{late}},
\]
where the split is at $\lfloor t/2\rfloor$.

\emph{Early part.} By \eqref{eq:early_exp_small} and $\eta_k^2\le b^2(k+1)^{-2\gamma}\le b^2$,
\[
A_t^{\mathrm{early}}
\le
\exp\!\Big(-\frac{\lambda a}{2}\,t^{1-\gamma}\Big)\sum_{k=0}^{\lfloor t/2\rfloor}\eta_k^2
\le
b^2 t\exp\!\Big(-\frac{\lambda a}{2}\,t^{1-\gamma}\Big)
= o(t^{-m})\quad \forall m>0.
\]

\emph{Late part.} For $\lfloor t/2\rfloor< k\le t-1$, we have $(k+1)\asymp t$, hence $\eta_k^2\asymp b^2(k+1)^{-2\gamma}\le b^2 t^{-2\gamma}=C t^{-2\gamma}$.
Using \eqref{eq:late_exp_geo} and the change of variables $u:=t-k-1\in\{0,1,\dots\}$,
\[
A_t^{\mathrm{late}}
\le
C t^{-2\gamma}\sum_{k=\lfloor t/2\rfloor+1}^{t-1}e^{-\lambda a (t-k-1)t^{-\gamma}}
=
C t^{-2\gamma}\sum_{u=0}^{t-\lfloor t/2\rfloor-2}e^{-\lambda a u t^{-\gamma}}
\le
C t^{-2\gamma}\sum_{u=0}^{\infty}e^{-\lambda a u t^{-\gamma}}.
\]
Applying \eqref{eq:geom_sum_bound} with $c=\lambda a$ gives
\[
A_t^{\mathrm{late}}\le C t^{-2\gamma}\cdot \frac{2}{c}t^{\gamma}=\frac{2b^2}{\lambda a} t^{-\gamma}.
\]
Combining early and late parts yields $A_t=\frac{1}{\lambda} \cdot O(t^{-\gamma})$, proving \eqref{eq:rate_eta2}.

For \eqref{eq:rate_eta_over_k}, 
let
\[
B_t:=\sum_{k=0}^{t-1}\eta_k\frac{1}{k+1}e^{-\lambda(S_t-S_{k+1})}
= B_t^{\mathrm{early}} + B_t^{\mathrm{late}},
\]
where the split is at $\lfloor t/2\rfloor$.

\emph{Early part.} Using \eqref{eq:early_exp_small} and $\eta_k/(k+1)\le b(k+1)^{-(1+\gamma)}\le b$,
\[
B_t^{\mathrm{early}}
\le
\exp\!\Big(-\frac{\lambda a}{2}\,t^{1-\gamma}\Big)\sum_{k=0}^{\lfloor t/2\rfloor}\eta_k\frac{1}{k+1}
\le
C t \exp\!\Big(-\frac{\lambda a}{2}\,t^{1-\gamma}\Big)
= o(t^{-m})\quad \forall m>0.
\]

\emph{Late part.} For $k>\lfloor t/2\rfloor$ we have $(k+1)\asymp t$, hence
\[
\eta_k\frac{1}{k+1}\le b(k+1)^{-(1+\gamma)}\le C t^{-(1+\gamma)}.
\]
Therefore, by \eqref{eq:late_exp_geo} and \eqref{eq:geom_sum_bound},
\[
B_t^{\mathrm{late}}
\le
C t^{-(1+\gamma)}\sum_{k=\lfloor t/2\rfloor+1}^{t-1}e^{-\lambda a (t-k-1)t^{-\gamma}}
\le
C t^{-(1+\gamma)}\sum_{u=0}^{\infty}e^{-\lambda a u t^{-\gamma}}
\le
\frac{C}{\lambda} t^{-(1+\gamma)}\cdot t^{\gamma}
=
\frac{C}{\lambda} t^{-1}.
\]
Now we obtain \eqref{eq:rate_eta_over_k}.

For \eqref{eq:rate_inv_eta_over_k}, 
let
\[
C_t:=\sum_{k=0}^{t-1}\eta_k^{-1}\frac{1}{(k+1)^2}e^{-\lambda(S_t-S_{k+1})}
= C_t^{\mathrm{early}} + C_t^{\mathrm{late}},
\]
where the split is at $\lfloor t/2\rfloor$.

\emph{Early part.} Using \eqref{eq:early_exp_small} and $\eta_k^{-1}/(k+1)^2\le a^{-1}(k+1)^{\gamma-2}\le a^{-1}$,
\[
C_t^{\mathrm{early}}
\le
\exp\!\Big(-\frac{\lambda a}{2}\,t^{1-\gamma}\Big)\sum_{k=0}^{\lfloor t/2\rfloor}\eta_k^{-1}\frac{1}{(k+1)^2}
\le
C t \exp\!\Big(-\frac{\lambda a}{2}\,t^{1-\gamma}\Big)
= o(t^{-m})\quad \forall m>0.
\]

\emph{Late part.} For $k>\lfloor t/2\rfloor$ we have $(k+1)\asymp t$, hence
\[
\eta_k^{-1}\frac{1}{(k+1)^2}
\le
a^{-1}(k+1)^{\gamma-2}
\le
C t^{\gamma-2}
=
C t^{-(2-\gamma)}.
\]
Therefore, by \eqref{eq:late_exp_geo} and \eqref{eq:geom_sum_bound},
\[
C_t^{\mathrm{late}}
\le
C t^{-(2-\gamma)}
\sum_{k=\lfloor t/2\rfloor+1}^{t-1}e^{-\lambda a (t-k-1)t^{-\gamma}}
\le
C t^{-(2-\gamma)}\sum_{u=0}^{\infty}e^{-\lambda a u t^{-\gamma}}
\le
\frac{C}{\lambda} t^{-(2-\gamma)}\cdot t^{\gamma}
=
\frac{C}{\lambda} t^{-(2-2\gamma)}.
\]
This proves \eqref{eq:rate_inv_eta_over_k}. 
\end{proof}

\subsection{Proofs for Corollary \ref{cor:Moment_Convergence_of_SGD-freeze_Oja} in Section~\ref{sec: FSGD}}

An immediate corollary of Theorem~\ref{thm:Moment_Convergence_of_SGD} 
is that an improved convergence rate can be established when the output of Oja's iteration is frozen after iteration $\lfloor t/2 \rfloor$. 
Note that the same rate of convergence can also be achieved when the factor subspace is estimated using offline PCA.

\begin{proof}[Proof for Corollary \ref{cor:Moment_Convergence_of_SGD-freeze_Oja}]
It suffices to consider the case $t_1=\lfloor t/2 \rfloor$, as the case $t_1=\lfloor t/c_0 \rfloor$ follows from identical arguments.

We keep using the proof constants $C_1, C_2, C_3, C_4$ in the proof of Theorem~\ref{thm:Moment_Convergence_of_SGD}. By following the same proof strategy as Theorem~\ref{thm:Moment_Convergence_of_SGD}, 
we obtain the recursive inequality holds for all $1\leq i \leq t$,
    \begin{equation}
\begin{aligned}
\label{eq:rec_small-2}
    X^2_i
\le&
\Big(1-\frac{\mu}{2}\eta_{i-1}\Big)X^2_{i-1}
+
\frac{8}{\mu}\eta^{-1}_{i-1}\kappa_{i-1}^2
+C_1^2 \kappa_{i-1}^2  \\
&+4C_1^2\eta_{i-1}^2\Xi_{i-1}^2
+(\frac{8}{\mu}L_{s,f}^2+1) \eta_{i-1}F_{i-1}^2
+C_1^2L_{s,f}^2\kappa_{i-1}^2\eta_{i-1}. 
\end{aligned}
\end{equation}

For iterations $i \ge \lfloor \frac{t}{2} \rfloor$, the minimizer shift no longer occurs, so that 
$\theta_{i}^*=\theta^*_{\lfloor \frac{t}{2} \rfloor-1}$. Under this regime, we can rewrite $\kappa,F,\Xi$ as:
For minimizer shift $\kappa_{i} (\geq \Big\| |\theta^*_{i-2} - \theta^*_{i-1}|_s \Big\|_q)$,
\[
\kappa_{i-1} =
\begin{cases}
C_4 i^{-1}, & i \le \lfloor \frac{t}{2} \rfloor, \\[4pt]
0, & i > \lfloor \frac{t}{2} \rfloor .
\end{cases}
\]
For factor estimation error
\[
F_{i-1} =
\begin{cases}
C_2 M_f i^{-1/2} + C_3 d^{-1/2}, & i \le \lfloor \frac{t}{2} \rfloor, \\[6pt]
\sqrt{2}C_2 M_f t^{-1/2} + C_3 d^{-1/2} , & i > \lfloor \frac{t}{2} \rfloor .
\end{cases}
\]
And for $\Xi$ which is related to mini-batch concentration
\[
\Xi_{i-1}^2
= C_{s,q}^2 m^{-1} + L_{s,f}^2 F_{i-1}^2
\]

Now we plug-in the definition of $\kappa,F,\Xi$ and apply \eqref{eq:weighted_gronwall} to solve the recursive bound from $k=0$.

Note that for the summation up to index $t_a :=\lfloor ta \rfloor$, where $\tfrac{1}{2} < a \le 1$, 
we can split the sum at $\lfloor t/2 \rfloor$. 
The early part has the same structure as in the proof of Lemma \ref{lemma:Polynomial step size and exponentially weighted sums}, 
while the late part involves only the terms $F_k$ and $\Xi_k$ since $\kappa_k = 0$. 
Therefore, we obtain

\begin{equation}
    \begin{aligned}
            X_{t_a}^2 &\leq \exp\!\Big(-\frac{\mu}{2}S_{t_a}\Big)X_0^2 
    + \frac{8}{\mu}C_4^2 \sum_{k=0}^{\lfloor\frac{t}{2}\rfloor-1}\eta_k^{-1}\,\frac{1}{(k+1)^2}\,\exp\!\Big(-\frac{\mu}{2}(S_{t_a}-S_{k+1})\Big)  \\
    & \quad + C_1^2C_4^2 \sum_{k=0}^{\lfloor\frac{t}{2}\rfloor-1}\frac{1}{(k+1)^2}\,\exp\!\Big(-\frac{\mu}{2}(S_{t_a}-S_{k+1}) \Big) \\
    &\quad + 4C_1^2(C_{s,q}^2m^{-1}+2L_{s,f}^2C_d^2) \sum_{k=0}^{\lfloor\frac{t}{2}\rfloor-1}\eta_k^2\,\exp\!\Big(-\frac{\mu}{2}(S_{t_a}-S_{k+1}) \Big)
     \\
    &\quad + 8C_1^2L_{s,f}^2C_2^2M_f^2 \sum_{k=0}^{\lfloor\frac{t}{2}\rfloor-1}\eta_k^{2}\,\frac{1}{k+1}\,\exp\!\Big(-\frac{\mu}{2}(S_{t_a}-S_{k+1})\Big)  \\
    &\quad + \big(2(\frac{8}{\mu}L_{s,f}^2+1)C_2^2M_f^2\big)  \sum_{k=0}^{\lfloor\frac{t}{2}\rfloor-1}\eta_k\,\frac{1}{k+1}\,\exp\!\Big(-\frac{\mu}{2}(S_{t_a}-S_{k+1})\Big)  \\
    &\quad + 2(\frac{8}{\mu}L_{s,f}^2+1)C_d^2  \sum_{k=0}^{\lfloor\frac{t}{2}\rfloor-1}\eta_k
    \exp\!\Big(-\frac{\mu}{2}(S_{t_a}-S_{k+1})\Big)\\
    &\quad + L_{s,f}^2C_1^2C_4^2 \sum_{k=0}^{\lfloor\frac{t}{2}\rfloor-1}\eta_k \frac{1}{(k+1)^2}
    \exp\!\Big(-\frac{\mu}{2}(S_{t_a}-S_{k+1})\Big)  \\
    &\quad + 4C_1^2 \sum_{k=\lfloor\frac{t}{2}\rfloor}^{t_a-1}\Xi_k^2\eta_k^2\,\exp\!\Big(-\frac{\mu}{2}(S_{t_a}-S_{k+1}) \Big)
    \\
    &\quad + (\frac{8}{\mu}L_{s,f}^2+1)  \sum_{k=\lfloor\frac{t}{2}\rfloor}^{t_a-1}F^2_k \eta_k
    \exp\!\Big(-\frac{\mu}{2}(S_{t_a}-S_{k+1})\Big)
    \end{aligned}
    \label{eq:cor-step5.3}
\end{equation}
According to the proof of Lemma~\ref{lemma:Polynomial step size and exponentially weighted sums}, 
all early-stage terms are of order 
$\Theta\!\Big(t_a \exp\{-\tfrac{\mu}{4} {t_a}^{1-\gamma}\}\Big)$,
so that decay faster than any polynomial rate. Thus, we could rewrite \eqref{eq:cor-step5.3} as
\begin{equation}
\begin{aligned}
            X_{t_a}^2 &\leq \exp\!\Big(-\frac{\mu}{2}S_{t_a}\Big)X_0^2 + \Theta\!\Big(t_a \exp\{-\tfrac{\mu}{4} {t_a}^{1-\gamma}\}\Big) \\
    &\quad + 4C_1^2 \sum_{k=\lfloor\frac{t}{2}\rfloor}^{t_a-1}\Xi_k^2\eta_k^2\,\exp\!\Big(-\frac{\mu}{2}(S_{t_a}-S_{k+1}) \Big)\\
    &\quad + (\frac{8}{\mu}L_{s,f}^2+1)  \sum_{k=\lfloor\frac{t}{2}\rfloor}^{t_a-1}F^2_k \eta_k
    \exp\!\Big(-\frac{\mu}{2}(S_{t_a}-S_{k+1})\Big) 
\end{aligned}
\label{eq: Xt bound in cor short}
\end{equation}

When $k \geq \lfloor\frac{t}{2} \rfloor$, it holds that
\begin{align}
    &F_{k} =\sqrt{2}C_2 M_f t^{-1/2} + C_3 d^{-1/2} =F_{\lfloor \frac{t}{2} \rfloor} \nonumber \\
    &\Xi_{k}^2 = C_{s,q}^2 m^{-1} + L_{s,f}^2 (\sqrt{2}C_2 M_f t^{-1/2} + C_3 d^{-1/2})^2 =: \Xi_{\lfloor \frac{t}{2} \rfloor}^2 \nonumber
\end{align}
which are only related to $t$.

Thus, it holds for \eqref{eq: Xt bound in cor short} that
\begin{align}
    X_{t_a}^2 &\leq \exp\!\Big(-\frac{\mu}{2}S_{t_a}\Big)X_{0}^2  + 4C_1^2\Xi_{\lfloor \frac{t}{2} \rfloor}^2 \sum_{k=0}^{t_a-1}\eta_k^2\,\exp\!\Big(-\frac{\mu}{2}(S_{t_a}-S_{k+1}) \Big)
    \nonumber \\
    &\quad + (\frac{8}{\mu}L_{s,f}^2+1)F_{\lfloor \frac{t}{2} \rfloor}^2  \sum_{k=0}^{t_a-1}\eta_k
    \exp\!\Big(-\frac{\mu}{2}(S_{t_a}-S_{k+1})\Big) +\Theta\!\Big(t_a \exp\{-\tfrac{\mu}{4} {t_a}^{1-\gamma}\}\Big),
\end{align}
where, after extracting the constants $F_{\lfloor \frac{t}{2} \rfloor}$ and $\Xi_{\lfloor \frac{t}{2} \rfloor}$ from the summation, we extend the lower summation index to $0$.

Applying Lemma \ref{lemma:Polynomial step size and exponentially weighted sums} with $\lambda = \frac{\mu}{2}$, we have
\[
\Big \| | \theta_{t_a} - \theta^*_{\lfloor \frac{t}{2} \rfloor-1} |_s \Big \|_q^2 \leq  \frac{4}{\mu}(\frac{8}{\mu}L_{s,f}^2+1)F_{\lfloor \frac{t}{2} \rfloor}^2 + 4C_1^2 \Xi^2_{\lfloor \frac{t}{2} \rfloor} \cdot O\!\big(t_a^{-\gamma}\big)+\exp\!\Big(-\frac{\mu}{2}S_{t_a}\Big)X_{0}^2 +\Theta\!\Big(t_a \exp\{-\tfrac{\mu}{4} {t_a}^{1-\gamma}\}\Big)
\]
We complete the proof by taking square root on both side, and applying inequality $\sqrt{\sum a_i} \leq \sum \sqrt{a_i}$.
\begin{equation}
    \Big \| | \theta_{t_a} - \theta^*_{\lfloor \frac{t}{2} \rfloor-1} |_s \Big \|_q \leq  \sqrt{\frac{4}{\mu}(\frac{8}{\mu}L_{s,f}^2+1)}F_{\lfloor \frac{t}{2} \rfloor} + 2C_1 \Xi_{\lfloor \frac{t}{2} \rfloor} \cdot O\!\big(t_a^{-\frac{\gamma}{2}}\big) + \Rcal_t,
\end{equation}
where $\Rcal_t$ decays faster than any polynomial rate. Specifically, when $a=1$, we obtain that
\begin{equation}
    \Big \| | \theta_{t} - \theta^*_{\lfloor \frac{t}{2} \rfloor-1} |_s \Big \|_q \leq  \sqrt{\frac{4}{\mu}(\frac{8}{\mu}L_{s,f}^2+1)}F_{\lfloor \frac{t}{2} \rfloor} + 2C_1 \Xi_{\lfloor \frac{t}{2} \rfloor} \cdot O\!\big(t^{-\frac{\gamma}{2}}\big) + \Rcal_t,
\end{equation}
\end{proof}
\begin{remark}[Proof constants]
\label{remark: proof constants}
For clarity, we collect the constants appearing in the proof. Let
\[
M_f:=\big\|\, |f|_2 \,\big\|_q,
\qquad 
C_1^2:=\max\{q,s\}-1.
\]
Furthermore, define
\begin{align*}
C_2
&:=Ck^{1/2} (C_{\alpha,\beta}+\frac{q^{1/2}}{\bar \rho_k}), \\
C_3
&:=C\left(
\frac{K_u^2}{\lambda_k(\Sigma_f)}
 M_f
+
K_u(k^{1/2}+q^{1/2})
+
k^{1/2} M_f
\right)d^{-1/2}, \\
C_4
&:=C\frac{L_{s,f}}{\mu}M_f(M_Ak^{1/2}\rho_k^{-1}+d^{-1/2}), \\
C_d
&:= C_3 d^{-1/2}.
\end{align*}
We also note that $C_{s,q}$ is a constant of order $\Ocal(p^{1/s})$ depending on $s,q$, and $\E[|\nabla l(\theta;f,y)|_q]$, introduced in Lemma \ref{lemma:Randomness of nabla_l}.

These constants are independent of the iteration index $t$ and are used only to simplify the notation in the proof.
\end{remark}

\end{document}